\pdfoutput=1

\documentclass[11pt]{article}

\usepackage[final]{acl}

\usepackage{times}
\usepackage{latexsym}
\usepackage{amssymb}
\usepackage{adjustbox}
\usepackage{subcaption}
\usepackage{graphicx}
\usepackage{spverbatim}
\usepackage[ruled]{algorithm2e}
\usepackage{booktabs}
\usepackage[misc]{ifsym}

\usepackage{multirow}
\usepackage[normalem]{ulem}
\useunder{\uline}{\ul}{}

\usepackage[T1]{fontenc}

\usepackage[utf8]{inputenc}

\usepackage{microtype}

\usepackage{inconsolata}

\usepackage{csquotes}

\title{Investigating Large Language Models for Complex Word Identification in Multilingual and Multidomain Setups}

\author{Răzvan-Alexandru Smădu$^{1}$, David-Gabriel Ion$^{1}$, Dumitru-Clementin Cercel$^{1,}$\textsuperscript{{\tiny\Letter}}, \\
\textbf{Florin Pop$^{1,2,3}$, Mihaela-Claudia Cercel$^{4,5}$}\\
$^{1}$National University of Science and Technology POLITEHNICA Bucharest, \\
Faculty of Automatic Control and Computers, Bucharest, Romania \\
$^{2}$National Institute for Research \& Development in Informatics - ICI Bucharest, \\
Bucharest, Romania \\
$^{3}$Academy of Romanian Scientists, Bucharest, Romania \\
$^{4}$Paris 1 Panthéon-Sorbonne University, Paris, France \\
$^{5}$University of Bucharest, Bucharest, Romania \\
\tt dumitru.cercel@upb.ro\\
}

\begin{document}
\maketitle
\begin{abstract}
Complex Word Identification (CWI) is an essential step in the lexical simplification task and has recently become a task on its own. Some variations of this binary classification task have emerged, such as lexical complexity prediction (LCP) and complexity evaluation of multi-word expressions (MWE). Large language models (LLMs) recently became popular in the Natural Language Processing community because of their versatility and capability to solve unseen tasks in zero/few-shot settings. Our work investigates LLM usage, specifically open-source models such as Llama 2, Llama 3, and Vicuna v1.5, and closed-source, such as ChatGPT-3.5-turbo and GPT-4o, in the CWI, LCP, and MWE settings. We evaluate zero-shot, few-shot, and fine-tuning settings and show that LLMs struggle in certain conditions or achieve comparable results against existing methods. In addition, we provide some views on meta-learning combined with prompt learning. In the end, we conclude that the current state of LLMs cannot or barely outperform existing methods, which are usually much smaller.
\end{abstract}

\section{Introduction}
\label{sec:introduction}

Complex word identification (CWI) aims to determine whether words or phrases are difficult for a target group of readers to understand. In the lexical simplification task, which targets replacing complex words and expressions with simplified alternatives~\cite{north2023deep}, CWI is the first step, and it was treated as part of this task until 2012 when it became a standalone task~\cite{shardlow-2013-comparison}.

CWI was initially addressed as a binary classification task \cite{paetzold-specia-2016-semeval}, identifying whether a word is complex in a given sentence. When the task became more popular \cite{north2023cwisurvey}, it was extended to the continuous domain as Lexical Complexity Prediction (LCP, also referred to as the probabilistic classification for CWI) \cite{yimam-etal-2018-report}, addressing multilingual and multidomain settings. Then, it was extended to multi-word expressions \cite{shardlow-etal-2021-semeval}. Recently, new datasets started to emerge in various languages and domains~\cite{ortiz-zambrano-montejo-raez-2021-clexis2,venugopal-etal-2022-cwid,10.1145/3582768.3582802,DBLP:journals/pdln/ZambranoEM23}.
Previous approaches to CWI ranged from using Support Vector Machines~\cite{s-p-etal-2016-amritacen} to deep neural networks based on Bidirectional Representation from Encoder Transformers~\cite{pan-etal-2021-deepblueai-semeval}, multi-task learning with domain adaptation~\cite{zaharia-etal-2022-domain}, and sequence modeling~\cite{gooding-kochmar-2019-complex}.

With the recent breakthrough in large language models (LLMs), particularly the work of OpenAI with Generative Pre-trained Transformer (GPT) models~\cite{radford2019language,brown2020language}, natural language processing has seen a significant leap. These models have shown the potential to improve performances on various tasks as we scale up the model size and the amount of training data. The announcement of ChatGPT\footnote{\url{https://openai.com/blog/chatgpt}} and its conversational capabilities has sparked a race in developing and fine-tuning new models for general purpose and domain-specific applications using PaLM~\cite{anil2023palm}, LLaMA~\cite{touvron2023llama,dubey2024llama}, Orca~\cite{mitra2023orca}, Mistral~\cite{jiang2023mistral}, GPT-4~\cite{openai2023gpt4}, GPT-4o\footnote{\url{https://openai.com/index/hello-gpt-4o/}}, and many others.

Our work aims to provide the current state of LLMs in addressing CWI and LCP compared against state-of-the-art approaches. We focus on evaluating open-source (pre-trained Llama 2~\cite{touvron2023llama}, Llama 3~\cite{dubey2024llama}, Vicuna~\cite{zheng2024judging}) and OpenAI’s close-source ChatGPT-3.5-turbo and GPT-4o.  We summarize the contributions as follows:
\begin{itemize}
    \item We evaluate LLMs in binary (discrete set of labels) and probabilistic classification (continuous space labels) on multidomain and multilingual corpora.
    \item We employ various techniques for prompting and fine-tuning.
    \item We show that LLMs struggle to address CWI and LCP tasks; however, in limited instances, they can achieve similar results with other, more lightweight approaches.
    \item In the end, we analyze and provide an insight into where the models struggle.
\end{itemize}

\section{Related Work}
\label{sec:related_work}

\subsection{Complex Word Identification}

CWI was previously addressed using straightforward baseline models based on feature engineering. For example, \citet{aroyehun-etal-2018-complex} compared CNN-based models with various feature engineering methods based on tree ensembles and features, such as inverse term frequency, parts-of-speech tagging, WordNet, and word2vec, achieving comparable results. \citet{finnimore-etal-2019-strong} proposed mono and cross-lingual models based on simple features and logistic regression, achieving similar results to more complex, language-specific state-of-the-art models. 
\citet{zaharia_2020_cwi} employed zero and few-shot learning techniques, along with Transformers and Recurrent Neural Networks, in a multilingual setting. \citet{gooding-kochmar-2019-complex} considered CWI a sequential task, using a bidirectional LSTM with word embeddings, character-level representations, and a language modeling objective to learn the complexity of words given the context. Other methods improved performances, employing graph-based~\cite{8999263}, domain adaptation~\cite{zaharia-etal-2022-domain}, and transformer-based models~\cite{pan-etal-2021-deepblueai-semeval,cheng-sheang-etal-2022-identification}.

\subsection{Large Language Models}

Recently, LLMs were successfully utilized in various generative tasks~\cite{pu2023summarization,DBLP:journals/corr/abs-2107-03374}. The new paradigm in solving other non-generative tasks is based on prompting pre-trained language models to perform the prediction task~\cite{10.1145/3560815,sun-etal-2023-text}. Fine-tuning models on instructions showed improved results in zero-shot settings, especially on unseen tasks~\cite{DBLP:conf/iclr/WeiBZGYLDDL22}. Prompt-based methods such as the use of demonstrations~\cite{min-etal-2022-rethinking}, intermediate reasoning steps by breaking down complex tasks into simpler subtasks (also known as a chain of thought)~\cite{DBLP:conf/nips/Wei0SBIXCLZ22}, and using LLMs to optimize their prompts~\cite{DBLP:conf/iclr/ZhouMHPPCB23} made zero-shot inference much more appealing due to reduced costs and more efficient than fine-tuning LLMs.

\subsection{Prompt Tuning and Meta-Learning}

Fine-tuning usually requires high computational costs, especially in the context of LLMs, to achieve good performances. Therefore, there was the need to find alternatives. Prompt Tuning ~\cite{lester2021power} is a soft-prompting technique for adapting a large language model for a custom task without training its parameters. While successful, prompt tuning falls short when applied in the few-shot learning regime, leading to a combination with meta-learning. MetaPrompting ~\cite{hou2022metaprompting} utilizes the meta-learning algorithms FOMAML ~\cite{finn2017model} and Reptile ~\cite{nichol2018first} to obtain optimized initialization embeddings. One shortcoming of this approach is the requirement for supervised training data during meta-learning, which is alleviated in other works ~\cite{pan2023self,huang2023learning} by generating the meta-training tasks in an unsupervised or self-supervised manner. Additionally, other works explored the use of an adaptable gradient regulating function ~\cite{li2023gradient} and domain-adversarial neural networks ~\cite{fang2023adversarial}, both techniques used to increase model generalization.

\section{Method}
\label{sec:method}

\subsection{Problem Formulation in the Pre-LLM Era}

Word complexity can be defined as absolute and relative~\cite{north2023cwisurvey}. Absolute complexity is determined by the objective linguistic properties (e.g., semantic, morphological, phonological). In contrast, relative complexity is related to the subjective speaker's point of view (e.g., familiarity with sound and meaning, culture, and context). In this work, we evaluate the relative complexity of words from non-native speakers' points of view. Considering an annotated dataset $D=\{(x_i, y_i)\}_{i=1}^N$ of N samples, the task can be viewed as a binary classification, CWI, where, given the pair $x_i=(C_i, w_i)$ of a sentence $C_i=(w_1, w_2, ...)$ and a word $w_i\in C_i$, the system outputs $y_i^{CWI} \in \{0,1\}$ (i.e., complex or non-complex)~\cite{paetzold-specia-2016-semeval}. A variation of the CWI task is to evaluate the complexity $y_i^{MWE} \in \{0,1\}$ of a multi-word expression $e_i = ( w_1, w_2, ... )$ containing multiple words $w_j$, $j=1:|e|$, from a given context $C_i$ (i.e., $x_i=(C_i, e_i)$) \cite{shardlow-etal-2021-semeval}. Later, CWI was considered in the continuous domain known as LCP, indicating the degree of difficulty $y_i^{LCP} \in [0,1]$ for the given word $w_i \in C_i$ in the context $C_i$ \cite{yimam-etal-2018-report}.

\subsection{Problem Formulation in the LLM Era}
\label{sec:cwi_problem_llm_era}

Starting from the previous formulation, we derive the formalism in the context of LLMs.

\textbf{Binary classification.} Given an example $x_i=(C_i, w_i)$, the model predicts if a given phrase $w_i$ from the sentence $C_i$ is complex. Especially in closed-source models, the access to the tokens' logits is limited (e.g., OpenAI's ChatGPT or GPT-4o). Therefore, we consider having access to only the model's predicted text labels ``true'' or ``false'' (or any equivalent form) without a confidence estimation.

\textbf{Probabilistic classification.} The model produces a real value between 0 and 1 in the existing approaches, representing the degree of complexity for $(C_i, w_i)$. LLMs are known to suffer from hallucinations~\cite{openai2023gpt4}, and reliably predicting real values is challenging. We abide by \citet{liu-etal-2023-g}'s solution for estimating the scoring function. In a nutshell, instead of letting the model predict the probability as a number, we let the model generate discrete signals, and then we estimate the score through averaging. Specifically, we prompt the model to predict on the 5-point Likert scale, in natural language, one of ``very easy'', ``easy'', ``neutral'', ``difficult'', or ``very difficult''. This scale is converted into a numerical representation using the following mapping: very easy - 0, easy - 0.25, neutral - 0.5, difficult - 0.75, and very difficult - 1. Since LLMs output tokens from a probability distribution, we set the temperature to a higher value (in our experiments, we use 0.8) to increase response variability. The numerical representation from LLM's output is denoted as $s_k \in S$ for a sampling step k, with $S = \{ 0, 0.25, 0.5, 0.75, 1 \}$. The model's probability of outputting one 5-point Likert score is $p(s_k)$. The final score is:

\begin{equation}
score = \mathbb{E}_{p}[S] = \sum_{s \in S} p(s) \cdot s
\end{equation}

For experiments, we use the sample mean estimator $\overline{score} = \frac{1}{K} \sum_{k=1}^K s_k$ of $K$ sampling steps.

In essence, we want to simulate the data annotation process involved. Instead of employing multiple human annotators, we use the same model that produces a more randomized output by setting a higher temperature. Note that randomized LLM outputs do not translate to randomized output labels. Variance in the results can be used as a confidence estimator, and we show (see \S \ref{sec:predictions_ft}) that models act more deterministically despite setting a high-temperature value. However, even for fine-tuned models, the variance is not 0; thus, it does not collapse to the multi-class classification setup (see \S \ref{sec:predictions_ft}).

\subsection{Prompting LLMs}
\label{label:prompting_llms}

We set the system prompt of the model with the task and how to format the output. Then, we prompt it through the user prompt to predict the example label. Each example is prompted individually to avoid leaking knowledge from other examples.

All system prompts are listed in the Appendix \ref{sec:prompting}. These are obtained after prompt engineering, i.e., multiple trial-and-errors. Our focus was on optimizing the prompt length and the LLM's performance. Because performing prompt engineering on every LLM is time-consuming, we optimized the prompts on ChatGPT-3.5-turbo and LLama 2 7B models on LCP and CWI English, each as a whole task, respectively. The prompts for German and Spanish are translations of the English prompts.

We investigate prompting strategies for zero-shot and few-shot settings. In every setting, we evaluate with and without employing the chain of thoughts approach~\cite{DBLP:conf/nips/Wei0SBIXCLZ22} to reduce hallucination. Therefore, the model needs to reproduce the original sentence and the word and then, before the final answer, provide a short proof about the reason for the response.

A similar prompting procedure is applied for open-source and close-source models. The difference mainly lies in how we format the query using either the Chat Message API\footnote{\url{https://platform.openai.com/docs/api-reference/chat/create}} for OpenAI's models or the chat template available in the HuggingFace tokenizer\footnote{\url{https://huggingface.co/docs/transformers/en/chat_templating}}. We provide details regarding the prompting format in Appendix \ref{sec:prompting} and the evaluation protocol in Appendix \ref{sec:eval_protocol}.

\subsection{Fine-tuning LLMs}

For fine-tuned models, we prepare the dataset to include a minimal system prompt and the query with the answer. First, we discretize the probabilities similar to \citet{shardlow-etal-2021-semeval}: scores between 0 and 0.2 are very easy, between 0.2 and 0.4 are easy, between 0.4 and 0.6 are neutral, between 0.6 and 0.8 are difficult, and between 0.8 and 1 are very difficult. Next, we apply the prompt for open-source models using the template specific to the model available in the HuggingFace tokenizer. Fine-tuning OpenAI's ChatGPT models involves uploading the training and validation files and starting the training job. After finishing the fine-tuning step, we follow a procedure similar to \S \ref{label:prompting_llms}. However, in this setting, the model does not generate a demonstration but directly generates the answer.

\subsection{Meta-Learning}

Meta-learning tries to adapt a model to a new task by acquiring knowledge from multiple learning tasks. Our approach involves training the model on various tasks using soft prompting methods. Therefore, we propose using FOMAML \cite{finn2017model} in conjunction with Prompt tuning \cite{lester2021power} and P-tuning v2~\cite{liu-etal-2022-p} to optimize the initial parameters of our adapters. In the prompt tuning setting, we prepend the input prompt with several virtual tokens learned during training. P-tuning v2 takes this idea to the next level by adding trainable prompts to different layers in the model. The model is only provided with the user prompt during input construction due to the system prompt's non-standardized nature and the variety of tasks it employs. We select 45 tasks from the BIG-bench suite, as detailed in \S \ref{subsection:datasets} and Appendix \ref{appendix:metalearning}.

Most optimization-based meta-learning algorithms, including FOMAML, involve copying the model's parameters trained so far, sampling a random set of tasks, training the parameters for several steps, and then performing the optimization step. The inner training steps are task-dependent. In our case, we use the same causal loss as in fine-tuning. The algorithm is described in Algorithm \ref{alg:fomaml}. It is noteworthy that training requires more computational resources than fine-tuning.

\SetKwComment{Comment}{//}{}
\begin{algorithm}
\caption{FOMAML algorithm}
\label{alg:fomaml}
\KwData{$\alpha$, $\beta$ learning rates, $n$ inner steps}
Randomly initialize $\theta$\;
\While{not converged}{
    \Comment{Sample support and query sets}
    Sample task $\mathcal{T} = (\mathcal{T}_\mathrm{s}, \mathcal{T}_\mathrm{q})$\;
    \Comment{Inner training loop}
    $\theta'_0 \gets \theta$\;
    \For{$i=0$ \KwTo $n$}{
        $\theta'_{i+1} \gets \theta'_i - \alpha\nabla_{\theta'_i}\mathcal{L}(\theta'_i, \mathcal{T}_\mathrm{s})$\;
    }
    \Comment{FOMAML optimization}
    \Comment{$\nabla_{\theta}\mathcal{L}(\theta'_n, \mathcal{T}_\mathrm{q}) \approx \nabla_{\theta'_n}\mathcal{L}(\theta'_n, \mathcal{T}_\mathrm{q})$}
    $\theta \gets \theta - \beta \nabla_{\theta'_n}\mathcal{L}(\theta'_n, \mathcal{T}_\mathrm{q})$ \Comment*{or Adam}
}
\end{algorithm}

\section{Experimental Setup}

\subsection{Datasets}
\label{subsection:datasets}

\textbf{CWI 2018 Shared Dataset.} It was proposed at the CWI Shared Task in 2018~\cite{yimam-etal-2018-report} and addressed English multidomain and multilingual settings. The English split contains samples from three sources (News, WikiNews, and Wikipedia) totaling approx. 35,000 samples. In the multilingual setting, the dataset features German and Spanish with approx. 8,000 and 17,600 samples, respectively, and a French test set containing 2,251 samples. We present the split in train, validation, and test sets in Table~\ref{tab:data_splits}. The dataset was developed to address binary and probabilistic classification tasks by assigning probabilities and labels such that samples with 0\% probability are non-complex and others as complex. We consider only the binary classification tasks (see Limitations \ref{sec:limitations}). 
 
\begin{table}[!htb]
\centering
\small
\begin{tabular}{lccc}
\toprule
\multicolumn{1}{c}{\textbf{Dataset}} & \textbf{Train} & \textbf{Valid.} & \textbf{Test} \\
\midrule
\hline
& \multicolumn{3}{c}{\textit{CWI 2018}} \\
\textbf{English News}      & 14,002 & 1,764 & 2,095 \\
\textbf{English WikiNews}  &  7,746 &   870 & 1,287 \\
\textbf{English Wikipedia} &  5,551 &   694 &   870 \\
\textbf{German}            &  6,151 &   795 &   959 \\
\textbf{Spanish}           & 13,750 & 1,622 & 2,233 \\
\hline
\hline
& \multicolumn{3}{c}{\textit{CompLex LCP 2021}} \\
\textbf{Single-Word} & 7,662 & 421 & 917 \\
\textbf{Multi-Word}  & 1,517 & 99 & 184 \\
\bottomrule
\end{tabular}
\caption{Dataset splits in train/validation/test for CWI 2018 and CompLex LCP 2021 datasets.}
\label{tab:data_splits}
\end{table}

\textbf{CompLex LCP 2021.} Proposed at SemEval 2021 Task 1~\cite{shardlow-etal-2021-semeval}, CompLex LCP 2021 comprises around 10,000 sentences in English from three domains: European Parliament proceedings, the Bible, and biomedical literature. The data is split across two tasks: single-word (Single-Word) and multi-word expressions (Multi-Word). The complexity is provided as continuous values between 0 and 1, addressed as the probabilistic classification task. The average complexity is 0.3 for single and 0.42 for multi-word. The data splits are shown in Table~\ref{tab:data_splits}.

\textbf{BIG-bench.} This recently proposed benchmark ~\cite{srivastava2023beyond} contains over 200 tasks for evaluating large language models. We use this collection as part of the meta-learning stage in our pipeline. Since not all tasks are suitable for our use case, we select 45 tasks (detailed in Appendix \ref{appendix:metalearning}) and only pre-train on those. The unsuitable tasks include non-categorical responses, requirements for external knowledge, or might be too dissimilar to the target task. Our main task selection criteria were prompt length and similarity to the complex word identification task since we wanted as much intrinsic knowledge to be transferred on fine-tuning as possible.

\subsection{Baselines}

We compare against top-performing methods at CWI 2018 Shared task and LCP 2021. Camb~\cite{gooding-kochmar-2018-camb} employs heterogeneous features combined with an ensemble of AdaBoost classifiers. The TMU system~\cite{kajiwara-komachi-2018-complex} uses a random forest classifier with multiple hand-crafted features. ITEC~\cite{de-hertog-tack-2018-deep} combines CNN and LSTM layers. SB@GU \cite{alfter-pilan-2018-sb} employs Random Forest and Extra Tree on top of hand-crafted features. In addition, we include the XLM-RoBERTa-based approach combined with text simplification and domain adaptation \cite{zaharia-etal-2022-domain}, the MLP combined with Sent2Vec solution \citet{almeida-etal-2021-c3sl}, and $RoBERTa_{LARGE}$ with an ensemble of RoBERTa-based models (LR-Ensemble) \cite{pan-etal-2021-deepblueai-semeval}.

\subsection{Models}

We evaluate several open- and closed-source LLMs. Specifically, we choose Llama 2 (7B and 13B parameters)~\cite{touvron2023llama}, Vicuna (7B and 13B parameters) \cite{zheng2024judging}, and Llama 3 8B \cite{dubey2024llama} for open-source models. For closed-source models, we employ OpenAI's ChatGPT-3.5-turbo and GPT-4o~\cite{openai2023gpt4} specifically for their relatively lower prices than GPT-4 (see also Appendix \ref{appendix:costs}). The chat model is used in the zero- and few-shot settings, and the base or instruct model is used as the starting checkpoint for fine-tuning. Details regarding specific checkpoints for all models are listed in Appendix~\ref{app:checkpoints}.

\subsection{Evaluation Metrics}

We adopt the same evaluation methodology in \citet{shardlow-etal-2021-semeval} for CompLex LCP 2021 and \citet{yimam-etal-2018-report} for CWI 2018 datasets. Therefore, we employ Pearson correlation (P) and Mean Average Error (MAE) on the LCP task and F1-score (F1) and Accuracy (Acc.) for the CWI task. We report all results on a single run for CWI and multiple runs (described by $K$ in \S \ref{sec:cwi_problem_llm_era}) for LCP.

\section{Results}

\begin{table*}[!t]
\centering
\small
\setlength{\tabcolsep}{3pt}

\begin{tabular}{lcccccccccccccc}
\toprule
\multicolumn{1}{c}{\multirow{3}{*}{\textbf{Model}}} &
\multicolumn{10}{c}{\textbf{CWI 2018}} &
\multicolumn{4}{c}{\textbf{CompLex LCP 2021}}
\\
  &
\multicolumn{2}{c}{\textbf{EN-N}} &
\multicolumn{2}{c}{\textbf{EN-WN}} &
\multicolumn{2}{c}{\textbf{EN-W}} &
\multicolumn{2}{c}{\textbf{DE}} &
\multicolumn{2}{c}{\textbf{ES}} &
\multicolumn{2}{c}{\textbf{Single-Word}} &
\multicolumn{2}{c}{\textbf{Multi-Word}}
\\
  &
\textbf{F1$\uparrow$} & \textbf{Acc$\uparrow$} &
\textbf{F1$\uparrow$} & \textbf{Acc$\uparrow$} &
\textbf{F1$\uparrow$} & \textbf{Acc$\uparrow$} &
\textbf{F1$\uparrow$} & \textbf{Acc$\uparrow$} &
\textbf{F1$\uparrow$} & \textbf{Acc$\uparrow$} &
\textbf{P$\uparrow$} & \textbf{MAE$\downarrow$} &
\textbf{P$\uparrow$} & \textbf{MAE$\downarrow$}
\\
\midrule
\hline
  & \multicolumn{14}{c}{\textit{Baseline}} \\
\textbf{Camb}                 & \textbf{87.4} & - & \textbf{84.0} & - & \textbf{81.2} & - & -             & - & -             & - & -              & -              & -              & -              \\
\textbf{ITEC}                 & 86.4          & - & 81.1          & - & 78.1          & - & -             & - & 76.3          & - & -              & -              & -              & -              \\
\textbf{TMU}                  & 86.3          & - & 78.7          & - & 76.2          & - & \textbf{74.5} & - & \textbf{77.0} & - & -              & -              & -              & -              \\
\textbf{SB@GU}                & 83.3          & - & 80.3          & - & 78.3          & - & 74.3          & - & 72.8          & - & -              & -              & -              & -              \\
\textbf{MLP+Sent2Vec}         & -             & - & -             & - & -             & - & -             & - & -             & - & .4598          & .0866          & .3941          & .1145          \\
\textbf{XLM-RoBERTa-based}    & -             & - & -             & - & -             & - & -             & - & -             & - & .7744          & .0652          & .8285          & .0708          \\
\textbf{RoBERTa{\tiny LARGE}} & -             & - & -             & - & -             & - & -             & - & -             & - & \textbf{.7903} & \textbf{.0648} & .7900          & .0753          \\
\textbf{LR-Ensemble}          & -             & - & -             & - & -             & - & -             & - & -             & - &  -             & -              & \textbf{.8612} & \textbf{.0616} \\
\hline
\hline
  & \multicolumn{14}{c}{\textit{Zero-shot}} \\
\textbf{Llama-2-7b-chat}      & 32.1          & 63.8          & 19.3          & 57.7          & 37.8          & 55.3          & 30.7          & 54.0          & 49.0          & 51.5          & .3133          & .3061          & .5200          & .2316          \\
\textbf{Llama-2-13b-chat}     & 11.9          & 63.2          & 12.5          & 58.8          & 20.1          & 53.3          & 56.7          & 54.9          & 44.2          & 40.8          & .2040          & .2475          & .3613          & .1737          \\
\textbf{Vicuna-v1.5-7b}       & 22.6          & 59.8          & 25.3          & 56.8          & 27.8          & 51.6          & 18.2          & 50.3          & 51.5          & 60.6          & .3108          & .3684          & .4502          & .2680          \\
\textbf{Vicuna-v1.5-13b}      & 13.0          & 63.1          & 12.0          & 59.1          & 16.0          & 51.7          & 53.4          & 59.6          & 11.7          & 50.8          & .2189          & .1987          & .4436          & \textbf{.1425} \\
\textbf{Llama-3-8b-chat}      & 43.0          & 70.0          & 29.3          & 62.9          & 43.1          & 61.5          & 50.2          & 60.3          & 10.7          & 55.9          & .3816          & .1880          & .6271          & .1626          \\
\textbf{ChatGPT-3.5-turbo}    & 40.1          & 69.5          & 37.0          & 64.6          & 45.6          & 62.0          & 53.3          & 60.1          & 35.3          & 63.3          & .5352          & \textbf{.1447} & .6284          & .1529          \\
\textbf{GPT-4o}               & \textbf{65.9} & \textbf{76.8} & \textbf{64.2} & \textbf{73.2} & \textbf{66.8} & \textbf{71.0} & \textbf{63.3} & \textbf{73.0} & \textbf{68.9} & \textbf{75.8} & \textbf{.5953} & .2346          & \textbf{.7753} & .2377          \\
\hline
\hline
  & \multicolumn{14}{c}{\textit{Zero-shot CoT}} \\
\textbf{Llama-2-7b-chat}      & 56.5          & 56.2          & 61.1          & 59.1          & 62.8          & 57.2          & 57.2          & 45.7          & 56.7          & 46.0          & .3617          & .1698          & .5040          & .1632          \\   
\textbf{Llama-2-13b-chat}     & 54.9          & 61.6          & 49.1          & 56.5          & 57.8          & 54.7          & 55.5          & 53.2          & 57.7          & 49.7          & .4335          & .1393          & .5905          & .1118          \\  
\textbf{Vicuna-v1.5-7b}       & 38.1          & 60.3          & 38.5          & 58.7          & 52.4          & 56.9          & 20.6          & 50.0          & 55.4          & 58.3          & .2558          & .1504          & .4916          & .1310          \\    
\textbf{Vicuna-v1.5-13b}      & 32.9          & 66.8          & 27.9          & 61.8          & 33.3          & 55.9          & 42.1          & 58.9          & 29.8          & 58.7          & .4664          & \textbf{.0922} & .6357          & \textbf{.1049} \\     
\textbf{Llama-3-8b-chat}      & 50.5          & 66.9          & 45.7          & 63.0          & 61.0          & 64.4          & 49.3          & 59.9          & 34.9          & 55.5          & .4617          & .1507          & .6923          & .1167          \\   
\textbf{ChatGPT-3.5-turbo}    & 64.0          & 69.9          & 64.0          & 68.0          & 66.7          & 64.6          & 59.1          & 58.8          & 63.4          & 47.2          & .5901          & .2012          & .6836          & .1624          \\ 
\textbf{GPT-4o}               & \textbf{72.9} & \textbf{75.7} & \textbf{74.9} & \textbf{75.4} & \textbf{76.1} & \textbf{73.9} & \textbf{69.5} & \textbf{69.8} & \textbf{70.3} & \textbf{71.9} & \textbf{.6228} & .2145          & \textbf{.7389} & .2586          \\            
\hline
\hline
  & \multicolumn{14}{c}{\textit{Few-shot}} \\
\textbf{Llama-2-7b-chat}      & 61.4          & 63.9          & \textbf{63.2} & 55.5          & \textbf{70.6} & 61.4          & 55.8          & 43.0          & 57.9          & 50.1          & .1409          & .2021          & .5016          & .1781          \\  
\textbf{Llama-2-13b-chat}     & 46.2          & 65.3          & 51.2          & 65.0          & 52.2          & 60.8          & 53.5          & 56.8          & 49.5          & 62.1          & .2010          & .2178          & .4412          & .2118          \\ 
\textbf{Vicuna-v1.5-7b}       & 43.9          & 64.5          & 46.0          & 59.5          & 48.1          & 57.8          & 50.4          & 58.8          & 45.1          & 63.3          & .1789          & .1767          & .4641          & .1522          \\   
\textbf{Vicuna-v1.5-13b}      & 42.3          & 65.3          & 53.4          & 63.7          & 54.9          & 59.0          & 54.6          & 56.9          & 44.6          & 63.6          & .2686          & .1871          & .4401          & .2157          \\  
\textbf{Llama-3-8b-chat}      & 53.0          & 72.9          & 55.3          & 69.9          & 61.7          & \textbf{67.6} & 56.6          & 60.3          & 54.1          & 61.7          & .3102          & .1730          & .5843          & .1796          \\  
\textbf{ChatGPT-3.5-turbo}    & 52.1          & 72.3          & 44.5          & 66.4          & 53.8          & 65.1          & 55.4          & 65.5          & 42.5          & 72.7          & .6385          & \textbf{.0979} & .6742          & \textbf{.1197} \\  
\textbf{GPT-4o}               & \textbf{63.8} & \textbf{76.6} & 60.7          & \textbf{71.8} & 58.3          & 66.8          & \textbf{60.2} & \textbf{75.9} & \textbf{66.0} & \textbf{75.7} & \textbf{.7111} & .1859          & \textbf{.8284} & .2195          \\           
\hline
\hline
  & \multicolumn{14}{c}{\textit{Few-shot CoT}} \\
\textbf{Llama-2-7b-chat}      & 54.9          & 64.2          & \textbf{60.7} & 53.0          & \textbf{67.5} & 57.5          & 43.5          & 45.3          & 58.6          & 59.0          & .4683          & .1988          & .5920          & .2170          \\   
\textbf{Llama-2-13b-chat}     & 45.2          & 68.9          & 56.3          & 60.9          & 59.3          & 57.5          & 58.0          & 59.0          & 58.5          & 63.5          & .5796          & \textbf{.1289} & .6468          & .1615          \\  
\textbf{Vicuna-v1.5-7b}       & 39.6          & 65.5          & 44.0          & 63.6          & 56.2          & 63.6          & 32.7          & 60.9          & 56.4          & 65.3          & .5832          & .1315          & .6463          & .1444          \\    
\textbf{Vicuna-v1.5-13b}      & 49.1          & 70.2          & 45.8          & 65.1          & 50.9          & 62.8          & 59.2          & 61.8          & 57.0          & 73.4          & .5576          & .1477          & .6832          & .1524          \\   
\textbf{Llama-3-8b-chat}      & 34.8          & 66.5          & 47.4          & 63.6          & 53.6          & 61.4          & 61.6          & 67.7          & 58.2          & 72.6          & .2723          & .2048          & .7148          & \textbf{.1146} \\   
\textbf{ChatGPT-3.5-turbo}    & 58.3          & 72.4          & 51.4          & 66.0          & 55.5          & 65.2          & \textbf{64.3} & 72.9          & \textbf{68.1} & \textbf{73.9} & .7175          & .1421          & .7568          & .1707          \\ 
\textbf{GPT-4o}               & \textbf{66.1} & \textbf{77.2} & 53.3          & \textbf{68.9} & 61.2          & \textbf{68.5} & 53.5          & \textbf{73.8} & 60.5          & 73.5          & \textbf{.7594} & .1609          & \textbf{.8211} & .1850          \\            
\hline
\hline
  & \multicolumn{14}{c}{\textit{Fine-tuned}} \\
\textbf{Llama-2-7b-ft}        & 78.0          & 82.9          & 78.2          & 81.1          & 77.4          & 76.7          & 70.5          & 75.7          & 74.6          & 79.4          & .7734          & \textbf{.0670} & .7919          & .0767          \\       
\textbf{Llama-2-13b-ft}       & 77.6          & 83.3          & 77.7          & 81.3          & 73.1          & 74.6          & \textbf{70.8} & 76.6          & \textbf{75.3} & \textbf{81.0} & \textbf{.7815} & .0798          & \textbf{.8318} & \textbf{.0718} \\      
\textbf{Vicuna-v1.5-7b-ft}    & 80.2          & 84.3          & 76.8          & 79.3          & 77.2          & 77.2          & 67.5          & 74.3          & 73.0          & 79.0          & .7613          & .0840          & .7862          & .0782          \\        
\textbf{Vicuna-v1.5-13b-ft}   & 81.2          & 85.2          & 77.4          & 80.3          & \textbf{81.2} & \textbf{80.8} & 70.0          & 75.4          & 72.1          & 76.7          & .7530          & .0914          & .8000          & .0763          \\       
\textbf{Llama-3-8b-ft}        & \textbf{82.1} & \textbf{86.3} & 79.6          & 82.9          & 76.8          & 77.5          & \textbf{70.8} & 76.9          & 72.2          & 78.1          & .7497          & .0909          & .7800          & .0834          \\       
\textbf{ChatGPT-3.5-turbo-ft} & 80.7          & 83.9          & \textbf{80.9} & \textbf{83.1} & 80.2          & 79.4          & 66.6          & \textbf{78.0} & 74.4          & 78.1          & .7397          & .1372          & .7537          & .1815          \\ 
\bottomrule                                                                                                                                                   
\end{tabular}

\caption{The results on the test sets from CWI 2018 and CompLex LCP 2021 datasets. Notation: EN - English, DE - German, and ES - Spanish; for English datasets, N - news domain, WN - WikiNews, W - Wikipedia. In bold, we denote the best scores.}
\label{tab:cwi_lcp_f1} 
\end{table*}

\subsection{English Multidomain Setup}

We present the results in Table~\ref{tab:cwi_lcp_f1}. The top-performing LLMs are ChatGPT-3.5-turbo and GPT-4o, which generally achieve higher scores than the open-source LLMs, especially in the zero- and few-shot settings. When fine-tuning, we notice that open-source models achieve competitive results with ChatGPT-3.5-tubo-ft. Meanwhile, all fine-tuned models match or outperform zero- and few-shot closed-source models. Fined-tuned ChatGPT-3.5-turbo-ft achieves over 80\% F1-score, while the highest scores for English-News and English-Wikipedia are surpassed by Llama-3-8b-ft and Vicuna-v1.5-13b-ft, respectively, by 1-2\%. However, LLMs fall behind baseline classifiers that are more lightweight and easier to run. On the Wikipedia domain, Vicuna-v1.5-13b-ft achieves the same F1-score as Camb. We found the main limitation is the task hallucination -- the model does not reproduce the task it needs to solve (see \S \ref{sec:llm_task_understanding}).

\subsection{Multilingual Setup}
Like the multidomain setup, the fine-tuned LLMs achieve the highest score in German and Spanish datasets (see Table \ref{tab:cwi_lcp_f1}). Notably, Llama-2-7b-ft and ChatGPT-3.5-turbo-ft achieve higher scores than the submitted systems, but we cannot consider LLMs a good solution for this problem as these models achieve under 80\% in F1-score. Because the models were trained with multilingual corpora, they perform similarly in German and Spanish. Zero-shot combined with the chain of thought performs better than other prompting techniques in most cases and falls behind fine-tuning by a small margin, especially in the case of the German split.

\subsection{Lexical Complexity Prediction Setup}
On the CompLex LCP 2021 dataset, \citet{pan-etal-2021-deepblueai-semeval} achieved the best scores. Refer to Table~\ref{tab:cwi_lcp_f1} for the results. Fine-tuned LLM-based models outperform RoBERTa-based models on the Multi-Word task, the best-performing model being Llama-2-13b-ft. However, RoBERTa{\tiny LARGE} has 355M parameters, while Llama 2 13B has 37 times more parameters, and the performance difference is only about 5\% on the Pearson correlation. In addition, RoBERTa{\tiny LARGE} outperforms all models on the single-word expressions task. The few-shot method, combined with the chain of thought, usually performs better when considering the prompting techniques.

\subsection{Meta-Learning Setup}

Due to the high computational cost of our meta-learning algorithm, we only test on Llama 2 7B, both the chat and base versions. We also only test using data in English since changing the meta-training tasks requires a new suite that can be difficult to obtain in the multilingual setting. The reasons are low data availability and lack of knowledge in the other languages since Llama 2 was trained predominantly on English data.

The soft prompting techniques show results comparable to the zero-shot regime, as illustrated in Table ~\ref{tab:ml_all}. All combinations of methods and chat versus base versions of the LLMs show similar performances. In addition, we show the performance impact of varying the optimization steps our meta-learner goes through before the evaluation process in Table ~\ref{tab:ml_wikinews}. The best number of steps is between 5 and 15 for the chat versions of Llama 2, as opposed to between 50 and 100 for the base model.

\begin{table}[!htbp]
\centering
\small
\setlength{\tabcolsep}{3pt}
\begin{tabular}{lcccccc}
\toprule
\multicolumn{1}{c}{\multirow{2}{*}{\textbf{Model}}} &
  \multicolumn{2}{c}{\textbf{News}} &
  \multicolumn{2}{c}{\textbf{WikiNews}} &
  \multicolumn{2}{c}{\textbf{Wikipedia}} \\ 
\multicolumn{1}{c}{} &
  \multicolumn{1}{c}{\textbf{F1$\uparrow$}} &
  \multicolumn{1}{c}{\textbf{Acc$\uparrow$}} &
  \multicolumn{1}{c}{\textbf{F1$\uparrow$}} &
  \multicolumn{1}{c}{\textbf{Acc$\uparrow$}} &
  \multicolumn{1}{c}{\textbf{F1$\uparrow$}} &
  \multicolumn{1}{c}{\textbf{Acc$\uparrow$}} \\
\midrule
\hline
& \multicolumn{6}{c}{\textit{P-tuning}}                                                    \\
\textbf{Llama-2-7b-chat} & 50.3 & \textbf{46.7} & \textbf{66.8} & 51.2 & \textbf{65.3} & \textbf{51.9} \\ 
\textbf{Llama-2-7b}      & \textbf{53.8} & 41.2 & 65.4 & \textbf{52.0} & 61.2 & 49.8 \\
\hline
\hline
& \multicolumn{6}{c}{\textit{Prompt-tuning}}                                                    \\
\textbf{Llama-2-7b-chat} & 46.8 & \textbf{53.0} & \textbf{66.1} & 51.4 & \textbf{61.9} & 48.6 \\ 
\textbf{Llama-2-7b}      & \textbf{51.6} & 43.8 & 64.0 & \textbf{53.2} & 61.3 & \textbf{49.7} \\
\bottomrule
\end{tabular}
\caption{Results on the multi-domain English test set from CWI 2018 Shared Dataset in the few-shot learning regime, starting from the meta-learned models. In bold, we denote the best score.}
\label{tab:ml_all} 
\end{table}

\begin{table}[!htbp]
\centering
\small
\setlength{\tabcolsep}{3pt}
\begin{tabular}{lcccccc}
\toprule
\multicolumn{1}{c}{\multirow{2}{*}{\textbf{Model}}} &
  \multicolumn{6}{c}{\textbf{Fine-tuning inner steps}} \\
\multicolumn{1}{c}{} &
  \multicolumn{1}{c}{\textbf{5}} &
  \multicolumn{1}{c}{\textbf{10}} &
  \multicolumn{1}{c}{\textbf{15}} &
  \multicolumn{1}{c}{\textbf{25}} &
  \multicolumn{1}{c}{\textbf{50}} &
  \multicolumn{1}{c}{\textbf{100}} \\
\midrule
\hline
& \multicolumn{6}{c}{\textit{P-tuning}}                                                    \\
\textbf{Llama-2-7b-chat} & \textbf{66.8} & 66.3 & 66.7 & 64.2 & 65.7 & 59.7 \\ 
\textbf{Llama-2-7b}      & 60.7 & 62.0 & \textbf{65.4} & 64.8 & \textbf{65.4} & 64.3 \\ 
\hline
\hline
& \multicolumn{6}{c}{\textit{Prompt-tuning}}                                                    \\
\textbf{Llama-2-7b-chat} & 64.8 & 65.7 & \textbf{66.1} & 65.4 & 62.4 & 61.8 \\ 
\textbf{Llama-2-7b}      & \textbf{64.2} & 58.8 & 55.8 & 52.2 & 59.6 & 64.0 \\ 
\bottomrule
\end{tabular}
\caption{The influence of the fine-tuning inner steps on the F1-score when evaluating on the CWI 2018 English WikiNews test set. In bold, we denote the best score.}
\label{tab:ml_wikinews} 
\end{table}

\section{Discussions}

\subsection{Task Hallucination}
\label{sec:llm_task_understanding}

We investigate the hallucination effects of LLMs reproducing the task firsthand before outputting the prediction. Before providing the final answer, we check whether the model correctly copied the sentence and target word from the input query. We report the sentence error rate (S) and the word error rate (W). In this scenario, we mainly focus on the prompting settings. The results are presented in Table~\ref{tab:llm_understand_err}. In the CWI setting, we enforced the output structure using the Outlines library~\cite{willard2023efficient}. We notice that the larger the model, the lower the error rates. The ChatGPT-3.5-turbo and GPT-4o models obtain the lowest error rates, while the Llama 2 7B model obtains the highest. The models usually struggle to recall the correct word to be evaluated. Investigating the outputs, we mostly see that the model considers more context than the target, for example, "America" (ground truth) vs "South America" (extracted by LLM). Other error cases we identified were extracting completely different words from the sentence. For example, the target ``years'' was replaced by Llama-2-13b-chat with ``Aegyptosaurus''. Text locality is not always the main reason; in the first example, we have text locality; however, in the second example, the words were in different parts of the sentence. Additionally, we note that few-shot prompting also reduces the error rates because of the emerging pattern from the system prompt.

In the LCP setting, we consider all the sampling runs, and thus, we report the average and standard deviation across those runs. We report lower error rates. Like the previous setting, employing few-shot prompting reduces the sentence and word error rates. In addition, ChatGPT-3.5-turbo and GPT-4o achieve error rates very close to 0, meaning that the models can produce better demonstrations for the results. In the zero-shot settings, the models struggle to recall the word and the sentence.

\begin{table*}[!ht]
\small
\centering
\setlength{\tabcolsep}{3pt}
\begin{tabular}{lcccccccccccccc}
\toprule
\multicolumn{1}{c}{\multirow{3}{*}{\textbf{Model}}} &
\multicolumn{10}{c}{\textbf{CWI Shared 2018}} &
\multicolumn{4}{c}{\textbf{CompLex LCP 2021}}
\\
&
\multicolumn{2}{c}{\textbf{EN-N}} &
\multicolumn{2}{c}{\textbf{EN-WN}} &
\multicolumn{2}{c}{\textbf{EN-W}} &
\multicolumn{2}{c}{\textbf{ES}} &
\multicolumn{2}{c}{\textbf{DE}} &
\multicolumn{2}{c}{\textbf{Single-Word}} &
\multicolumn{2}{c}{\textbf{Multi-Word}}
\\
&
\textbf{S $\downarrow$} & \textbf{W $\downarrow$} &
\textbf{S $\downarrow$} & \textbf{W $\downarrow$} &
\textbf{S $\downarrow$} & \textbf{W $\downarrow$} &
\textbf{S $\downarrow$} & \textbf{W $\downarrow$} &
\textbf{S $\downarrow$} & \textbf{W $\downarrow$} &
\textbf{S $\downarrow$} & \textbf{W $\downarrow$} &
\textbf{S $\downarrow$} & \textbf{W $\downarrow$}
\\
\midrule
\hline
& \multicolumn{14}{c}{\textit{Zero-shot}} \\
\textbf{Llama-2-7b-chat} & 0.4 & 2.3 & 1.4 & 2.4 & 1.2 & 2.9 & 12.9 & 7.1 & 15.6 & 3.8 & $2.1_{\pm 0.2}$ & $5.1_{\pm 0.4}$ & $1.0_{\pm 0.7}$ & $0.7_{\pm 0.3}$ \\ 
\textbf{Llama-2-13b-chat} & 0.1 & 0.6 & 0.3 & 1.1 & 0.4 & 1.1 & 6.9 & 6.9 & 17.8 & 8.2 & $1.9_{\pm 0.3}$ & $8.1_{\pm 0.8}$ & $1.1_{\pm 0.5}$ & $0_{\pm 0}$ \\    
\textbf{Vicuna-v1.5-7b} & 0.2 & 0.1 & 1.1 & 0.1 & 0.2 & 0.2 & 2.0 & 1.0 & 1.7 & 0.4 & $1.6_{\pm 0.3}$ & $0.2_{\pm 0.2}$ & $0.7_{\pm 0.5}$ & $0_{\pm 0}$ \\      
\textbf{Vicuna-v1.5-13b} & 0 & 0.1 & 0.1 & 0 & 0.2 & 0.2 & 0.3 & 1.2 & 5.7 & 4.7 & $2.7_{\pm 0.5}$ & $1.0_{\pm 0.3}$ & $1.2_{\pm 0.5}$ & $0.1_{\pm 0.2}$ \\ 
\textbf{Llama-3-8b-chat} & 0 & 0.4 & 0.5 & 0.3 & 0.5 & 0.5 & 1.3 & 0.3 & 2.0 & 11.3 & $1.3_{\pm 0.3}$ & $1.0_{\pm 0.2}$ & $1.4_{\pm 0.5}$ & $0.1_{\pm 0.2}$ \\ 
\textbf{ChatGPT-3.5-turbo} & 0 & 0 & 0.1 & 0 & 0 & 0 & 0.1 & 0.2 & 4.6 & 6.1 & $0.6_{\pm 0.2}$ & $0.1_{\pm 0.1}$ & $0.9_{\pm 0.4}$ & $0_{\pm 0}$ \\   
\textbf{GPT-4o} & 0 & 0 & 0.3 & 0 & 0 & 0 & 0 & 0.1 & 0 & 0 & $0_{\pm 0}$ & $0_{\pm 0}$ & $0_{\pm 0}$ & $0_{\pm 0}$ \\                          
\hline
\hline
& \multicolumn{14}{c}{\textit{Zero-shot CoT}} \\
\textbf{Llama-2-7b-chat} & 1.0 & 3.7 & 6.0 & 4.9 & 1.0 & 2.8 & 16.8 & 4.0 & 34.4 & 6.8 & $6.4_{\pm 0.5}$ & $5.3_{\pm 0.4}$ & $5.6_{\pm 1.0}$ & $0.9_{\pm 0.5}$ \\ 
\textbf{Llama-2-13b-chat} & 1.1 & 3.6 & 1.1 & 5.0 & 4.1 & 3.0 & 8.1 & 8.4 & 15.0 & 7.5 & $3.4_{\pm 0.2}$ & $1.7_{\pm 0.5}$ & $2.4_{\pm 1.1}$ & $0_{\pm 0}$ \\    
\textbf{Vicuna-v1.5-7b} & 0.4 & 0 & 1.5 & 0 & 0.5 & 0.3 & 2.2 & 0.8 & 1.4 & 0.6 & $2.2_{\pm 0.4}$ & $0.7_{\pm 0.3}$ & $1.9_{\pm 0.8}$ & $0.1_{\pm 0.2}$ \\  
\textbf{Vicuna-v1.5-13b} & 0.2 & 0.1 & 0.7 & 0.1 & 0.7 & 1.0 & 0.6 & 2.1 & 5.0 & 4.1 & $1.3_{\pm 0.3}$ & $0.8_{\pm 0.2}$ & $0.9_{\pm 0.5}$ & $0_{\pm 0}$ \\     
\textbf{Llama-3-8b-chat} & 0.5 & 1.2 & 1.7 & 1.1 & 3.2 & 1.2 & 1.2 & 0.2 & 2.6 & 6.2 & $3.9_{\pm 0.6}$ & $0.9_{\pm 0.2}$ & $3.0_{\pm 0.9}$ & $0.2_{\pm 0.3}$ \\ 
\textbf{ChatGPT-3.5-turbo} & 0.1 & 0 & 0.7 & 0.1 & 0.2 & 0 & 0.2 & 0.3 & 5.8 & 5.5 & $0.9_{\pm 0.2}$ & $0_{\pm 0}$ & $0.4_{\pm 0.4}$ & $0_{\pm 0}$ \\       
\textbf{GPT-4o} & 0 & 0 & 1.3 & 0 & 0 & 0 & 0 & 0 & 0 & 0 & $0.1_{\pm 0.1}$ & $0_{\pm 0}$ & $0_{\pm 0}$ & $0_{\pm 0}$ \\                      
\hline
\hline
& \multicolumn{14}{c}{\textit{Few-shot}} \\
\textbf{Llama-2-7b-chat} & 0.2 & 0 & 0.4 & 0.1 & 0.6 & 0 & 16.4 & 0.2 & 20.5 & 0.2 & $1.1_{\pm 0.2}$ & $0.1_{\pm 0.1}$ & $2.9_{\pm 0.8}$ & $0_{\pm 0}$ \\ 
\textbf{Llama-2-13b-chat} & 0.2 & 0.1 & 0.2 & 0 & 0 & 0 & 5.8 & 0.1 & 7.1 & 0 & $0.3_{\pm 0.1}$ & $0.1_{\pm 0.1}$ & $0.6_{\pm 0}$ & $0_{\pm 0}$ \\  
\textbf{Vicuna-v1.5-7b} & 0.1 & 0 & 0.4 & 0 & 0.4 & 0 & 0.1 & 0 & 0.3 & 0 & $0.3_{\pm 0.2}$ & $0_{\pm 0}$ & $0.6_{\pm 0.5}$ & $0_{\pm 0}$ \\      
\textbf{Vicuna-v1.5-13b} & 0 & 0 & 0 & 0 & 0.3 & 0 & 0.2 & 0 & 0 & 0 & $0.3_{\pm 0.2}$ & $0.1_{\pm 0.1}$ & $0.2_{\pm 0.4}$ & $0_{\pm 0}$ \\ 
\textbf{Llama-3-8b-chat} & 0.1 & 0 & 0 & 0 & 0.9 & 0 & 0.9 & 0 & 0.5 & 0 & $1.0_{\pm 0.2}$ & $0.1_{\pm 0.1}$ & $1.6_{\pm 0.6}$ & $0_{\pm 0}$ \\ 
\textbf{ChatGPT-3.5-turbo} & 0 & 0 & 0 & 0 & 0 & 0 & 0.1 & 0 & 0.5 & 0 & $0.1_{\pm 0.1}$ & $0_{\pm 0}$ & $0.4_{\pm 0.3}$ & $0_{\pm 0}$ \\   
\textbf{GPT-4o} & 0.1 & 0 & 0 & 0 & 0 & 0 & 0 & 0 & 0 & 0 & $0_{\pm 0}$ & $0_{\pm 0}$ & $0.2_{\pm 0.3}$ & $0_{\pm 0}$ \\                  
\hline
\hline
& \multicolumn{14}{c}{\textit{Few-shot CoT}} \\
\textbf{Llama-2-7b-chat} & 0.3 & 0 & 0.7 & 0.1 & 0.6 & 0 & 13.8 & 0.2 & 14.9 & 0 & $1.1_{\pm 0.2}$ & $0.1_{\pm 0.1}$ & $3.0_{\pm 0.9}$ & $0_{\pm 0}$ \\ 
\textbf{Llama-2-13b-chat} & 0.2 & 0 & 0.1 & 0.3 & 0 & 0.2 & 4.3 & 0.1 & 5.8 & 0 & $0.3_{\pm 0.1}$ & $0_{\pm 0}$ & $0.6_{\pm 0.2}$ & $0_{\pm 0}$ \\    
\textbf{Vicuna-v1.5-7b} & 0 & 0 & 0.3 & 0 & 0.5 & 0 & 0.1 & 0 & 0 & 0 & $0.4_{\pm 0.2}$ & $0_{\pm 0}$ & $0.6_{\pm 0.6}$ & $0_{\pm 0}$ \\      
\textbf{Vicuna-v1.5-13b} & 0.1 & 0 & 0 & 0 & 0.3 & 0 & 0.2 & 0 & 0 & 0 & $0.2_{\pm 0.1}$ & $0.1_{\pm 0.1}$ & $0.2_{\pm 0.3}$ & $0_{\pm 0}$ \\ 
\textbf{Llama-3-8b-chat} & 0.5 & 0 & 0 & 0 & 1.1 & 0 & 0.9 & 0.1 & 0.5 & 0 & $2.6_{\pm 0.4}$ & $0.1_{\pm 0.1}$ & $0.9_{\pm 0.6}$ & $0_{\pm 0}$ \\ 
\textbf{ChatGPT-3.5-turbo} & 0.1 & 0 & 0 & 0 & 0 & 0 & 0.3 & 0 & 0 & 0 & $0.6_{\pm 0.3}$ & $0_{\pm 0}$ & $1.7_{\pm 0.9}$ & $0_{\pm 0}$ \\   
\textbf{GPT-4o} & 0 & 0 & 0 & 0 & 0 & 0 & 0 & 0 & 0 & 0 & $0_{\pm 0}$ & $0_{\pm 0}$ & $0_{\pm 0}$ & $0_{\pm 0}$ \\                      
\bottomrule
\end{tabular}

\caption{LLMs hallucination rate on the CWI 2018 and CompLex LCP 2021 datasets. S indicates the percentage of wrong sentences, and W indicates the percentage of wrong target words in the LLMs' output.}
\label{tab:llm_understand_err}
\end{table*}

\subsection{Generated Demonstrations}
\label{sec:llm_difficulty}

In the chain-of-thought settings, we prompt the model to provide a brief proof before generating the final label. The reasoning behind letting the model first produce a demonstration and then generate the answer is to enforce the model to ``think before answer'', i.e., the generated proof guides the model to produce a better solution based on some reasons. If we let the model answer and then provide proof, the proof would have been influenced by the initial answer, altering the model's internal bias. In Table \ref{tab:llm_proofs_llama2_13b_cwi} from Appendix~\ref{sec:figures}, we show some examples of proofs regarding the answer provided by Llama-2-13b-chat on the CWI English dataset in the zero-shot setting. The generated proofs motivate the answers, but we notice some flaws in the reasoning. For example, the model says that "ft" (i.e., feet as a unit of measurement) is standard in English. Meanwhile, it tends to contradict that being an abbreviation makes it difficult to understand. We notice this pattern quite often in the outputs. 

\subsection{Confusion Matrices on CWI}
\label{sec:cm_cwi}

\begin{figure*}[!t]
\centering
\graphicspath{{figures/distr/figs/}}
\begin{subfigure}{\textwidth}
    \includegraphics[width=\textwidth]{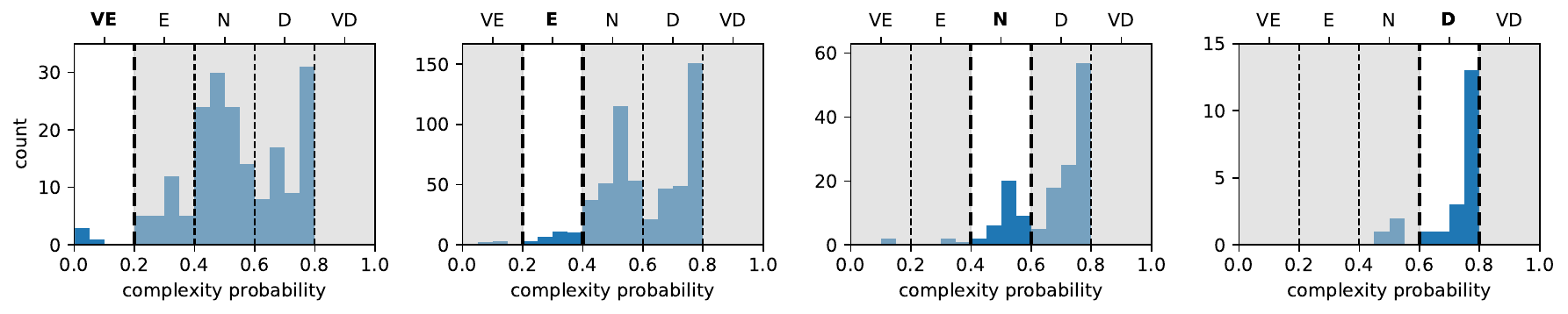}
    \caption{Zero-shot.}
\end{subfigure}
\hfill
\begin{subfigure}{\textwidth}
    \includegraphics[width=\textwidth]{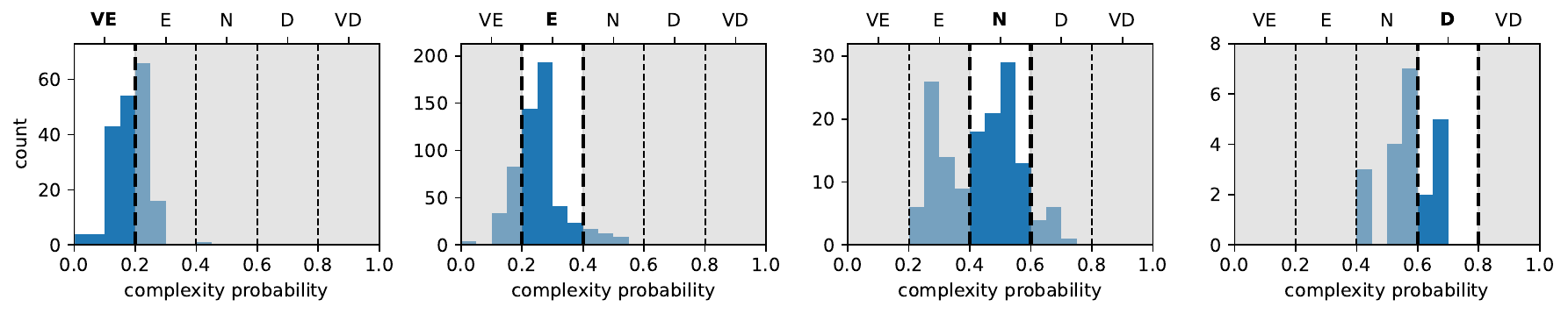}
    \caption{Fine-tune.}
\end{subfigure}
\caption{Predictive probability distribution of Llama 2 7B on LCP 2021 Single Word dataset, zero-shot, and fine-tuned settings. The gray area indicates the outside of the expected label region (i.e., wrong labels); the white stripe indicates the correctly predicted labels. Neither model predicts in the very difficult interval. Notation: VE - very easy, E - easy, N - neutral, D - difficult, VD - very difficult.}
\label{fig:lcp_distr_llama2_7b_single_extract}
\end{figure*}

\begin{figure*}[!h]
\centering
\includegraphics[width=0.9\textwidth]{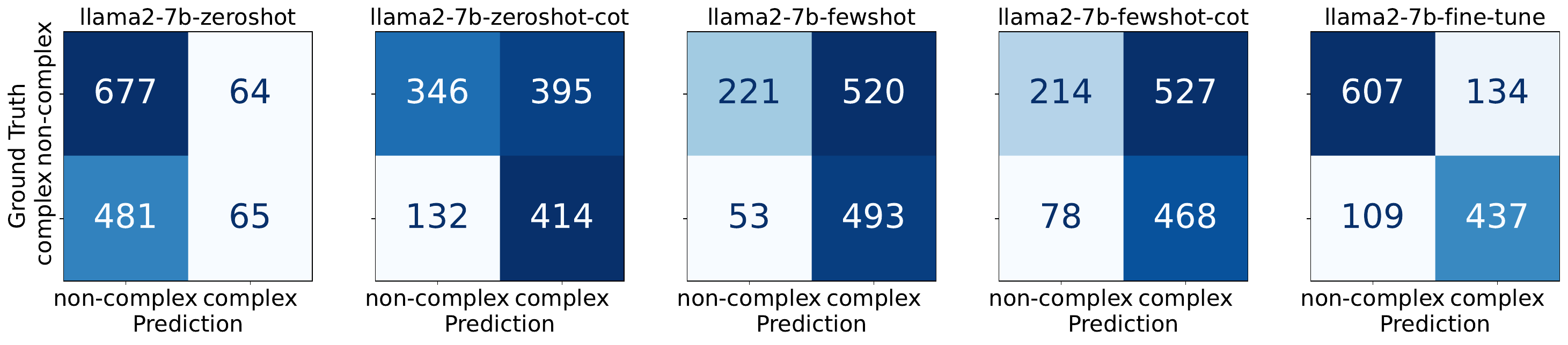}
\caption{Confusion matrices computed on the CWI 2018 English WikiNews test set for Llama 2 7B.}
\label{fig:cm_cwi_llama2-7b_wikinews}
\end{figure*}

We generate the confusion matrices to investigate how the predictions are affected by the domain, language, and model architecture. This section showcases the Llama 2 7B model on the CWI 2018 English WikiNews test set in Figure \ref{fig:cm_cwi_llama2-7b_wikinews}. More figures can be found in Appendix~\ref{sec:figures}. 

The general tendency is that chat models have higher false-positive or false-negative rates. The same model checkpoints have the same bias towards one false positive and false negative rate in the multidomain setting. For example, Llama-2-7b-chat has a high false-positive rate, while Llama-2-13b-chat has a high false rate. Correlated with the proofs generated by the LLMs, this is motivated by the fact that LLMs tend to produce either overestimates or underestimates of the word difficulty. It is especially true if the model finds a synonym for the target word. Also, the errors correlate with the model's degree of hallucination. On the other hand, fine-tuned models show lower false-positive/negative rates, meaning that this approach reduces the hallucination, and the model learns latent instructions directly from the data.

\subsection{Prediction Distribution on LCP}
\label{sec:predictions_ft}

We analyze the complexity probability distribution outputted by the LLMs and present in Figure~\ref{fig:lcp_distr_llama2_7b_single_extract}
the case for zero-shot and fine-tuned Llama 2 7B model. More figures across more settings, models, and datasets are shown in Appendix \ref{sec:figures}. The analysis is constructed by binning the models' real-valued estimates (i.e., the x-axis) and generating a histogram (i.e., the y-axis). The discrete labels were mapped equidistantly in the range 0-1, as such: very easy (VE) in 0-0.2, easy (E) in 0.2-0.4, neutral (N) in 0.4-0.6, difficult (D) in 0.6-0.8, and very difficult (VE) in 0.8-1.

We notice a more uniform distribution among models' predictions, especially for the low-complexity words. The absolute error is more than one step in the difficulty scale. We notice that the models struggle to identify the very difficult label, regardless of whether the model was fine-tuned or not. There is a tendency to label very easy words as easy and misclassify neutral and complex words, generally considering the words easier. This yields a smoother distribution across the labels. However, in the case of ChatGPT-3.5 and GPT-4o, the outputs tend to be more deterministic -- the majority of labels lie on the class scores.

\section{Conclusions}
In conclusion, we addressed CWI and LCP using LLMs, specifically Llama-based and OpenAI's GPT models. We observed that these models can determine the word difficulty level in multiple domains and languages, although with limited performances. Meanwhile, these models struggle to label very difficult phrases correctly. Future directions imply investigating multiple models in more languages. Also, the prompts and example selection greatly influence the models' performance. Thus, other future work should rely on reducing hallucination and determining which adversarial examples affect the model's capabilities the most.

\section{Limitations}
\label{sec:limitations}
Our approach has some limitations regarding prompt design. During experiments, we noticed that prompt design can highly influence the results, especially in the case of zero-shot settings. Using the same prompt across all models is not optimal, but we tried to find those instructions that benefit all models. Providing the model with specific instructions helps reduce the hallucination. One way to mitigate hallucinations was to use a specific structured output like JSON format (see Appendix~\ref{sec:prompting}), which required task validation through query reproduction.

Also, we know that random sampling is not the optimal solution for choosing fine-tuning examples for ChatGPT-3.5-turbo. The size and quality of data can significantly impact the prediction performance. To reduce this effect, we created a balanced dataset among label difficulties, such that the model is equally trained on easy and complex words. We also kept a uniform distribution among complexity probabilities strictly greater than zero for both tasks (CWI and LCP).

\section{Ethical Considerations}

Since we used pre-trained LLMs, all their limitations apply to our work. Developing CWI and LCP systems can benefit new language learners (e.g., chat-based applications in which LLMs help new language learners understand difficult words and even provide alternatives). However, because of the hallucinations and inaccuracies such models may provide, these systems can violate codes of ethics and harm or address attacks on such individuals. We are aware of the fast-paced development in the LLM area, and we think this area of research needs some attention. Therefore, we make the fine-tuned models publicly available for transparency and fair comparison with feature works\footnote{\url{https://github.com/razvanalex-phd/cwi_llm}}. These models should only be used for research. All the data we used is already publicly available, and the pre-trained Llama models are available on HuggingFace\footnote{\url{https://huggingface.co/meta-llama}}, under the Llama 2 License Agreement\footnote{\url{https://github.com/facebookresearch/llama/blob/main/LICENSE}}. We did not use the resources for other purposes than the ones allowed.

\bibliography{anthology,custom}

\clearpage
\appendix

\section{Prompts}
\label{sec:prompting}
\subsection{Inference Prompts}

\subsubsection{LCP English Prompt}
\begin{displayquote}
You are a helpful, honest, and respectful assistant for identifying the word complexity for beginner English learners. You are given one sentence in English and a phrase from that sentence. Your task is to evaluate the complexity of the word. Answer with one of the following: very easy, easy, neutral, difficult, very difficult. Be concise. Please, answer using the following JSON format: 
\end{displayquote}
\begin{spverbatim}
{
    "sentence": "the sentence you were provided",
    "word": "the word or words you have to analyze",
    "proof": "explain your response in maximum 50 words",
    "complex": "either very easy, easy, neutral, difficult, or very difficult"
}
\end{spverbatim}
\begin{displayquote}
What is the difficulty of \texttt{`\{token\}`} from \texttt{`\{sentence\}`}?
\end{displayquote}

\subsubsection{CWI English Prompt}
\begin{displayquote}
You are a helpful, honest, and respectful assistant for identifying the word complexity for beginner English learners. You are given one sentence in English and a phrase from that sentence. Your task is to say whether the phrase is complex. Assess the answer for the phrase, given the context from the sentence. Be concise. Please, use the following JSON schema:
\end{displayquote}
\begin{spverbatim}
{
    "sentence": "the sentence you were provided",
    "word": "the word or words you have to analyze",
    "proof": "explain your response in maximum 50 words",
    "complex": "either false (for simple) or true (for complex)",
}
\end{spverbatim}
\begin{displayquote}
Is \texttt{`\{token\}`} complex in \texttt{`\{sentence\}`}?
\end{displayquote}

\subsubsection{CWI German Prompt}
\begin{displayquote}
Sie sind ein hilfsbereiter, ehrlicher und respektvoller Assistent, um die Wortkomplexität für Anfänger im Deutschen zu identifizieren. Sie erhalten einen Satz auf Deutsch und eine Phrase aus diesem Satz. Ihre Aufgabe ist es zu sagen, ob die Phrase komplex ist. Bewerten Sie die Antwort für die Phrase, anhand des Kontexts aus dem Satz. Seien Sie kurz. Bitte verwenden Sie das folgende JSON-Schema:
\end{displayquote}
\begin{spverbatim}
{
    "sentence": "der Satz, den Sie erhalten haben",
    "word": "das Wort oder die Wörter, die Sie analysieren müssen",
    "proof": "erklären Sie Ihre Antwort in maximal 50 Wörtern",
    "complex": "entweder false (für einfach) oder true (für komplex)",
}
\end{spverbatim}
\begin{displayquote}
Ist \texttt{`\{token\}`} von \texttt{`\{sentence\}`} complex?
\end{displayquote}

\subsubsection{CWI Spanish Prompt}
\begin{displayquote}
Eres un asistente útil, honesto y respetuoso para identificar la complejidad de las palabras para los principiantes que aprenden español. Se te da una oración en español y una frase de esa oración. Tu tarea es decir si la frase es compleja. Evalúa la respuesta para la frase, dada el contexto de la oración. Sé conciso. Por favor, usa el siguiente esquema JSON:
\end{displayquote}
\begin{spverbatim}
{
    "sentence": "la oración que se te proporcionó",
    "word": "la palabra o palabras que tienes que analizar",
    "proof": "explica tu respuesta en máximo 50 palabras",
    "complex": "false (para simple) o true (para complejo)"
}
\end{spverbatim}
\begin{displayquote}
¿Es \texttt{`\{token\}`} complejo en \texttt{`\{sentence\}`}?
\end{displayquote}

\subsection{Fine-Tune Prompts}

\subsubsection{LCP English Prompt}
\begin{displayquote}
You are a helpful, honest, and respectful assistant for identifying the word difficulty for non-native English speakers. You are given one sentence in English and a word from that sentence. Your task is to evaluate the difficulty of the word. Answer only with one of the following: very easy, easy, neutral, difficult, very difficult.
\end{displayquote}
\begin{spverbatim}
sentence: `{sentence}`
word: `{token}`
\end{spverbatim}

\subsubsection{CWI English Prompt}
\begin{displayquote}
You are a helpful, honest, and respectful assistant for identifying the word complexity for non-native English speakers. You are given one sentence in English and a word from that sentence. Your task is to say whether a word is complex or not. Answer only with one of the following: yes or no.
\end{displayquote}
\begin{spverbatim}
sentence: `{sentence}`
word: `{token}`
\end{spverbatim}

\subsubsection{CWI German Prompt}
\begin{displayquote}
Du bist ein hilfsbereiter, ehrlicher und respektvoller Assistent für die Identifizierung der Wortkomplexität für nicht-deutsche Muttersprachler. Dir wird ein Satz auf Deutsch und ein Wort aus diesem Satz gegeben. Deine Aufgabe ist es zu sagen, ob ein Wort komplex ist oder nicht. Antworten nur mit einem der Folgenden: ja, nein.
\end{displayquote}
\begin{spverbatim}
Satz: `{sentence}`
Wort: `{token}`
\end{spverbatim}

\subsubsection{CWI Spanish Prompt}
\begin{displayquote}
Eres un asistente útil, honesto y respetuoso para identificar la complejidad de las palabras para hablantes no nativos de inglés. Se te da una oración en inglés y una palabra de esa oración. Tu tarea es decir si una palabra es compleja o no. Responde solo con una de las siguientes opciones: sí, no.
\end{displayquote}
\begin{spverbatim}
oracion: `{sentence}`
palabra: `{token}`
\end{spverbatim}

\section{Choice for Number of Inference Steps}
\label{sec:inference_steps_ap}

\begin{figure}[!h]
\graphicspath{{figures/bootstrap/figs/}}
\begin{subfigure}{\columnwidth}
    \includegraphics[width=\textwidth]{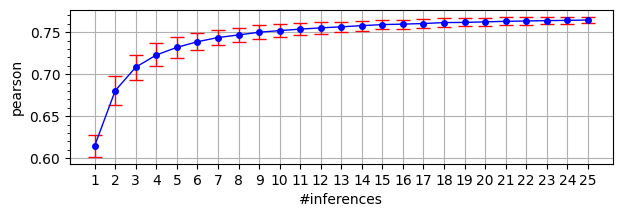}
    \caption{Pearson score on Llama-2-7b-ft}
\end{subfigure}
\hfill
\begin{subfigure}{\columnwidth}
    \includegraphics[width=\textwidth]{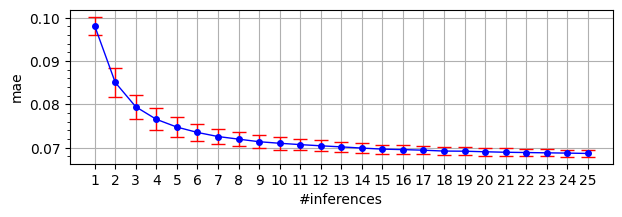}
    \caption{MAE score on Llama-2-7b-ft}
\end{subfigure}
\hfill
\begin{subfigure}{\columnwidth}
    \includegraphics[width=\textwidth]{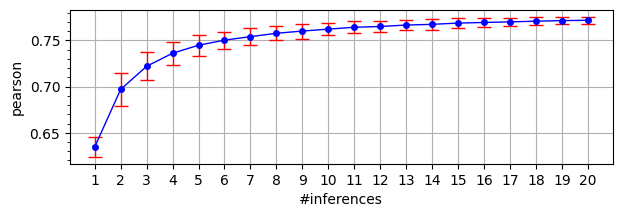}
    \caption{Pearson score on Llama-2-13b-ft}
\end{subfigure}
\hfill
\begin{subfigure}{\columnwidth}
    \includegraphics[width=\textwidth]{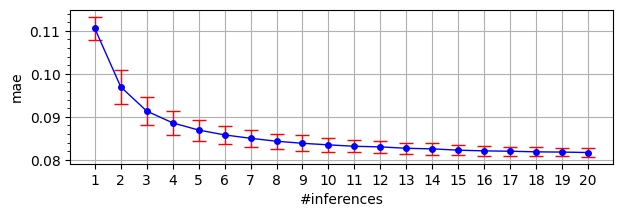}
    \caption{MAE score on Llama-2-13b-ft}
\end{subfigure}
\hfill
\begin{subfigure}{\columnwidth}
    \includegraphics[width=\textwidth]{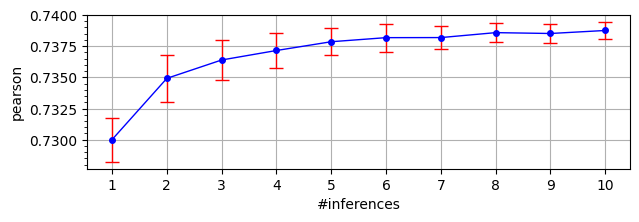}
    \caption{Pearson score on ChatGPT-3.5-turbo-ft}
\end{subfigure}
\hfill
\begin{subfigure}{\columnwidth}
    \includegraphics[width=\textwidth]{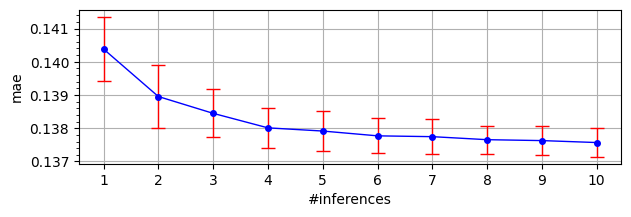}
    \caption{MAE score on ChatGPT-3.5-turbo-ft}
\end{subfigure}

\caption{Estimated Pearson and MAE scores against the number of LLM inference steps.}
\label{fig:inference_steps} 
\end{figure}

As presented in \S \ref{sec:method}, the estimated score in the LCP setting is an average of scores obtained after $K$ inference steps. We wanted to know what is the minimum number of inference steps required until the results do not change significantly anymore. Therefore, we set $K=25$ for Llama-2-7b-ft, $K=20$ for Llama-2-13b-ft, and $K=10$ for ChatGPT-3.5-turbo-ft, and then estimated the average score per number of iterations using bootstrapping, with 100 samples. The plots are shown in Figure~\ref{fig:inference_steps}. We obtained that at least 10 to 15 runs are required, after which the scores do not change significantly.

\begin{figure*}[!ht]
\centering
\includegraphics[width=\textwidth]{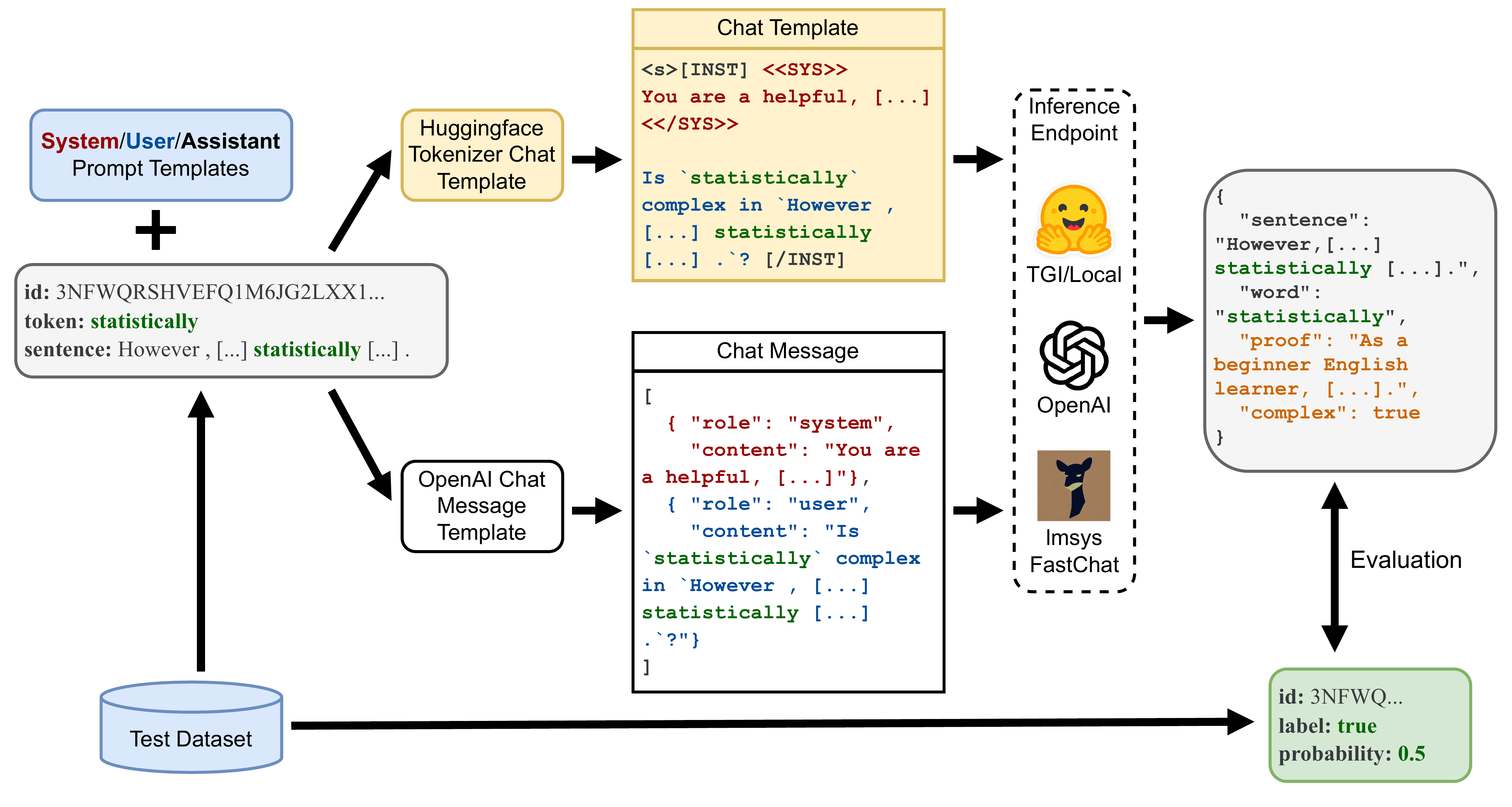}
\caption{Evaluation protocol.}
\label{fig:eval_proto}
\end{figure*}

\section{Evaluation Protocol}
\label{sec:eval_protocol}

To evaluate the LLMs effectively, we employ an approach that uses optimized inference servers and a generic way to interface with the models. The overall protocol is showcased in Figure~\ref{fig:eval_proto}. First, we load the dataset, and for every example, we apply the system, user, and assistant prompt templates. We use the assistant prompt template for the few-shot examples only in the few-shot setting. The final prompt is sent to the server, which processes the input and returns the LLM's prediction. We send the queries in parallel to use batching and other optimizations, thus reducing the execution time.  For local inference endpoints, we use HuggingFace Text Generation Inference (TGI)\footnote{\url{https://huggingface.co/docs/text-generation-inference/en/index}} for most LLMs available in HuggingFace and lmsys' FastChat\footnote{\url{https://github.com/lm-sys/FastChat}} with vllm~\cite{kwon2023efficient} integration for improved inference throughput. OpenAI's models are evaluated using their endpoints.  Ultimately, we aggregate and assess the results against the ground truth labels. We compute and report the metrics and perform analysis depending on the task.

During experiments, we noticed that some smaller LLMs struggle to output the response in the requested format, which makes it challenging to extract the prediction. To address this limitation, we employ guidance\footnote{\url{https://github.com/guidance-ai/guidance}} and outlines \cite{willard2023efficient}, which force the model to follow our custom output structure by manipulating the output logits.

\section{Computational Costs and Hardware Infrastructure}
\label{appendix:costs}

Due to the number of experiments, available resources, and costs, we used hardware from multiple sources:

\begin{itemize}
    \item A desktop PC with an Intel(R) Core i7-13700k CPU, 64GB RAM, 3TB NVMe SSD storage, and an Nvidia RTX 4080 16GB GPU.
    \item The cluster shared inside our organization through the SLURM cluster management\footnote{\url{https://slurm.schedmd.com/}}. The experiments were run on systems with Intel(R) Xeon(R) Gold 6326 CPU, 500GB RAM, NVIDIA A100-PCIE 40GB, and 250TB NFS storage.
    \item GPUs rented from the vast.ai\footnote{\url{https://vast.ai}} platform using NVIDIA RTX 4090 (approx. \$0.5/hr), A6000 (approx. \$0.9/hr), or H100 80GB-SXM (approx. \$3.2/hr), and the minimum required storage was 200GB.
\end{itemize}

We trained and ran inferences on NVIDIA RTX 4080 and 4090 (consumer-class GPUs) and NVIDIA RTX A6000, A100 40GB-PCIe, and H100 80GB-SXM (server-class GPUs), depending on the minimal requirements to run the model and execution time. The desktop PC could run most inference experiments regarding zero-shot settings. However, due to the limited VRAM of the RTX 4080, few-shot and fine-tuning experiments required at least 40GB of GPU VRAM because more input/output tokens result in higher video memory requirements. We rented H100 to speed up the experiments, with the observed speed-up of 2-6 times faster than the A100 40GB GPU. We decided to employ A6000 and A100 GPUs for training and inference in the meta-learning setting. The estimated costs for these experiments rise to about \$400.

For OpenAI's API, we used inference endpoints for ChatGPT-3.5-turbo and GPT-4o as well as training endpoints for ChatGPT-3.5-turbo, with the pricing at the time of writing this paper: \$0.0005 per 1k input tokens and \$0.0015 per 1k output tokens for ChatGPT-3.5-turbo; and \$0.0080 per training tokens, \$0.003 per 1k input tokens, and \$0.006 per 1k output tokens. For GPT-4o, we had access only to inference, with \$5 per 1M  input tokens and \$15 per 1M output tokens. All experiments related to OpenAI's models totaled about \$300. Because of the high costs, we limited our experiments to only classification on large test sets.

\section{Model Checkpoints}
\label{app:checkpoints}

In Table \ref{tab:model_checkpoints}, we present the checkpoints used in this work, which are available on the Huggingface platform. We indicate with ``ft'' where we use a different checkpoint for fine-tuning. Fine-tuned models are available at \url{https://github.com/razvanalex-phd/cwi_llm}.

\begin{table}[!h]
\centering
\small
\begin{tabular}{lp{4cm}}
\toprule
\textbf{Model} & \textbf{Checkpoint} \\
\midrule
\hline
Llama 2 7B & meta-llama/Llama-2-7b-chat-hf \\
Llama 2 7B ft & meta-llama/Llama-2-7b-hf \\
Llama 2 13B & meta-llama/Llama-2-13b-chat-hf \\
Llama 2 13B ft & meta-llama/Llama-2-13b-hf \\
Vicuna 1.5 7B & lmsys/vicuna-7b-v1.5 \\
Vicuna 1.5 7B ft & TheBloke/vicuna-7B-v1.5-AWQ \\
Vicuna 1.5 13B & lmsys/vicuna-13b-v1.5 \\
Vicuna 1.5 13B ft & TheBloke/vicuna-13B-v1.5-AWQ \\
Llama 3 8B & meta-llama/Meta-Llama-3-8B-Instruct \\
Llama 3 8B ft & meta-llama/Meta-Llama-3-8B \\
ChatGPT-3.5-turbo & gpt-3.5-turbo-0125 \\
GPT-4o & gpt-4o-2024-05-13 \\
\bottomrule
\end{tabular}
\caption{Checkpoints used during experiments.}
\label{tab:model_checkpoints}
\end{table}

\section{Few-Shot Examples and Proofs}

The few-shot examples for CWI datasets (see Tables \ref{tab:cwi-fewshot-proofs-en-news}, \ref{tab:cwi-fewshot-proofs-en-wikinews}, \ref{tab:cwi-fewshot-proofs-en-wikipedia}, \ref{tab:cwi-fewshot-proofs-de}, and \ref{tab:cwi-fewshot-proofs-es}) were chosen such that we provide two samples for false complexity (probability is 0\%) and one sample for every discrete label as presented in \S \ref{sec:cwi_problem_llm_era}. This way, even if we are biased towards complex sentences, the distribution among discrete labels is uniform. The sentences were randomly chosen so that they fulfilled the previous criteria. We use GPT-4o to generate proofs for any given sentence and token regarding why the label is correct. Similarly, we also do this for LCP datasets (see Tables \ref{tab:lcp-fewshot-proofs-single}, \ref{tab:lcp-fewshot-proofs-multi}) by sampling an example from every category. In both tasks, the few-shot samples were selected from the train set. When generating the prompt, we shuffle the few-shot examples to reduce the chances of a position bias; that is, the model would predict a specific label due to how the few-shot samples are ordered. Examples of generated outputs for Llama 2 13B on the CWI English Wikipedia test set are presented in Table \ref{tab:llm_proofs_llama2_13b_cwi}.

\begin{table*}[!ht]
\centering
\small
\begin{tabular}{p{0.7cm}p{5cm}p{1.5cm}p{1.6cm}p{5cm}}
\toprule
\textbf{ID} & \textbf{Sentence} & \textbf{Token} &\textbf{Complex} & \textbf{Proof} \\
\midrule
\hline
7329 & Northern Ireland's deputy first minister and Mid-Ulster MP, Martin McGuinness, said his heart went out to the family of the girl tragically killed. & MP & false & The abbreviation 'MP' for Member of Parliament is commonly understood in contexts involving government or politics, making it familiar and not complex. \\
\hline
7700 & Activists said at least 30 people died on Sunday - mainly civilians - as violence surged at flashpoints across the country despite an increase of UN observers. & civilians & true & The word 'civilians' may be considered complex because it is specific to contexts involving military or emergency situations, and not everyone might be familiar with its precise meaning. \\
\hline
407 & The regime is trying to punish these villages and to put an end to this revolution as quickly as it can, he said. & punish & false & The word 'punish' is a common verb used in everyday language to describe the act of imposing a penalty or suffering for an offense, making it familiar and not complex. \\
\hline
5182 & The Philippines and Vietnam complained last year of increasingly aggressive acts by China in staking its claim to the South China Sea. & aggressive acts & true & The phrase 'aggressive acts' may be considered complex as it involves understanding both 'aggressive' and 'acts' together, implying a specific type of behavior which might not be immediately clear without context. \\
\hline
12472 & Goodyear said police previously responded to the Florida 'Space Coast' home, about 15 miles south of Cape Canaveral, for domestic disturbance calls involving Jaxs Johnson. & domestic disturbance & true & The term 'domestic disturbance' can be complex as it combines 'domestic', related to the home or family, with 'disturbance', indicating trouble or conflict, requiring an understanding of both terms in context. \\
\hline
7131 & Spain is set to intensify the clean-up of its banks on Friday after difficult last-minute talks between the government and lenders on details of planned financial system reforms. & Friday & false & The word 'Friday' is a basic term indicating a day of the week, universally understood and not complex. \\
\hline
10459 & The country's leaders have to admit that there were numerous falsifications and rigging and the results do not reflect the will of the people, Gorbachev told Interfax, according to the AFP. & rigging & true & The word 'rigging' can be considered complex as it refers to the act of manipulating or tampering with something, often in a fraudulent way, which may not be a familiar concept to everyone. \\
\bottomrule
\end{tabular}
\caption{Few shot examples for CWI 2018 English - news domain.}
\label{tab:cwi-fewshot-proofs-en-news}
\end{table*}

\begin{table*}[!ht]
\centering
\small
\begin{tabular}{p{0.7cm}p{5cm}p{1.5cm}p{1.6cm}p{5cm}}
\toprule
\textbf{ID} & \textbf{Sentence} & \textbf{Token} &\textbf{Complex} & \textbf{Proof} \\
\midrule
\hline
4055 & \#29-17 He joins 139 other Republican Party presidential candidates who have done likewise. & Party & false & The word 'Party' is common and widely understood in political contexts, making it familiar to both native and non-native speakers. \\
\hline
5461 & \#11-14 The experiments were funded by national research organizations in the United States and China and the government of Brazil. & national & false & The word 'national' is a basic adjective used to describe something related to a nation, and is commonly used in many contexts, making it easy for most speakers. \\
\hline
4911 & \#42-4 The team used Formica fusca, an ant species that can form thousand-strong colonies. & Formica fusca & true & The term 'Formica fusca' is a scientific name for a specific ant species, which is likely unfamiliar to most people outside of entomology or biological sciences. \\
\hline
3758 & \#22-5 According to doctors at Bethany Hospital, Kalam was dead by 7 p.m. but they waited for the arrival of Meghalaya chief minister V. Shanmuganathan, about an hour later, before announcing the death. & announcing & true & The word 'announcing' can be challenging due to its length, the presence of a silent letter, and the necessity to understand the appropriate context for its use. \\
\hline
2220 & \#36-16 Another had been to tether the nose cone to the car; Hunter-Reay mentioned renderings developed of a boomerang-like debris-deflector positioned in front of the driver. & tether & true & The word 'tether' is less commonly used and may not be familiar to many people, leading to difficulty in understanding its meaning and usage. \\
\hline
1951 & \#24-37 Furthermore, the data of radars at Maldives airports have also been analysed and shows no indication of the said flight", said Malaysian Transport Minister Hishamuddin Hussein. & analysed & true & The word 'analysed' can be difficult due to its British English spelling (with 's' instead of 'z'), which might confuse those more familiar with American English. \\
\hline
1498 & \#3-10 Pavlensky and Oksana were detained in December at Sheremetyevo airport for questioning, which went on for seven hours. & detained & true & The word 'detained' may be difficult due to its legal context and the less frequent use in everyday language, requiring a higher level of vocabulary knowledge. \\
\bottomrule
\end{tabular}
\caption{Few shot examples for CWI 2018 English - WikiNews domain.}
\label{tab:cwi-fewshot-proofs-en-wikinews}
\end{table*}

\begin{table*}[!ht]
\centering
\small
\begin{tabular}{p{0.7cm}p{5cm}p{1.5cm}p{1.6cm}p{5cm}}
\toprule
\textbf{ID} & \textbf{Sentence} & \textbf{Token} &\textbf{Complex} & \textbf{Proof} \\
\midrule
\hline
3595 & Once the series had received the backing of the FIA , a management structure including new executive directors Brian Menell and Tony Teixeira were appointed to oversee the sale of franchises for the operation of international teams . & Brian & false & The word 'Brian' is a common proper noun and a typical English name, which is familiar to both native and non-native speakers. Its presence in the sentence is straightforward and does not add complexity. \\
\hline
3400 & The first recorded case of an actor performing took place in 534 BC ( though the changes in calendar over the years make it hard to determine exactly ) when the Greek performer Thespis stepped on to the stage at the Theatre Dionysus and became the first known person to speak words as a character in a play or story . & play & false & The word 'play' is a basic English word frequently used in both its noun and verb forms. It is easily understood by both native and non-native speakers, especially in the context of theater. \\
\hline
1048 & Also , if the reviewing administrator concludes that the block was justified , you will not be unblocked unless the reviewing administrator is convinced that you understand what you are blocked for , and that you will not do it again . & administrator & true & The word 'administrator' is long and contains multiple syllables, which can make it challenging to pronounce and remember. Additionally, its specific meaning in the context of authority or management may not be immediately clear to non-native speakers. \\
\hline
3670 & Two is the base of the simplest numeral system in which natural numbers can be written concisely , being the length of the number a logarithm of the value of the number ( whereas in base 1 the length of the number is the value of the number itself ) ; the binary system is used in computers . & numeral & true & The word 'numeral' is less commonly used in everyday language and pertains to a specific field (mathematics). This specialization can make it less familiar and harder to understand for some readers. \\
\hline
2767 & The Angara rocket family is a family of space-launch vehicles being developed by the Moscow-based Khrunichev State Research and Production Space Center . & space-launch & true & The term 'space-launch' is a compound word that refers to a specific and technical concept related to aerospace. Its specialized nature and the combination of two words can make it more difficult to understand. \\
\hline
1155 & Early references from the Vadstena Abbey show how the Swedish nuns were baking gingerbread to ease indigestion in 1444 . & indigestion & true & The word 'indigestion' is relatively long and describes a specific medical condition related to digestion, which might not be commonly known or used in daily conversation, making it harder for some readers. \\
\hline
919 & The roof of the nave is composed of a pair and knuckle frame , coated inside by pieces of tracery . & tracery & true & The word 'tracery' is an architectural term that may not be widely recognized outside of specialized contexts. Its specific meaning and less frequent use contribute to its complexity. \\
\bottomrule
\end{tabular}
\caption{Few shot examples for CWI 2018 English - Wikipedia domain.}
\label{tab:cwi-fewshot-proofs-en-wikipedia}
\end{table*}

\begin{table*}[!ht]
\centering
\small
\begin{tabular}{p{0.6cm}p{3.2cm}p{1.4cm}p{1.0cm}p{3.2cm}p{3.2cm}}
\toprule
\textbf{ID} & \textbf{Sentence} & \textbf{Token} &\textbf{Complex} & \textbf{Proof} & \textbf{Proof (En.)}  \\
\midrule
\hline
4890 & Unmittelbar nach den Anschlägen vom 11. & Unmittelbar & false & Das Wort 'Unmittelbar' ist nicht komplex, da es ein häufig verwendetes deutsches Adjektiv ist und weder selten noch schwierig zu verstehen ist. & The word 'Immediate' is not complex as it is a commonly used German adjective and is neither rare nor difficult to understand. \\
\hline
713 & Janukowytsch findet dort die größte Unterstützung , während Juschtschenko das größte Wählerpotenzial sieht . & größte & false & Das Wort 'größte' ist ein Basisadjektiv in der deutschen Sprache und stellt keine besondere Schwierigkeit dar. & The word 'largest' is a basic adjective in the German language and does not pose any particular difficulty. \\
\hline
4106 & Sie berichtete unter anderem über ihre derzeitige Tournee mit dem Thema Hitler-Tagebücher . & Tournee & true & Das Wort 'Tournee' stammt aus dem Französischen und wird in der deutschen Sprache seltener verwendet, was es für Nicht-Muttersprachler schwieriger macht. & The word 'tournee' comes from French and is used less frequently in the German language, making it more difficult for non-native speakers. \\
\hline
2738 & Die Anwälte Berlusconis kündigten an , gegen die Verjährung einen Einspruch einzureichen , um einen Freispruch erster Klasse zu erreichen . & Freispruch & true & Das Wort 'Freispruch' kann komplex sein, da es ein spezifischer juristischer Begriff ist, der in alltäglichen Gesprächen selten vorkommt. & The word 'acquittal' can be complex because it is a specific legal term that rarely appears in everyday conversations. \\
\hline
3535 & Der eineinhalbstündige feierliche Trauergottesdienst fand in der zu zwei Drittel gefüllten Friedenskirche im Nürnberger Stadtteil St. -Johannis statt . & Trauer-gottesdienst & true & Das Wort 'Trauergottesdienst' ist komplex, da es ein zusammengesetztes Substantiv ist und selten verwendet wird. & The word 'funeral service' is complex because it is a compound noun and is rarely used. \\
\hline
185 & Konvergenz als Ursache der Fehleinordnung : Nach ihrer Analyse des Fibrinogen-Gens stellen etwa die äußerlich sehr ähnlichen Flamingos und Löffler zwei weit auseinanderliegende Gruppen auf den beiden Evolutionsästen dar . & Fibrinogen-Gens & true & Das Wort 'Fibrinogen-Gens' ist komplex, da es ein wissenschaftlicher Begriff ist, der in der allgemeinen Sprache nicht häufig vorkommt. & The word 'fibrinogen gene' is complex because it is a scientific term that is not commonly used in common language. \\
\hline
5726 & Hauptgrund für die Verschlechterung des Zustandes sei der heiße und trockene Sommer 2003 mit hohen Ozonwerten . & Ozonwerten & true & Das Wort 'Ozonwerten' kann für Nicht-Muttersprachler schwierig sein, da es ein wissenschaftlicher Begriff ist und spezifisches Wissen über Luftqualität erfordert. & The word 'ozone levels' can be difficult for non-native speakers as it is a scientific term and requires specific knowledge of air quality. \\
\bottomrule
\end{tabular}
\caption{Few shot examples for CWI 2018 German. For proofs, we also provide the translation in English.}
\label{tab:cwi-fewshot-proofs-de}
\end{table*}

\begin{table*}[!ht]
\centering
\small
\begin{tabular}{p{0.6cm}p{3.2cm}p{1.4cm}p{1.0cm}p{3.2cm}p{3.2cm}}
\toprule
\textbf{ID} & \textbf{Sentence} & \textbf{Token} &\textbf{Complex} & \textbf{Proof} & \textbf{Proof (En.)}  \\
\midrule
\hline
11798 & En 1911, escapó de su casa y se alistó en una expedición militar, organizada por Ricciotti Garibaldi, para liberar a Albania del control turco. & Garibaldi & false & El apellido 'Garibaldi' no es difícil porque es un nombre propio conocido, especialmente en el contexto de la historia y la cultura italiana. & The surname 'Garibaldi' is not difficult because it is a well-known proper name, especially in the context of Italian history and culture. \\
\hline
10963 & Estos magos fueron, según la tradición, adorar al Mesías que acababa de nacer en Belén de Judea, el que posteriormente se llamaría Jesús de Nazaret. & adorar & true & La palabra 'adorar' puede considerarse difícil debido a su uso menos común y su connotación religiosa específica. & The word 'worship' may be considered difficult due to its less common use and its specific religious connotation. \\
\hline
8294 & En marzo de 2011 firma con el BK Jimki dónde sustituirá a Meleschenko, entrenador interino desde la renuncia de Sergio Scariolo tras no conseguir el pase para el Top-16 de la Euroliga. & interino & true & La palabra 'interino' puede ser difícil debido a su uso en un contexto específico y profesional, lo que requiere un conocimiento preciso del término. & The word 'interim' can be difficult due to its use in a specific and professional context, which requires precise knowledge of the term. \\
\hline
6171 & Linda con las poblaciones de Yepes, Huerta de Valdecarábanos y el término segregado de La Guardia, todas de Toledo. & Linda & true & La palabra 'Linda' es difícil porque se trata de un término geográfico específico que puede no ser conocido por todos los hablantes. & The word 'Linda' is difficult because it is a specific geographical term that may not be known to all speakers. \\
\hline
5911 & Estuvieron presentes el presidente de Estados Unidos Bill Clinton y el presidente de la República de Corea Kim Young Sam, y se dedicó a los hombres y mujeres que sirvieron en la guerra. & Bill & false & El nombre 'Bill' no es difícil porque es un nombre propio común y fácil de reconocer, especialmente en el contexto de figuras públicas como Bill Clinton. & The name 'Bill' is not difficult because it is a common and easy to recognize proper name, especially in the context of public figures like Bill Clinton. \\
\hline
2673 & Cada uno de los vectores columna de la matriz "A" se llama modo propio de vibración, y los "Ci" son las amplitudes relativas de cada modo propio. & amplitudes & true & La palabra 'amplitudes' es técnica y específica del campo de las matemáticas y la física, lo que puede hacerla difícil para quienes no están familiarizados con estos temas. & The word 'amplitudes' is technical and specific to the field of mathematics and physics, which can make it difficult for those unfamiliar with these topics. \\
\hline
1945 & El Ducado de Prusia o Prusia Ducal (en alemán: "Herzogtum Preußen"; en polaco: "Prusy Książęce") fue un ducado entre 1525-1701 en la región más oriental de Prusia heredero del Estado monástico de los Caballeros Teutónicos. & monástico & true & La palabra 'monástico' es difícil porque es un término especializado que se refiere a la vida y organización de los monasterios, lo que puede no ser familiar para todos. & The word 'monastic' is difficult because it is a specialized term referring to the life and organization of monasteries, which may not be familiar to everyone. \\
\bottomrule
\end{tabular}
\caption{Few shot examples for CWI 2018 Spanish. For proofs, we also provide the translation in English.}
\label{tab:cwi-fewshot-proofs-es}
\end{table*}

\begin{table*}[!ht]
\centering
\small
\begin{tabular}{p{0.7cm}p{5cm}p{1.5cm}p{1.6cm}p{5cm}}
\toprule
\textbf{ID} & \textbf{Sentence} & \textbf{Token} &\textbf{Complexity} & \textbf{Proof} \\
\midrule
\hline
6043 & Containers lost at sea and compensation (debate) & Containers & Very Easy & The word 'Containers' is a common and easily understood term in English, referring to objects used for holding or transporting items. \\
\hline
4290 & We have also shown that chondrogenesis can be initiated and chondrogenic differentiation will take place even in the absence of both BMP2 and BMP4 or BMP2 and BMP7. & differentiation & Easy & The word 'differentiation' is slightly technical and commonly used in biological contexts, making it easy but not very easy. \\
\hline
2143 & Their scribes and the Pharisees murmured against his disciples, saying, "Why do you eat and drink with the tax collectors and sinners?" & scribes & Neutral & The term 'scribes' is not commonly used in everyday language and refers to a specific historical role, requiring some background knowledge to understand. \\
\hline
5144 & Our data suggest that while recombination events destined to be resolved as COs can proceed normally in Trip13 mutants, DSBs that enter the NCO repair pathway are incompletely resolved or processed inefficiently. & COs & Difficult & The acronym 'COs' is specialized and requires specific knowledge in genetics to understand that it refers to 'crossovers' in the context of recombination events. \\
\hline
4873 & In the mouse model of RA, small genetic contributions are also often observed. & RA & Very Difficult & The acronym 'RA' stands for 'rheumatoid arthritis,' a term that is highly specialized and not immediately clear without specific medical knowledge. \\
\bottomrule
\end{tabular}
\caption{Few shot examples for LCP 2021 single-word expressions.}
\label{tab:lcp-fewshot-proofs-single}
\end{table*}

\begin{table*}[!ht]
\centering
\small
\begin{tabular}{p{0.7cm}p{5cm}p{1.5cm}p{1.6cm}p{5cm}}
\toprule
\textbf{ID} & \textbf{Sentence} & \textbf{Token} &\textbf{Complexity} & \textbf{Proof} \\
\midrule
\hline
526 & Therefore, TGF$\beta$ and BMP signaling are playing distinct but necessary roles to maintain articular cartilage. & necessary roles & Very Easy & The phrase 'necessary roles' is straightforward, commonly used in English, and easily understood within the context of the sentence. \\
\hline
212 & In this confidence, I was determined to come first to you, that you might have a second benefit; & second benefit & Easy & The phrase 'second benefit' is relatively simple, but the context may slightly challenge the reader, making it less immediate to understand. \\
\hline
1376 & We will be very strict on enforcing this fundamental principle in this case as well. & fundamental principle & Neutral & The term 'fundamental principle' requires a moderate understanding of abstract concepts and formal language, making it neutral in difficulty. \\
\hline
503 & neither to pay attention to myths and endless genealogies, which cause disputes, rather than God's stewardship, which is in faith-- & endless genealogies & Difficult & The phrase 'endless genealogies' is less common and refers to complex and potentially obscure biblical or historical references, adding to its difficulty. \\
\hline
1008 & Such polymorphisms should yield biomarkers suitable for more readily accessible samples, such as peripheral blood or buccal smears. & buccal smears & Very Difficult & The term 'buccal smears' is highly specialized and technical, typically known only to those with specific biomedical knowledge, making it very difficult. \\
\bottomrule
\end{tabular}
\caption{Few shot examples for LCP 2021 multi-word expressions.}
\label{tab:lcp-fewshot-proofs-multi}
\end{table*}

\section{Meta-Learning Datasets}
\label{appendix:metalearning}

We select 45 tasks from the BIG-bench benchmark, all being classification tasks. Some tasks offer the choices in the original prompt, whereas others do not. For the ones that don't, we manually append them in the prompt. The tasks can be viewed in Table \ref{tab:bigbench}.

\begin{table*}[!h]
\small
\centering
\begin{tabular}{lll}
\toprule
\multicolumn{3}{c}{\textbf{Tasks}} \\ \midrule
\hline
abstract\_narrative\_understanding & fantasy\_reasoning & nonsense\_words\_grammar \\
analytic\_entailment & figure\_of\_speech\_detection & odd\_one\_out \\
bbq\_lite\_json & formal\_fallacies\_syllogisms\_negation & penguins\_in\_a\_table \\
causal\_judgment & general\_knowledge & phrase\_relatedness \\
cause\_and\_effect & human\_organs\_senses & play\_dialog\_same\_or\_different \\
codenames & hyperbaton & presuppositions\_as\_nli \\
contextual\_parametric\_knowledge\_conflicts & implicatures & question\_selection \\
crash\_blossom & implicit\_relations & reasoning\_about\_colored\_objects \\
crass\_ai & intent\_recognition & riddle\_sense \\
dark\_humor\_detection & irony\_identification & ruin\_names \\
disambiguation\_qa & logical\_deduction & strange\_stories \\
empirical\_judgments & logical\_fallacy\_detection & temporal\_sequences \\
entailed\_polarity & metaphor\_boolean & timedial \\
epistemic\_reasoning & metaphor\_understanding & tracking\_shuffled\_objects \\
evaluating\_information\_essentiality & movie\_dialog\_same\_or\_different & winowhy \\
\bottomrule
\end{tabular}
\caption{All tasks selected from the BIG-bench benchmark that were used during the meta-learning process.}
\label{tab:bigbench}
\end{table*}

\section{Hyperparameters}

\textbf{Inference.} During inference, we set the LLM to use a maximum of 4,096 tokens, the repetition penalty was set to 1.2, and the temperature to 0.8. We set the top-k parameter to 10 and the top-p to 0.95. The open-source models were loaded with quantized parameters using the nf4 format through \texttt{bitsandbytes} \cite{CANNIZZO201837}. For LCP, we set the number of inference steps $K=20$ for all open-source models, while for OpenAI models, we evaluate on $K=10$ inferences (see also the discussion from Appendix~\ref{sec:inference_steps_ap}).

\textbf{Fine-tuning open-source models.} The open-source models are downloaded from HuggingFace. We employ QLoRA~\cite{dettmers2023qlora} with 4-bit quantization to reduce GPU memory usage for fine-tuning. We set $R$ to 16, $\alpha$ to 32, and dropout to $0.05$. The batch size varies between 10 and 32, and the learning rate uses a linear scheduler with a 10\% warmup and a maximum value of $1e-4$. The LLM is limited to handling a maximum of 1,024 tokens. We trained the models for three epochs using the AdamW optimizer \cite{DBLP:journals/corr/KingmaB14}.

\textbf{Fine-tuning GPT models.} We use OpenAI's platform with the default hyperparameters to fine-tune OpenAI models. We limit the training size to 250 samples uniformly sampled among labels from the train set specific to the dataset task and language.

\textbf{Meta-Learning.} For meta-learning, the inner learning rate is 0.1 and 0.03 for prompt learning and P-tuning, respectively, while the outer learning rates are 0.01 and 0.003. The batch size is set to 1, with the number of inner steps set to 5. We run each experiment for a total of 3,000 steps. During P-tuning, we also choose the LSTM \cite{hochreiter1997long} model as our architecture. We use 16 virtual tokens, and the support and query sets contain six examples from the same task each.

\section{Results Discussions}
\label{sec:figures}

\begin{figure*}[!ht]
\centering
\graphicspath{{figures/cm/figs/}}
\begin{subfigure}{\textwidth}
    \centering
    \includegraphics[width=\textwidth]{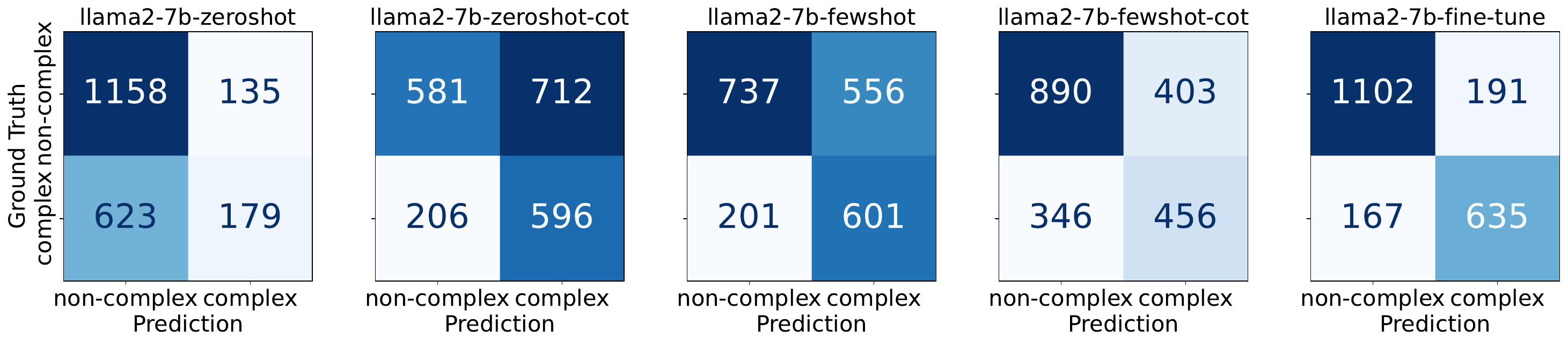}
    \caption{CWI 2018 English News.}
\end{subfigure}
\hfill
\begin{subfigure}{\textwidth}
    \centering
    \includegraphics[width=\textwidth]{cm_cwi-2018-en-wikinews_cwi-llama2-7b}
    \caption{CWI 2018 English WikiNews.}
\end{subfigure}
\hfill
\begin{subfigure}{\textwidth}
    \centering
    \includegraphics[width=\textwidth]{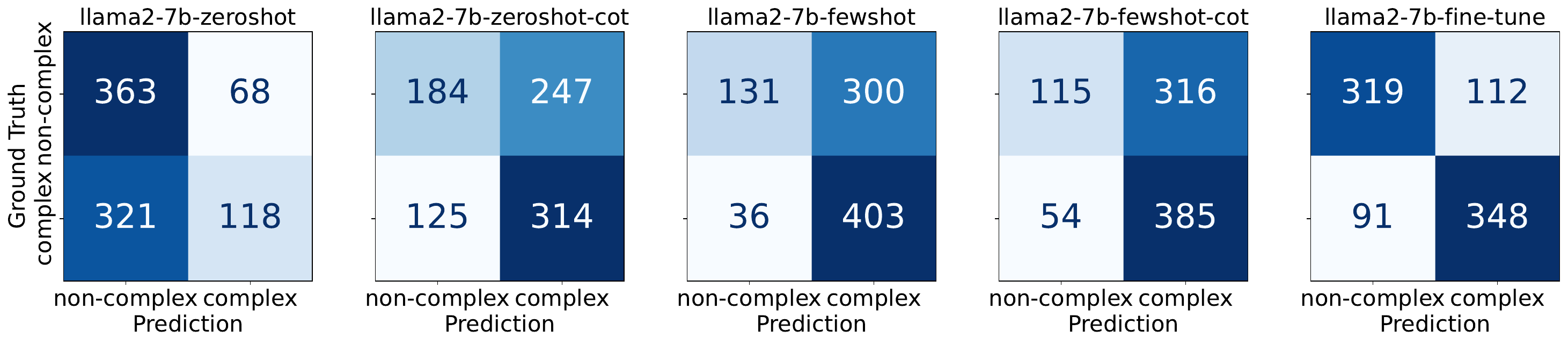}
    \caption{CWI 2018 English Wikipedia.}
\end{subfigure}
\hfill
\begin{subfigure}{\textwidth}
    \centering
    \includegraphics[width=\textwidth]{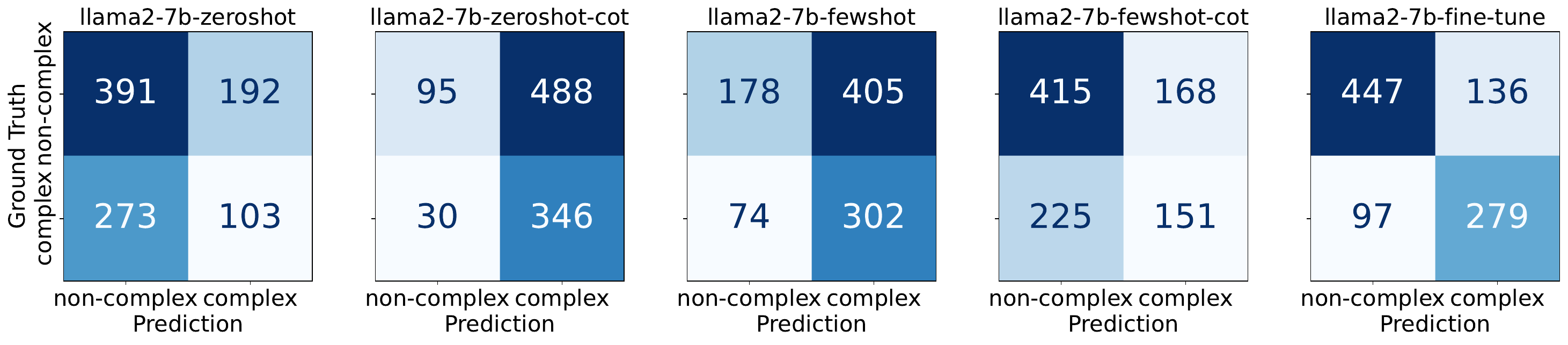}
    \caption{CWI 2018 German.}
\end{subfigure}
\hfill
\begin{subfigure}{\textwidth}
    \centering
    \includegraphics[width=\textwidth]{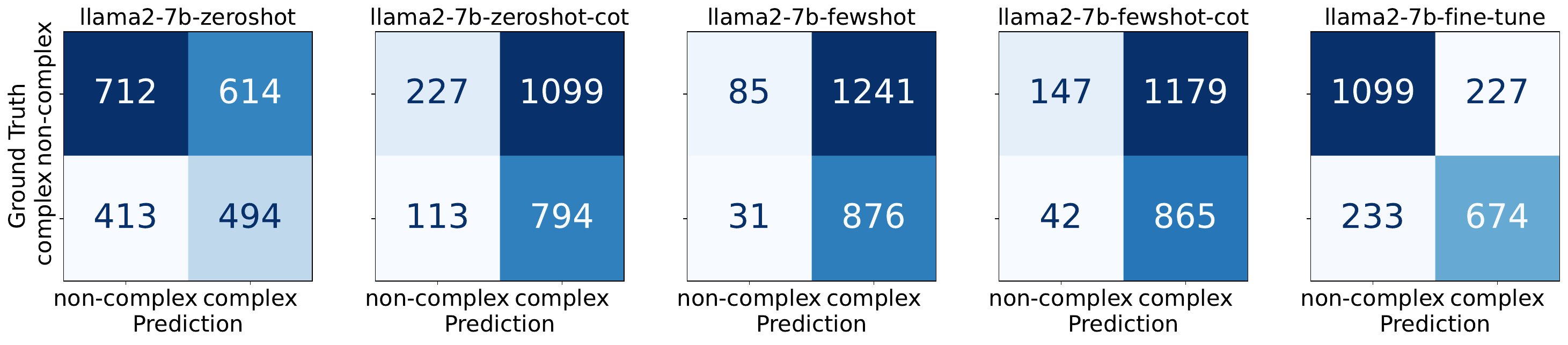}
    \caption{CWI 2018 Spanish.}
\end{subfigure}
\caption{Confusion matrices computed on the CWI 2018 datasets for Llama2 7b.}
\label{fig:cm_cwi_llama2-7b}
\end{figure*}

\begin{figure*}[!ht]
\centering
\graphicspath{{figures/cm/figs/}}
\begin{subfigure}{\textwidth}
    \centering
    \includegraphics[width=\textwidth]{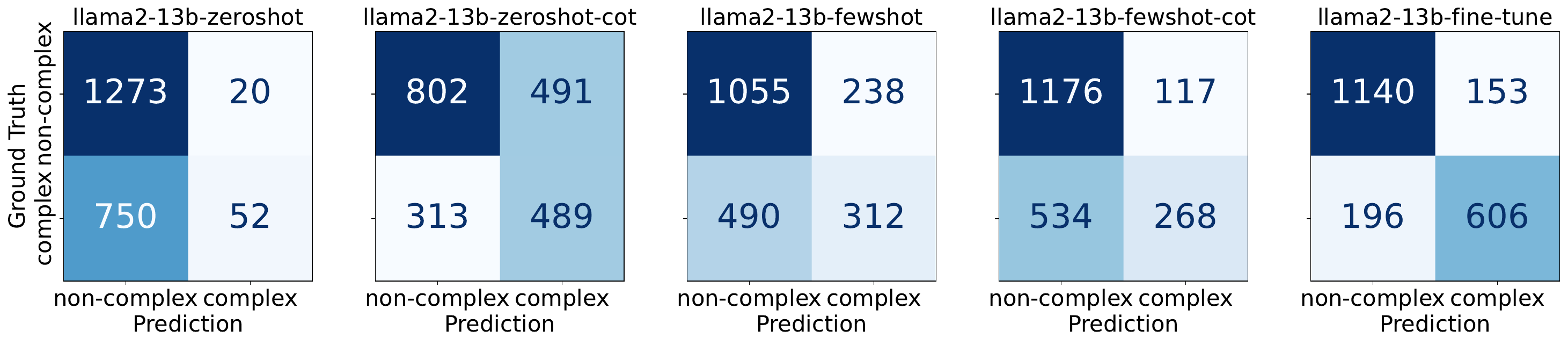}
    \caption{CWI 2018 English News.}
\end{subfigure}
\hfill
\begin{subfigure}{\textwidth}
    \centering
    \includegraphics[width=\textwidth]{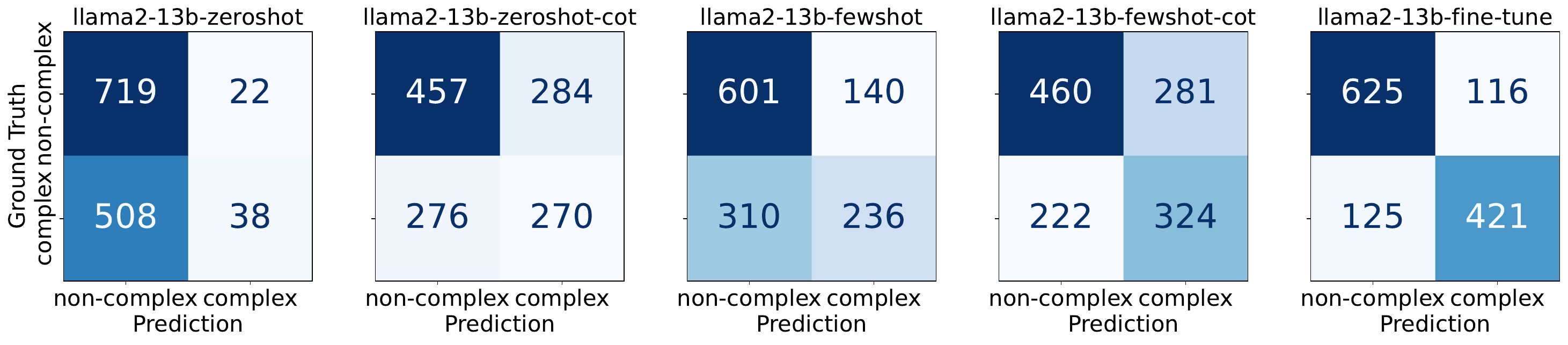}
    \caption{CWI 2018 English WikiNews.}
\end{subfigure}
\hfill
\begin{subfigure}{\textwidth}
    \centering
    \includegraphics[width=\textwidth]{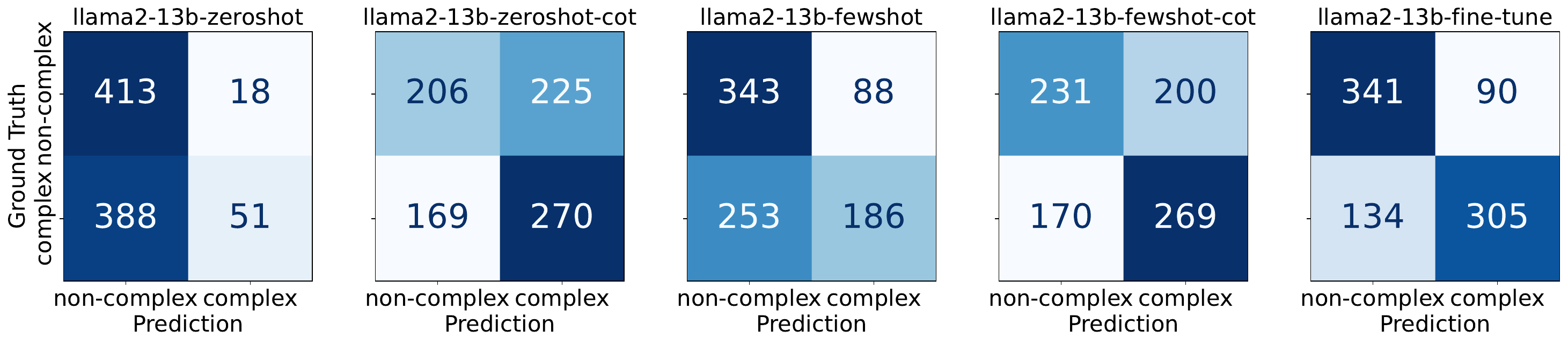}
    \caption{CWI 2018 English Wikipedia.}
\end{subfigure}
\hfill
\begin{subfigure}{\textwidth}
    \centering
    \includegraphics[width=\textwidth]{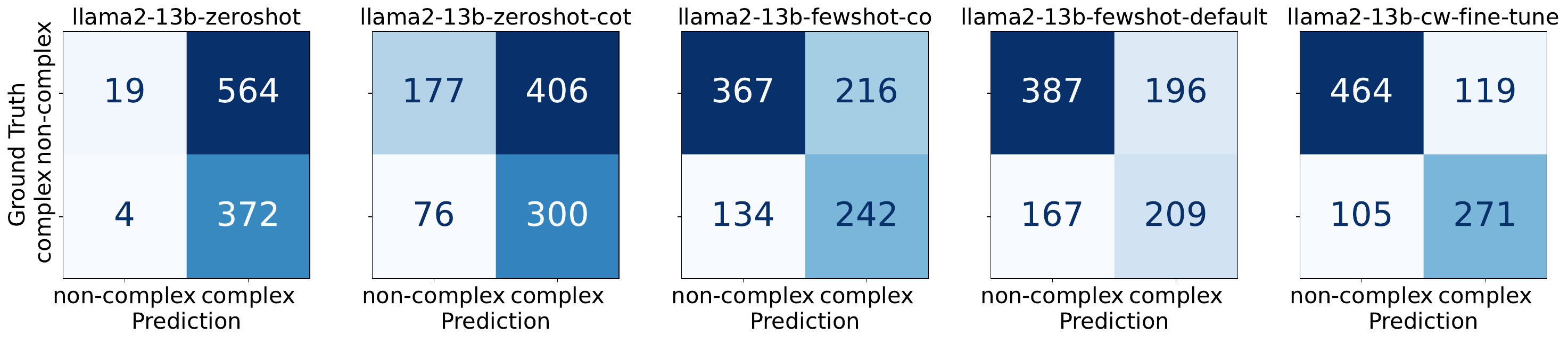}
    \caption{CWI 2018 German.}
\end{subfigure}
\hfill
\begin{subfigure}{\textwidth}
    \centering
    \includegraphics[width=\textwidth]{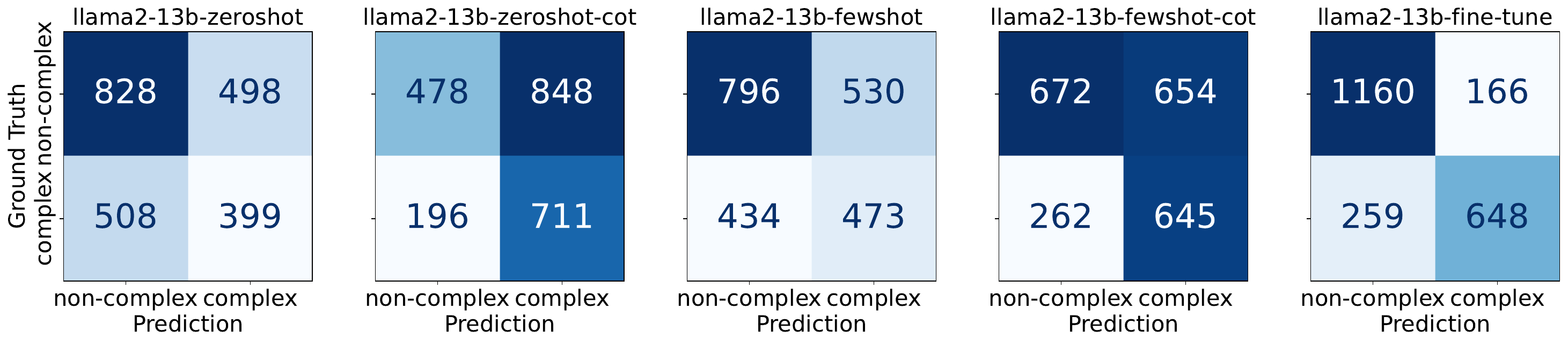}
    \caption{CWI 2018 Spanish.}
\end{subfigure}
\caption{Confusion matrices computed on the CWI 2018 datasets for Llama2 13b.}
\label{fig:cm_cwi_llama2-13b}
\end{figure*}

\begin{figure*}[!ht]
\centering
\graphicspath{{figures/cm/figs/}}
\begin{subfigure}{\textwidth}
    \centering
    \includegraphics[width=\textwidth]{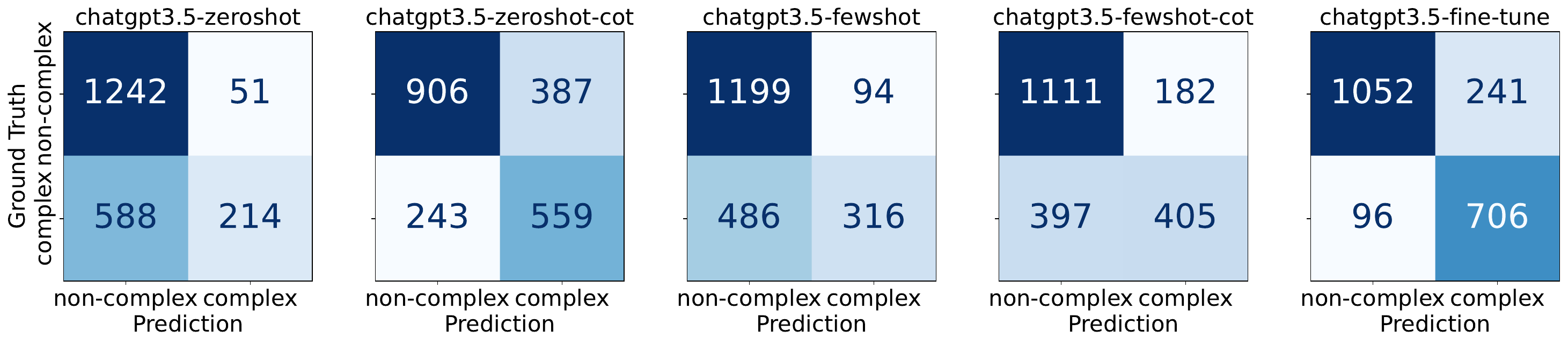}
    \caption{CWI 2018 English News.}
\end{subfigure}
\hfill
\begin{subfigure}{\textwidth}
    \centering
    \includegraphics[width=\textwidth]{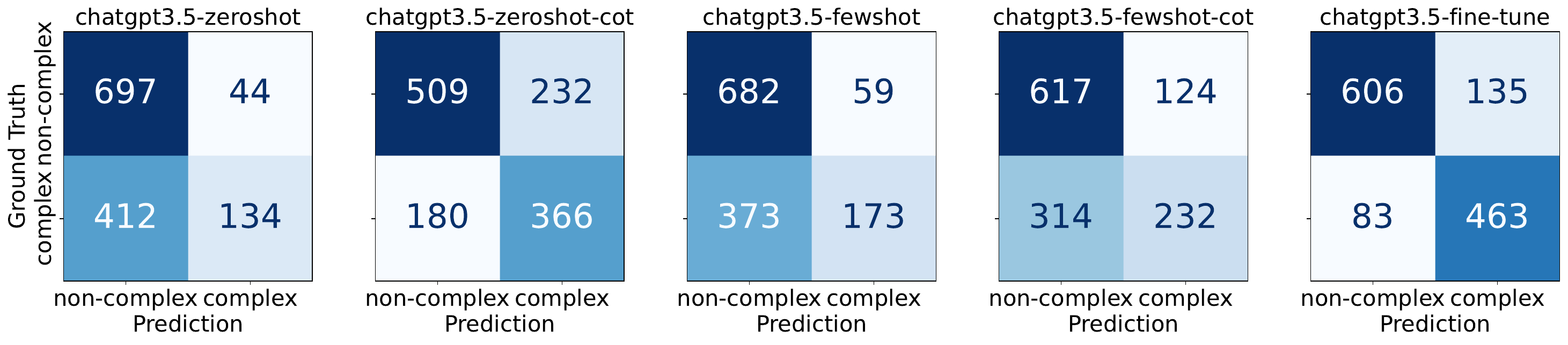}
    \caption{CWI 2018 English WikiNews.}
\end{subfigure}
\hfill
\begin{subfigure}{\textwidth}
    \centering
    \includegraphics[width=\textwidth]{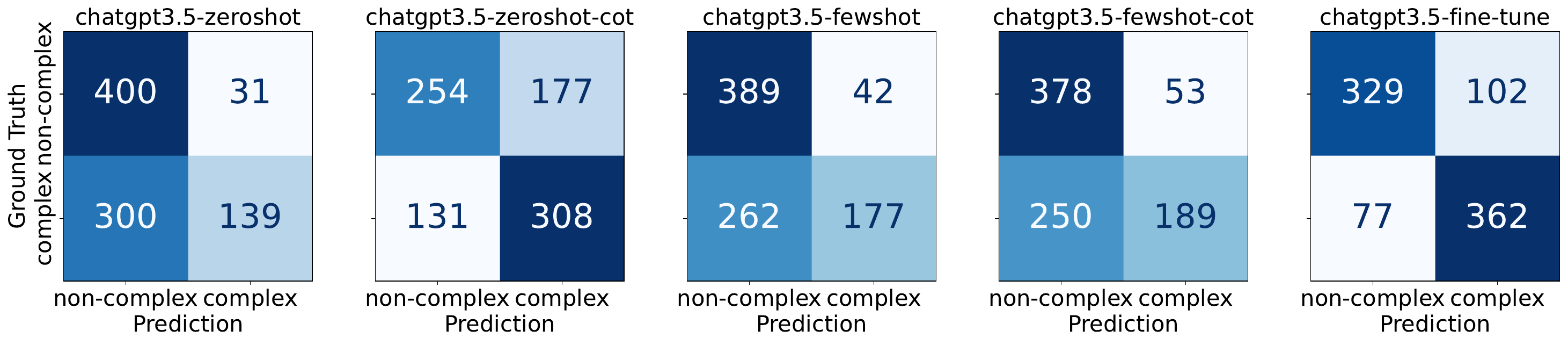}
    \caption{CWI 2018 English Wikipedia.}
\end{subfigure}
\hfill
\begin{subfigure}{\textwidth}
    \centering
    \includegraphics[width=\textwidth]{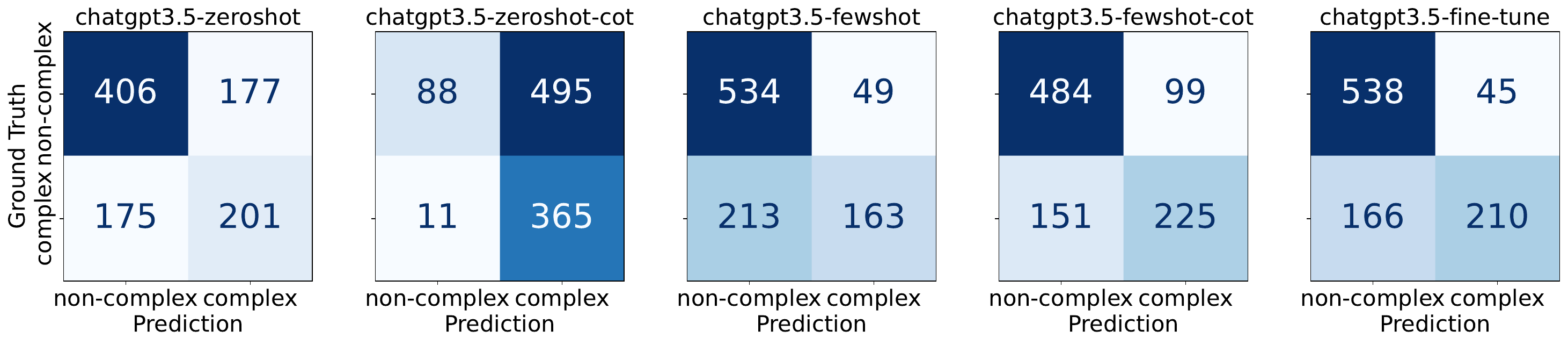}
    \caption{CWI 2018 German.}
\end{subfigure}
\hfill
\begin{subfigure}{\textwidth}
    \centering
    \includegraphics[width=\textwidth]{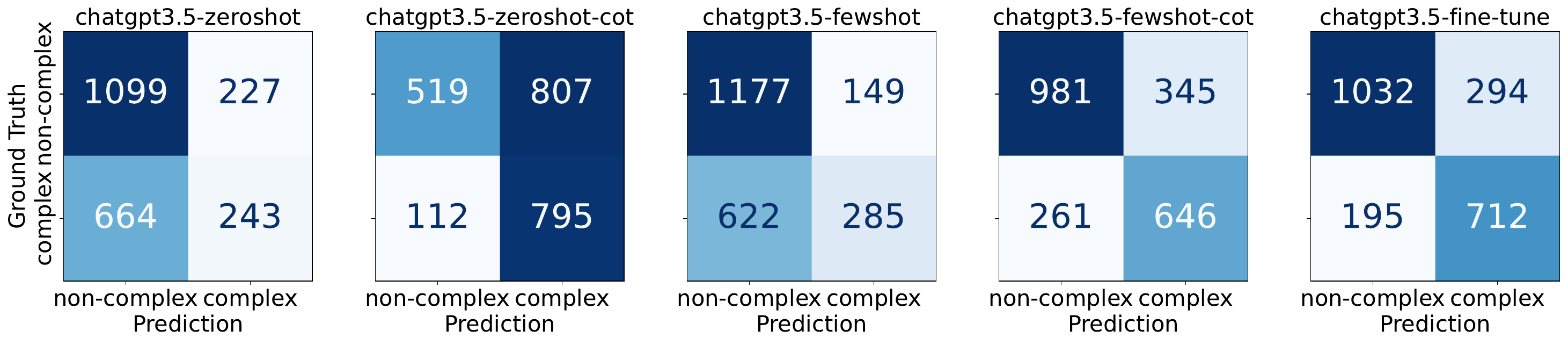}
    \caption{CWI 2018 Spanish.}
\end{subfigure}
\caption{Confusion matrices computed on the CWI 2018 datasets for ChatGPT-3.5-turbo.}
\label{fig:cm_cwi_chatgpt35}
\end{figure*}

\begin{figure*}[!ht]
\centering
\graphicspath{{figures/cm/figs/}}
\begin{subfigure}{0.8\textwidth}
    \centering
    \includegraphics[width=\textwidth]{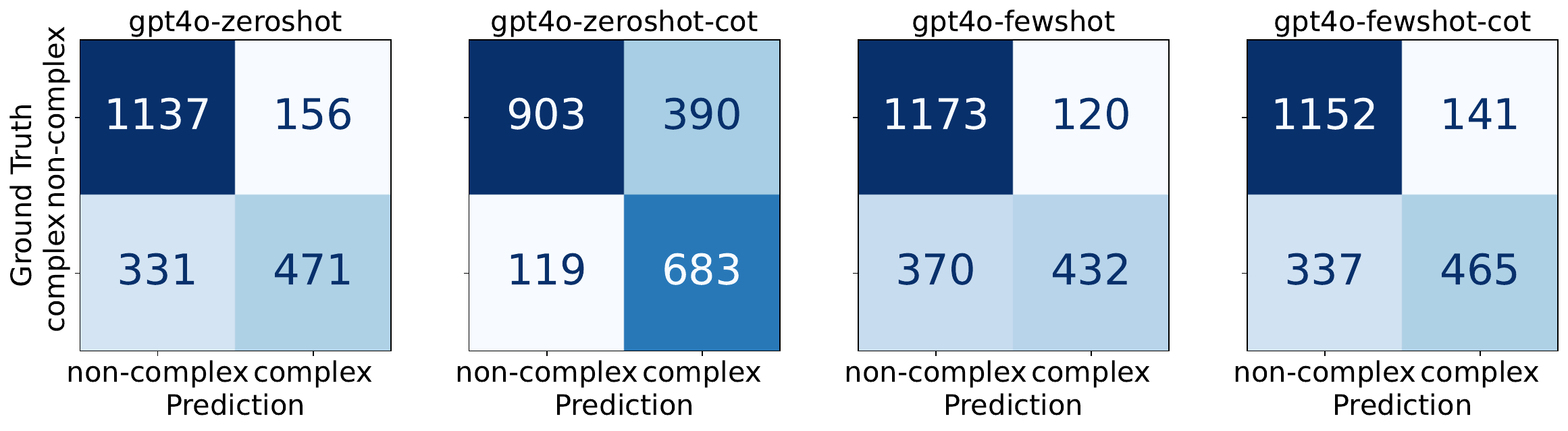}
    \caption{CWI 2018 English News.}
\end{subfigure}
\hfill
\begin{subfigure}{0.8\textwidth}
    \centering
    \includegraphics[width=\textwidth]{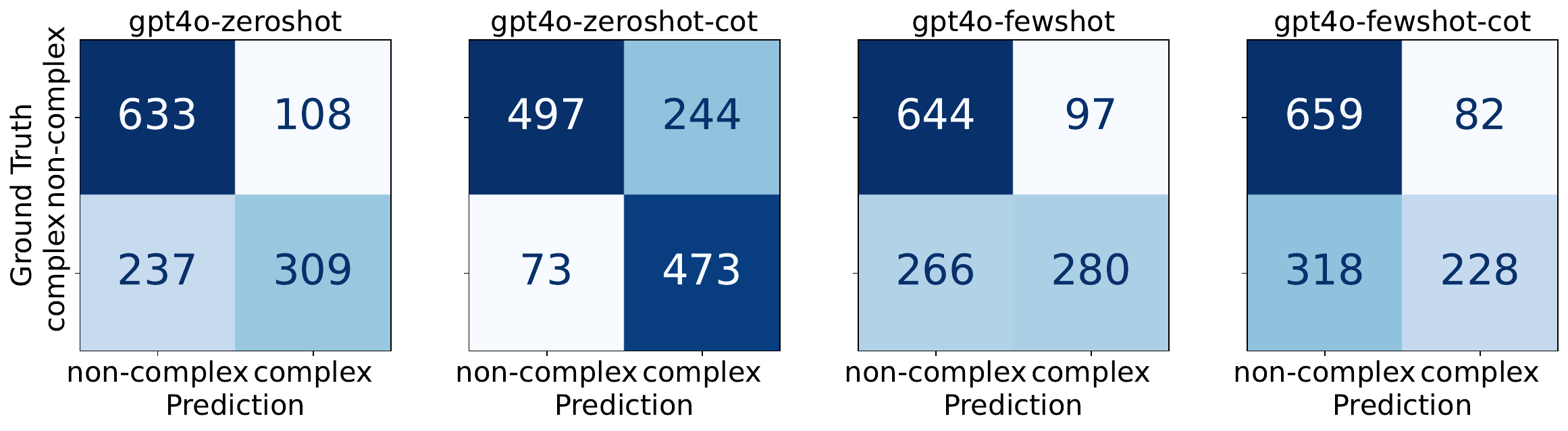}
    \caption{CWI 2018 English WikiNews.}
\end{subfigure}
\hfill
\begin{subfigure}{0.8\textwidth}
    \centering
    \includegraphics[width=\textwidth]{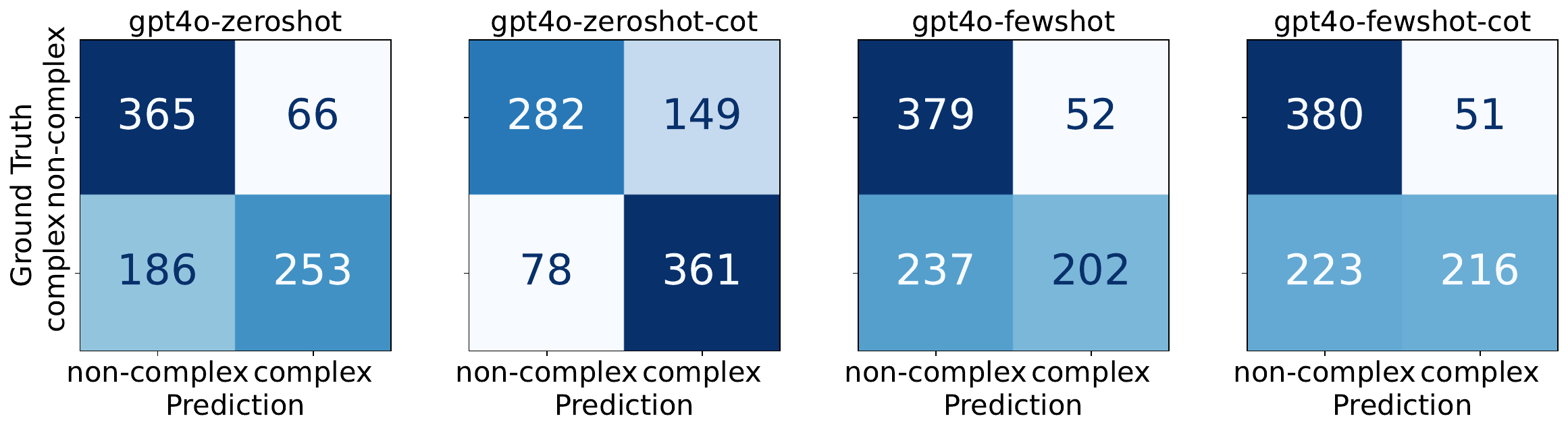}
    \caption{CWI 2018 English Wikipedia.}
\end{subfigure}
\hfill
\begin{subfigure}{0.8\textwidth}
    \centering
    \includegraphics[width=\textwidth]{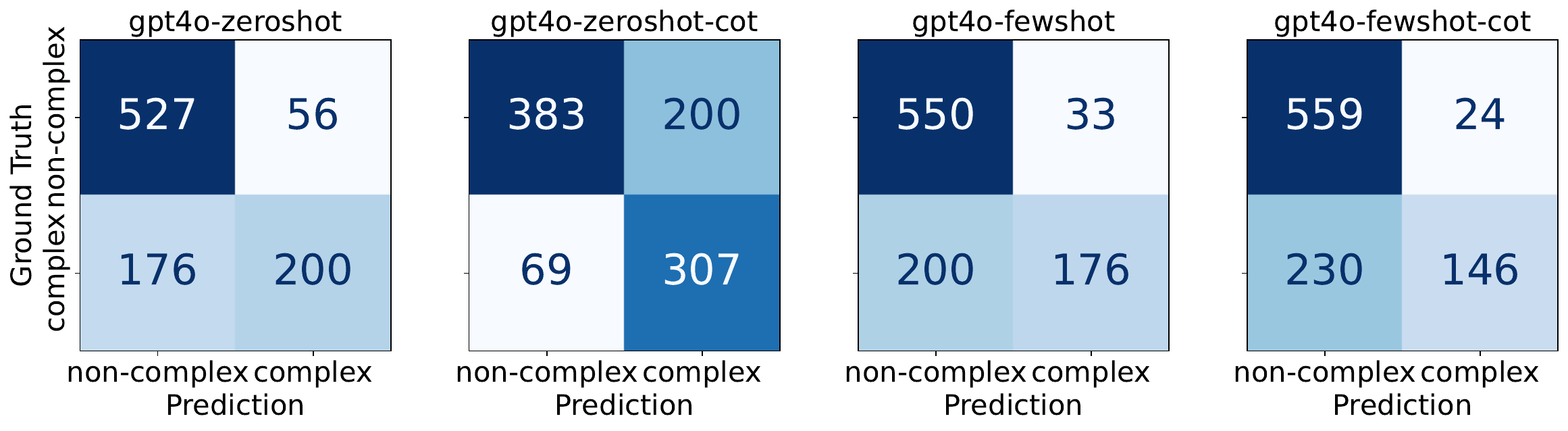}
    \caption{CWI 2018 German.}
\end{subfigure}
\hfill
\begin{subfigure}{0.8\textwidth}
    \centering
    \includegraphics[width=\textwidth]{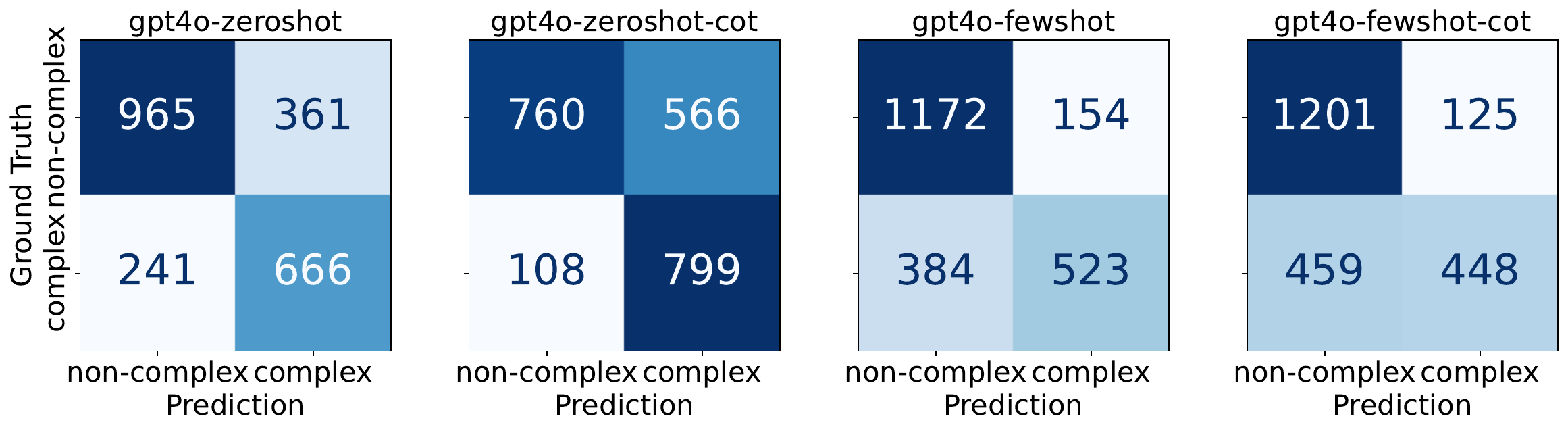}
    \caption{CWI 2018 Spanish.}
\end{subfigure}
\caption{Confusion matrices computed on the CWI 2018 datasets for GPT-4o.}
\label{fig:cm_cwi_gpt4o}
\end{figure*}

\begin{figure*}[!ht]
\centering
\graphicspath{{figures/distr/figs/}}
\begin{subfigure}{\textwidth}
    \includegraphics[width=\textwidth]{distr_llama2-7b-lcp-zeroshot-default-en_CWI-CompLex-single}
    \caption{Zero-shot.}
\end{subfigure}
\hfill
\begin{subfigure}{\textwidth}
    \includegraphics[width=\textwidth]{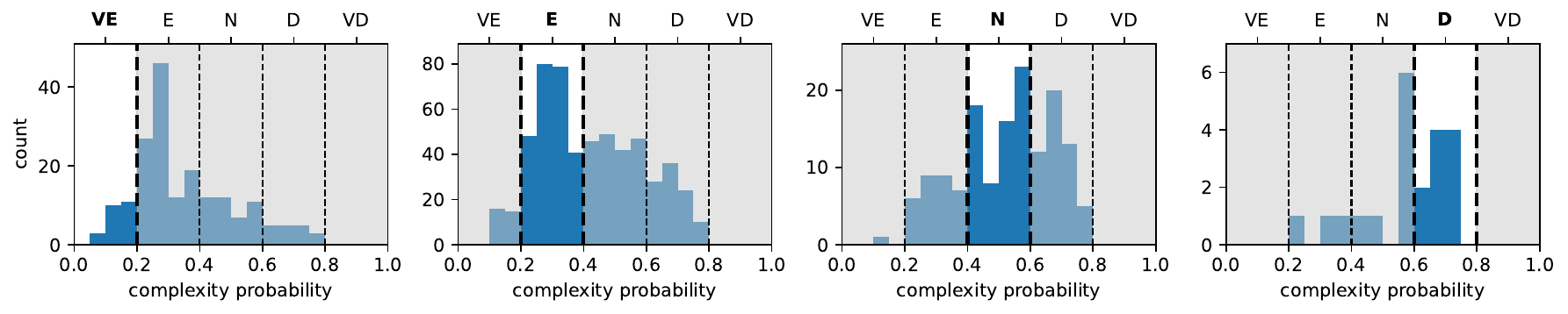}
    \caption{Zero-shot CoT.}
\end{subfigure}
\hfill
\begin{subfigure}{\textwidth}
    \includegraphics[width=\textwidth]{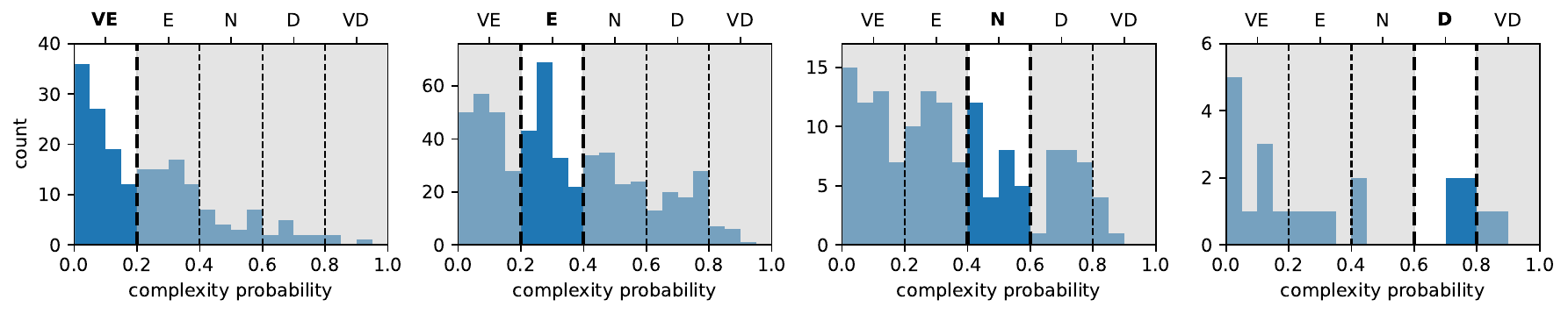}
    \caption{Few-shot.}
\end{subfigure}
\hfill
\begin{subfigure}{\textwidth}
    \includegraphics[width=\textwidth]{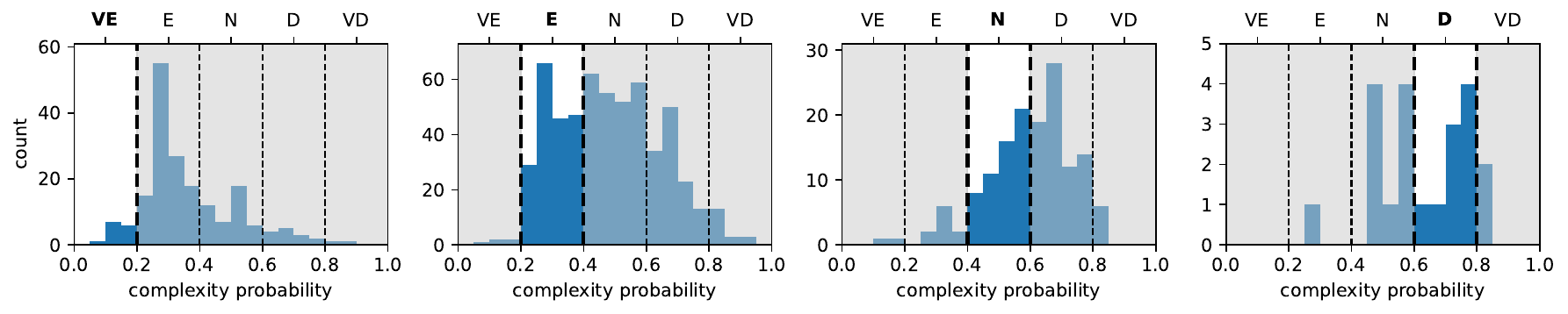}
    \caption{Few-shot CoT.}
\end{subfigure}
\hfill
\begin{subfigure}{\textwidth}
    \includegraphics[width=\textwidth]{distr_ft-llama2-7b-lcp_CWI-CompLex-single}
    \caption{Fine-tune.}
\end{subfigure}

\caption{Predictive probability distribution of Llama2 7b on LCP 2021 Single Word dataset. Highlighted in white is the ground truth interval. Neither model predicts in the VD interval. Notation: VE - very easy, E - easy, N - neutral, D - difficult, VD - very difficult.}
\label{fig:lcp_distr_llama2_7b_single}
\end{figure*}

\begin{figure*}[!ht]
\centering
\graphicspath{{figures/distr/figs/}}
\begin{subfigure}{\textwidth}
    \includegraphics[width=\textwidth]{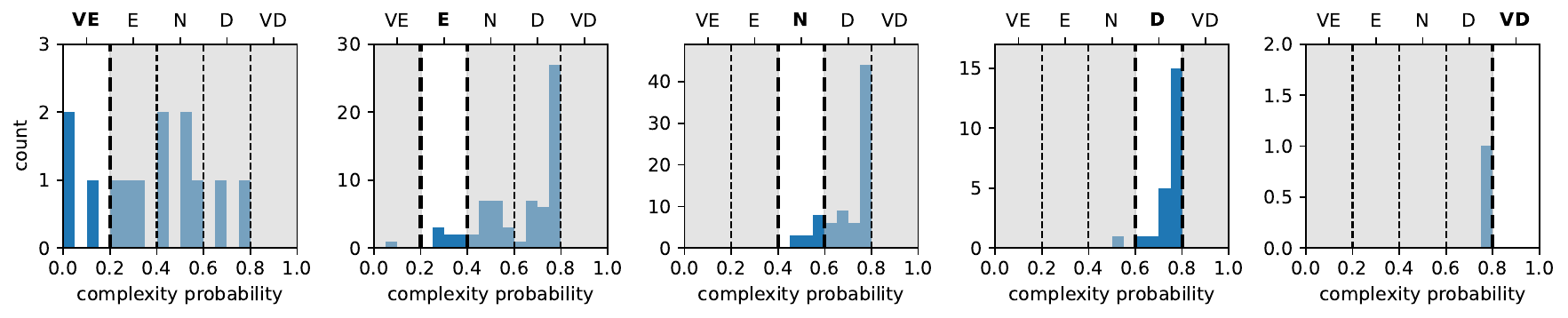}
    \caption{Zero-shot.}
\end{subfigure}
\hfill
\begin{subfigure}{\textwidth}
    \includegraphics[width=\textwidth]{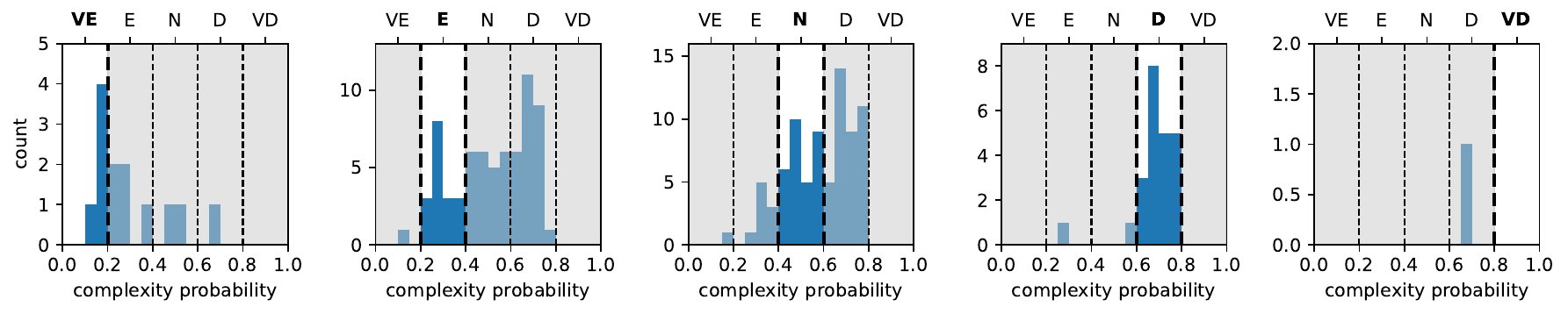}
    \caption{Zero-shot CoT.}
\end{subfigure}
\hfill
\begin{subfigure}{\textwidth}
    \includegraphics[width=\textwidth]{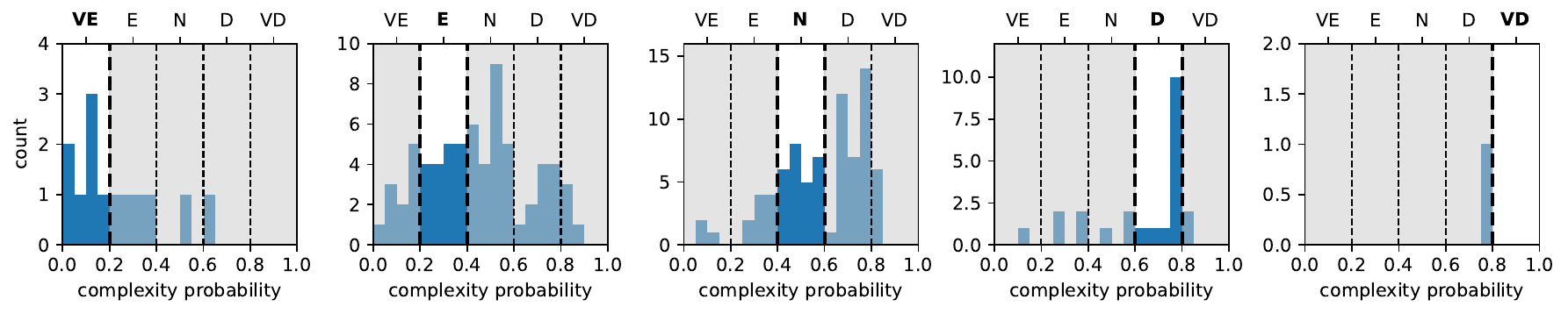}
    \caption{Few-shot.}
\end{subfigure}
\hfill
\begin{subfigure}{\textwidth}
    \includegraphics[width=\textwidth]{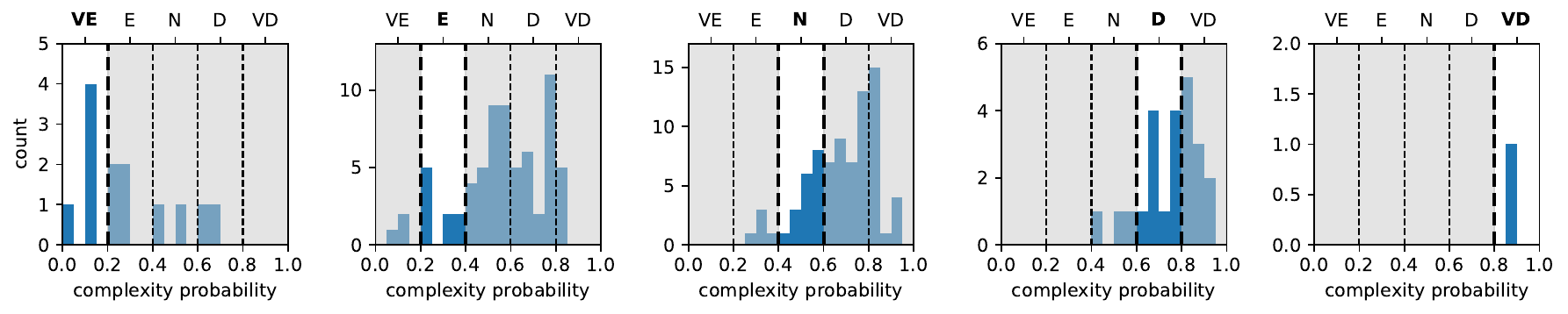}
    \caption{Few-shot CoT.}
\end{subfigure}
\hfill
\begin{subfigure}{\textwidth}
    \includegraphics[width=\textwidth]{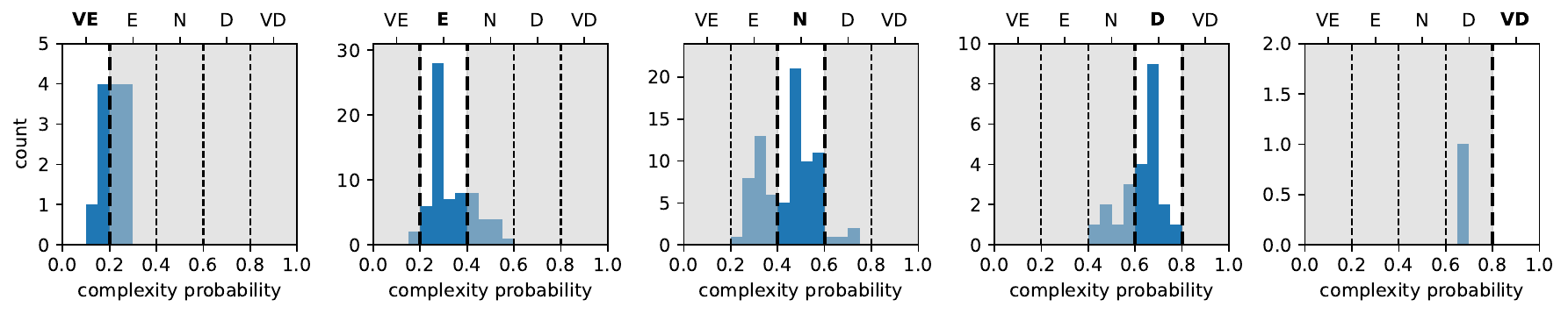}
    \caption{Fine-tune.}
\end{subfigure}

\caption{Predictive probability distribution of Llama2 7b on LCP 2021 Multi Word dataset. Highlighted in white is the ground truth interval. Neither model predicts in the VD interval. Notation: VE - very easy, E - easy, N - neutral, D - difficult, VD - very difficult.}
\label{fig:lcp_distr_llama2_7b_multi}
\end{figure*}

\begin{figure*}[!ht]
\centering
\graphicspath{{figures/distr/figs/}}
\begin{subfigure}{\textwidth}
    \includegraphics[width=\textwidth]{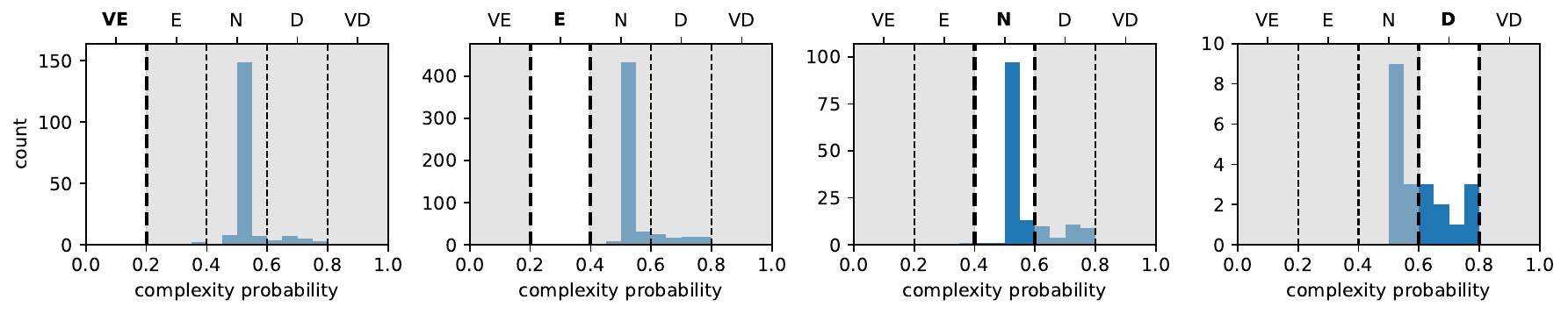}
    \caption{Zero-shot.}
\end{subfigure}
\hfill
\begin{subfigure}{\textwidth}
    \includegraphics[width=\textwidth]{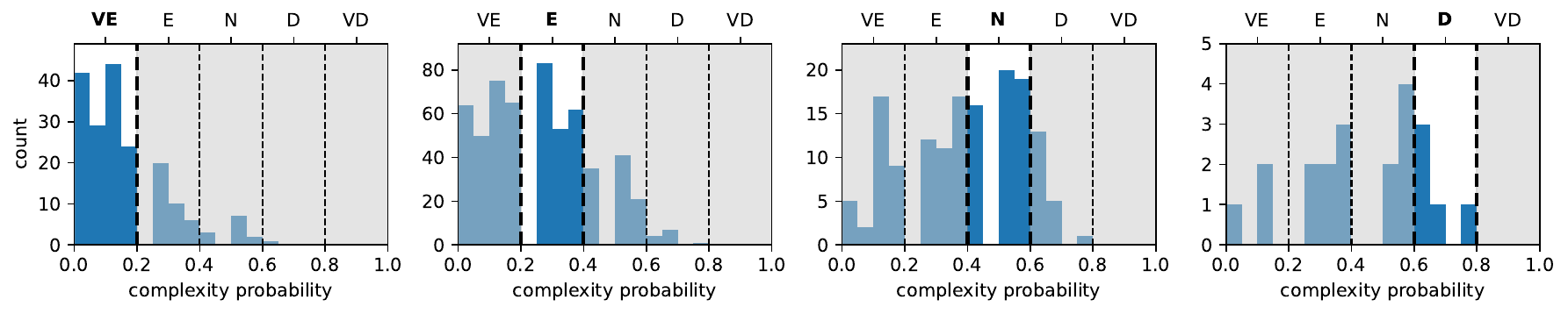}
    \caption{Zero-shot CoT.}
\end{subfigure}
\hfill
\begin{subfigure}{\textwidth}
    \includegraphics[width=\textwidth]{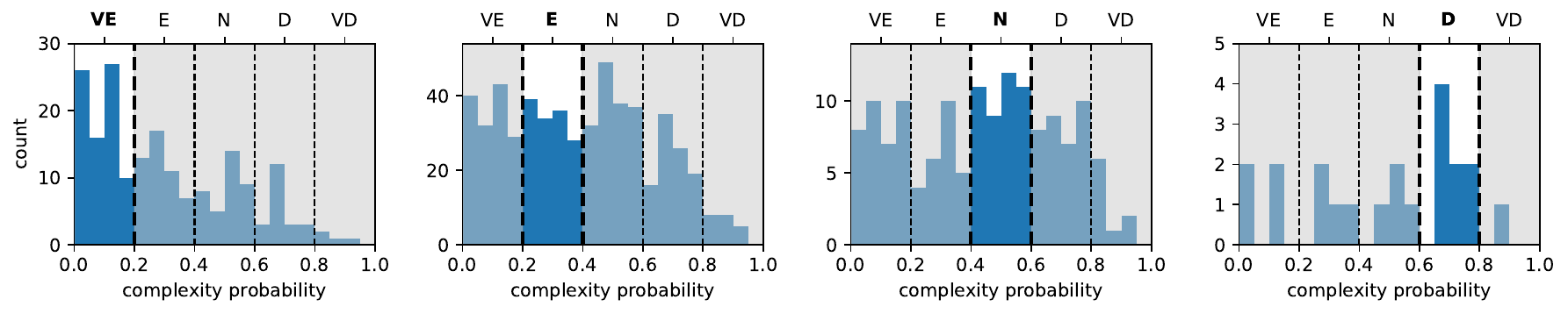}
    \caption{Few-shot.}
\end{subfigure}
\hfill
\begin{subfigure}{\textwidth}
    \includegraphics[width=\textwidth]{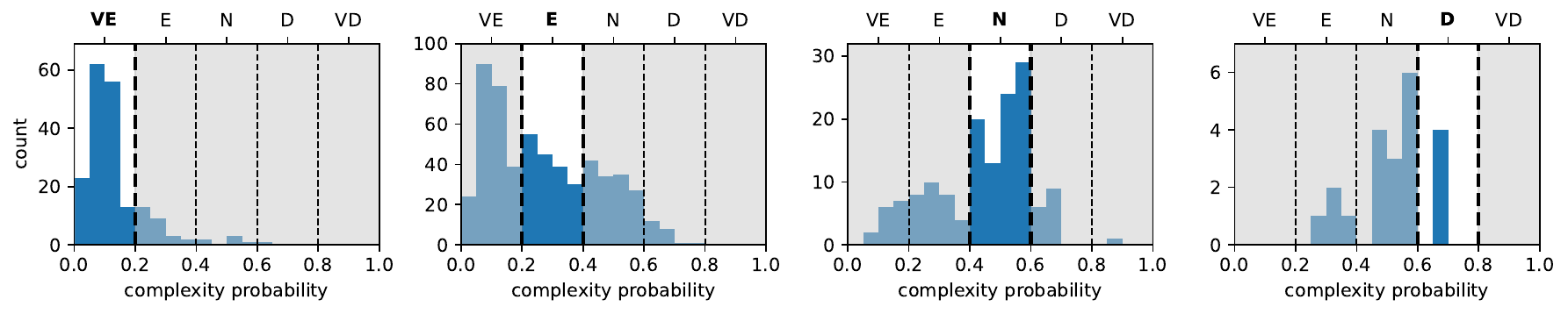}
    \caption{Few-shot CoT.}
\end{subfigure}
\hfill
\begin{subfigure}{\textwidth}
    \includegraphics[width=\textwidth]{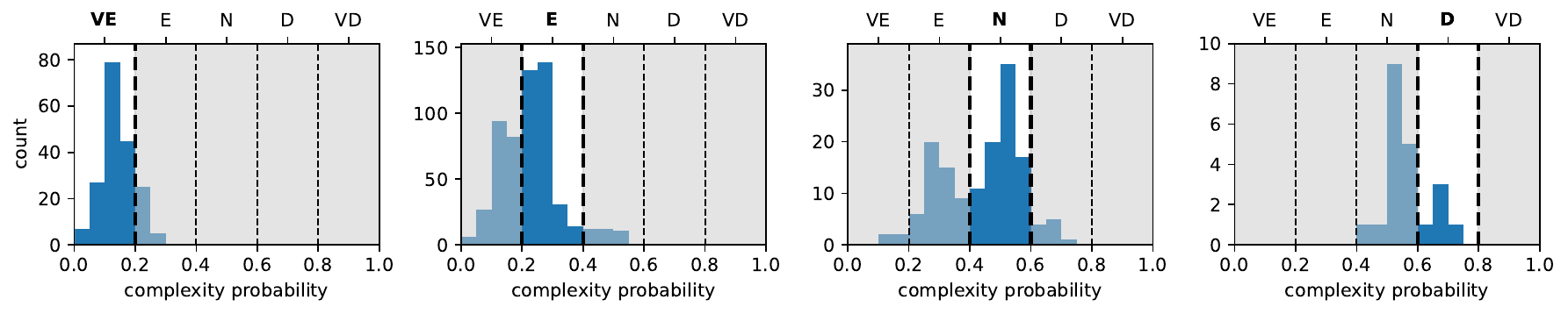}
    \caption{Fine-tune.}
\end{subfigure}

\caption{Predictive probability distribution of Llama2 13b on LCP 2021 Single Word dataset. Highlighted in white is the ground truth interval. Neither model predicts in the VD interval. Notation: VE - very easy, E - easy, N - neutral, D - difficult, VD - very difficult.}
\label{fig:lcp_distr_llama2_13b_single}
\end{figure*}

\begin{figure*}[!ht]
\centering
\graphicspath{{figures/distr/figs/}}
\begin{subfigure}{\textwidth}
    \includegraphics[width=\textwidth]{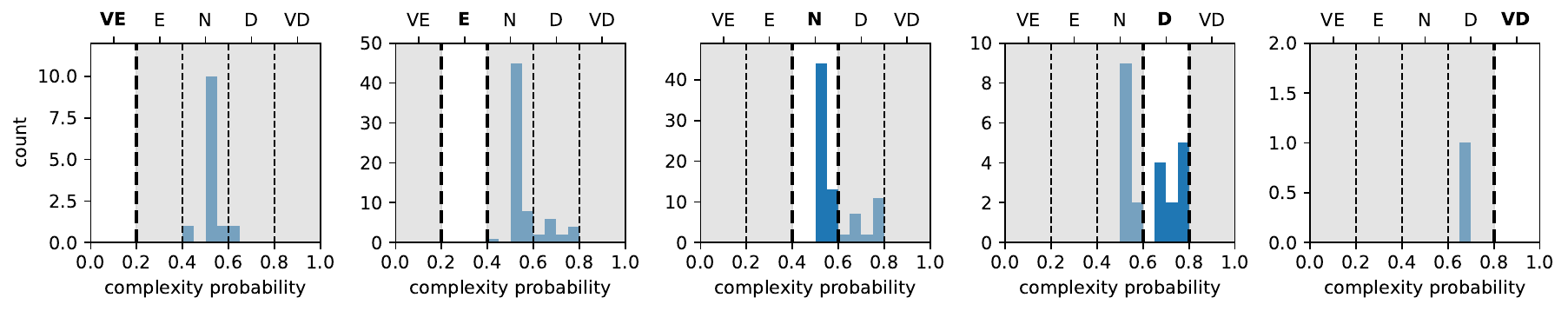}
    \caption{Zero-shot.}
\end{subfigure}
\hfill
\begin{subfigure}{\textwidth}
    \includegraphics[width=\textwidth]{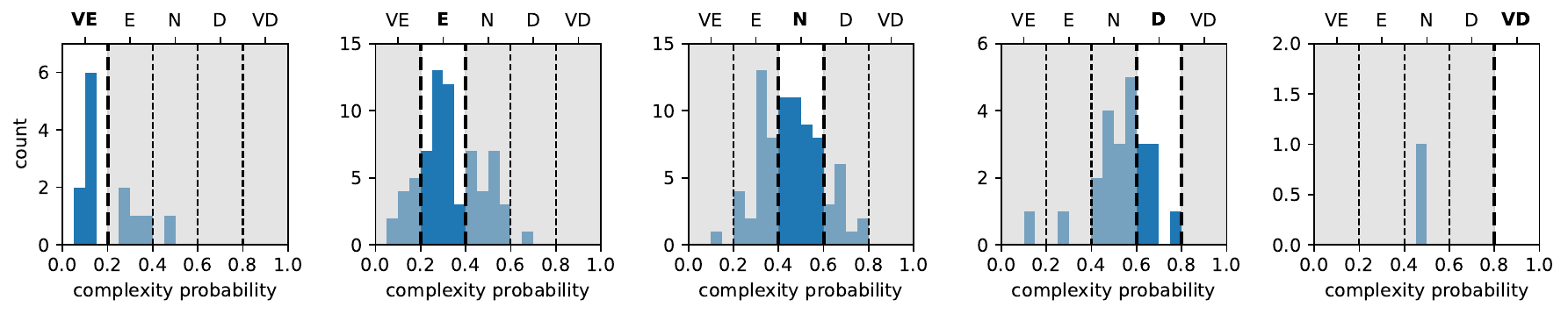}
    \caption{Zero-shot CoT.}
\end{subfigure}
\hfill
\begin{subfigure}{\textwidth}
    \includegraphics[width=\textwidth]{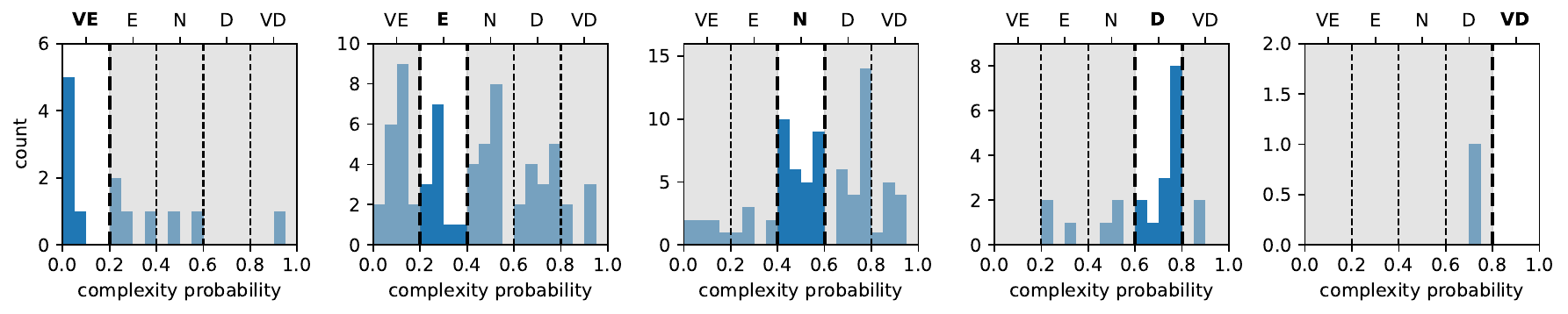}
    \caption{Few-shot.}
\end{subfigure}
\hfill
\begin{subfigure}{\textwidth}
    \includegraphics[width=\textwidth]{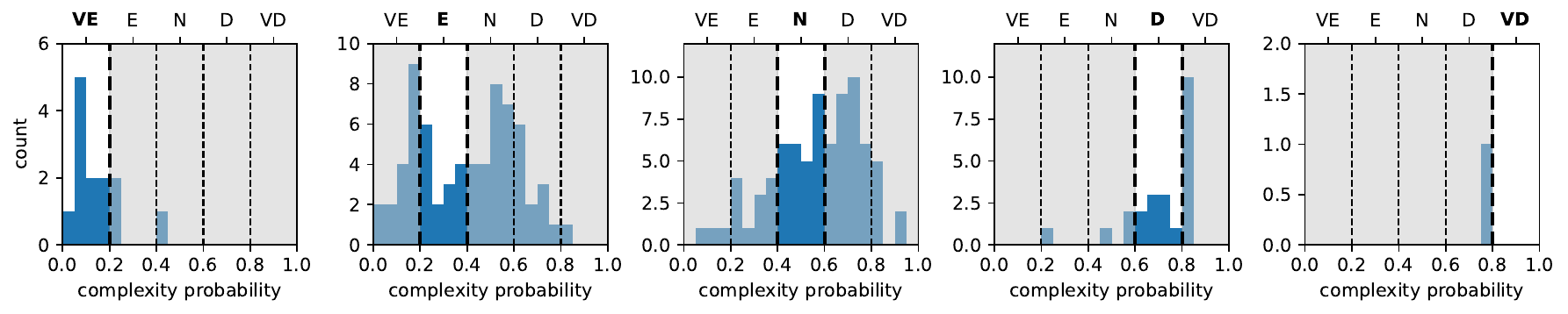}
    \caption{Few-shot CoT.}
\end{subfigure}
\hfill
\begin{subfigure}{\textwidth}
    \includegraphics[width=\textwidth]{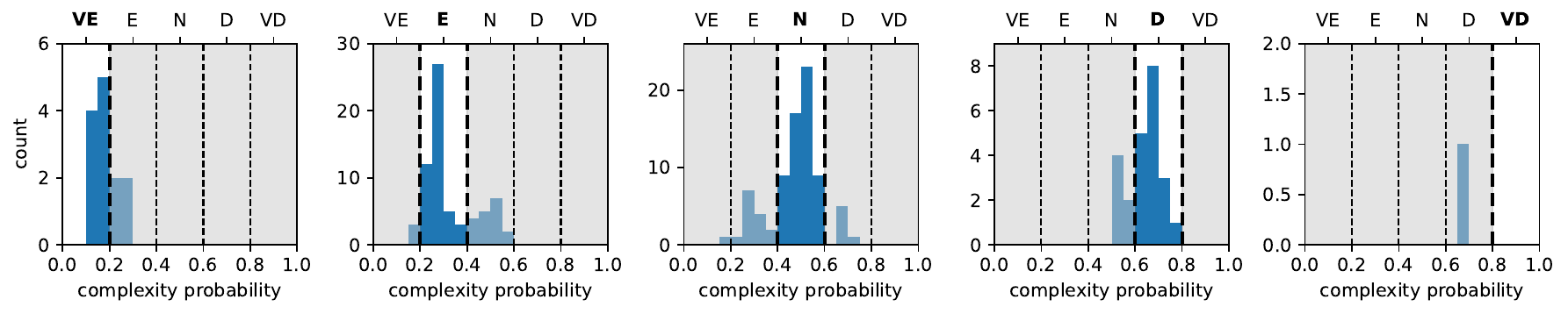}
    \caption{Fine-tune.}
\end{subfigure}

\caption{Predictive probability distribution of Llama2 13b on LCP 2021 Multi Word dataset. Highlighted in white is the ground truth interval. Neither model predicts in the VD interval. Notation: VE - very easy, E - easy, N - neutral, D - difficult, VD - very difficult.}
\label{fig:lcp_distr_llama2_13b_multi}
\end{figure*}

\begin{figure*}[!ht]
\centering
\graphicspath{{figures/distr/figs/}}
\begin{subfigure}{\textwidth}
    \includegraphics[width=\textwidth]{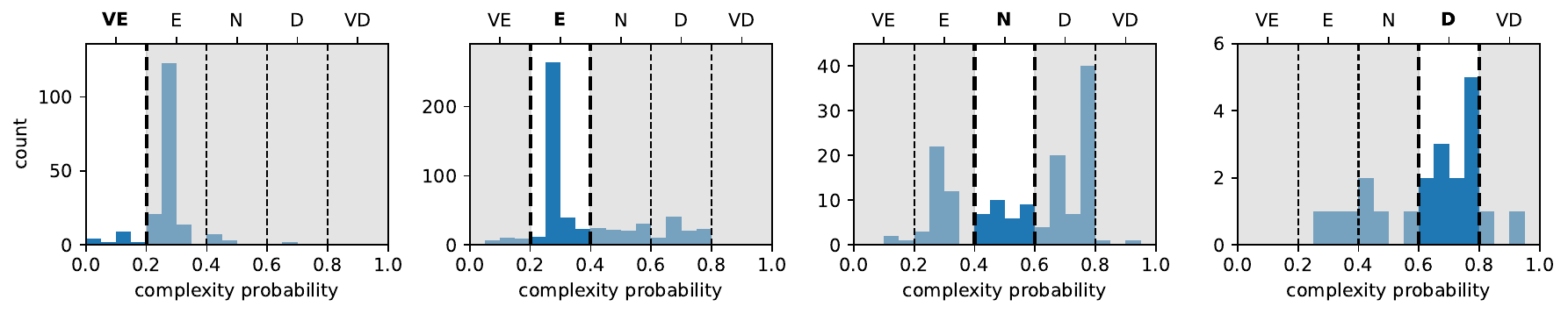}
    \caption{Zero-shot.}
\end{subfigure}
\hfill
\begin{subfigure}{\textwidth}
    \includegraphics[width=\textwidth]{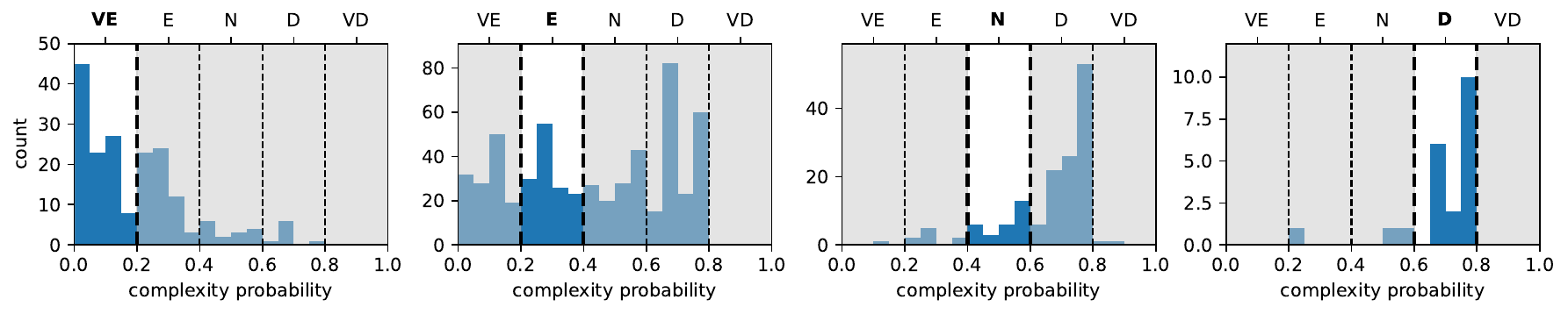}
    \caption{Zero-shot CoT.}
\end{subfigure}
\hfill
\begin{subfigure}{\textwidth}
    \includegraphics[width=\textwidth]{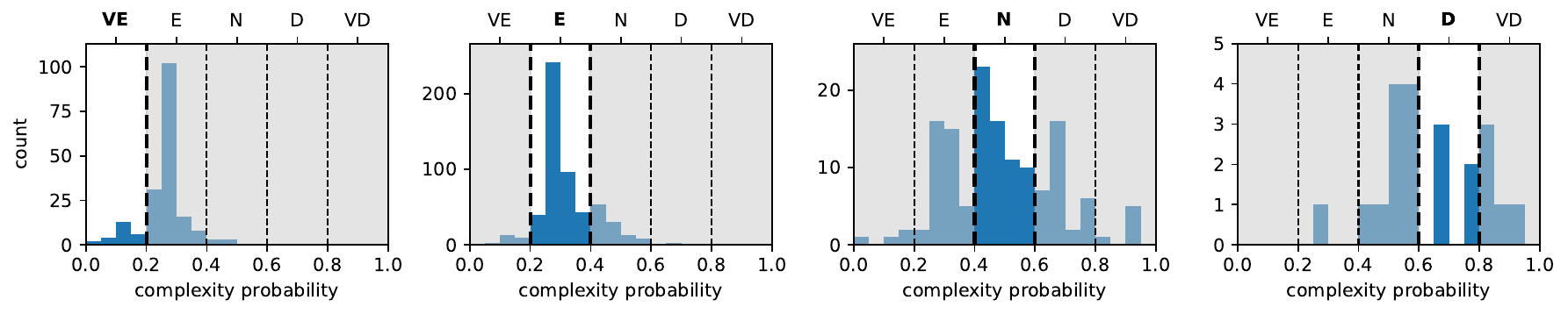}
    \caption{Few-shot.}
\end{subfigure}
\hfill
\begin{subfigure}{\textwidth}
    \includegraphics[width=\textwidth]{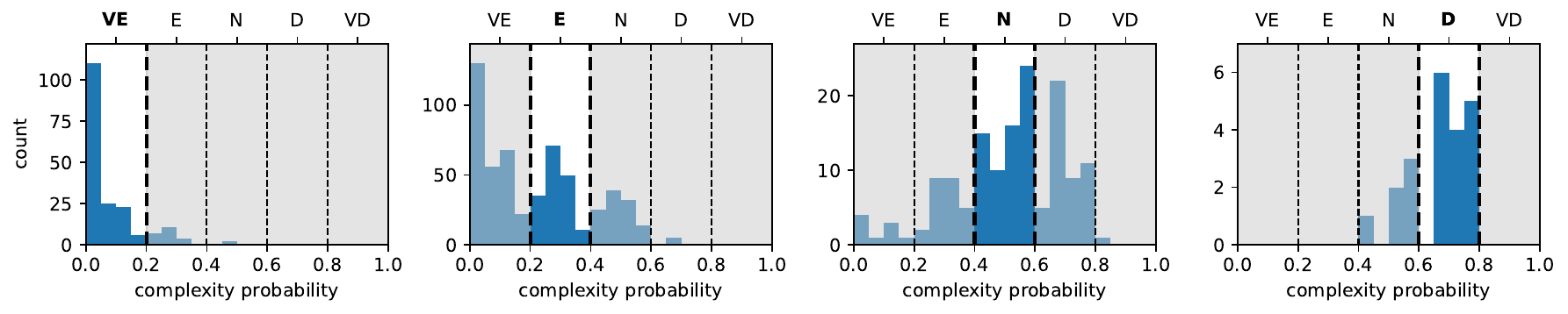}
    \caption{Few-shot CoT.}
\end{subfigure}
\hfill
\begin{subfigure}{\textwidth}
    \includegraphics[width=\textwidth]{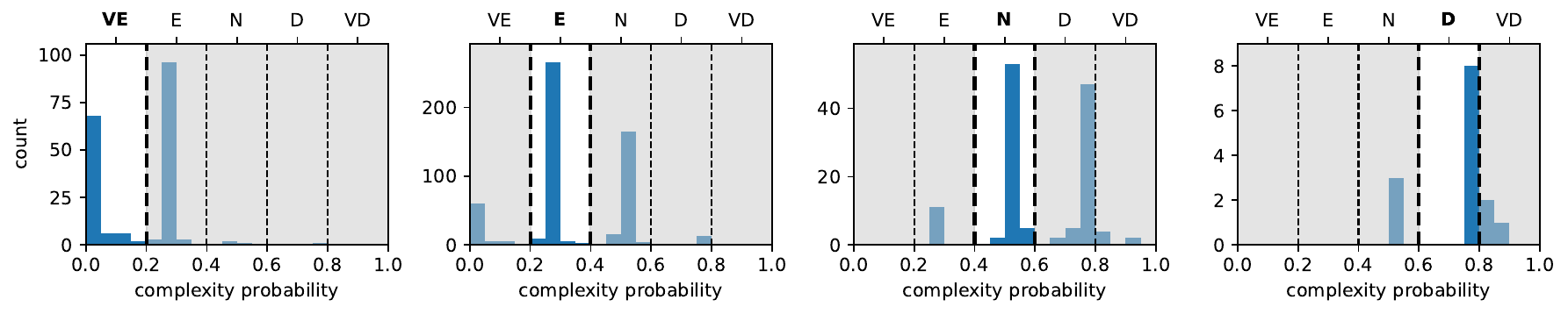}
    \caption{Fine-tune.}
\end{subfigure}

\caption{Predictive probability distribution of ChatGPT-3.5-turbo on LCP 2021 Single Word dataset. Highlighted in white is the ground truth interval. Neither model predicts in the VD interval. Notation: VE - very easy, E - easy, N - neutral, D - difficult, VD - very difficult.}
\label{fig:lcp_distr_chatgpt35_single}
\end{figure*}

\begin{figure*}[!ht]
\centering
\graphicspath{{figures/distr/figs/}}
\begin{subfigure}{\textwidth}
    \includegraphics[width=\textwidth]{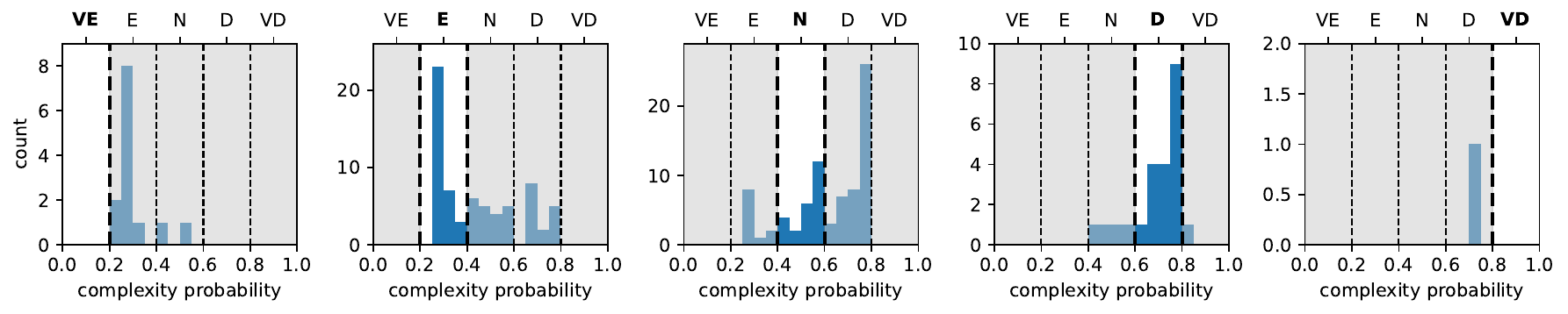}
    \caption{Zero-shot.}
\end{subfigure}
\hfill
\begin{subfigure}{\textwidth}
    \includegraphics[width=\textwidth]{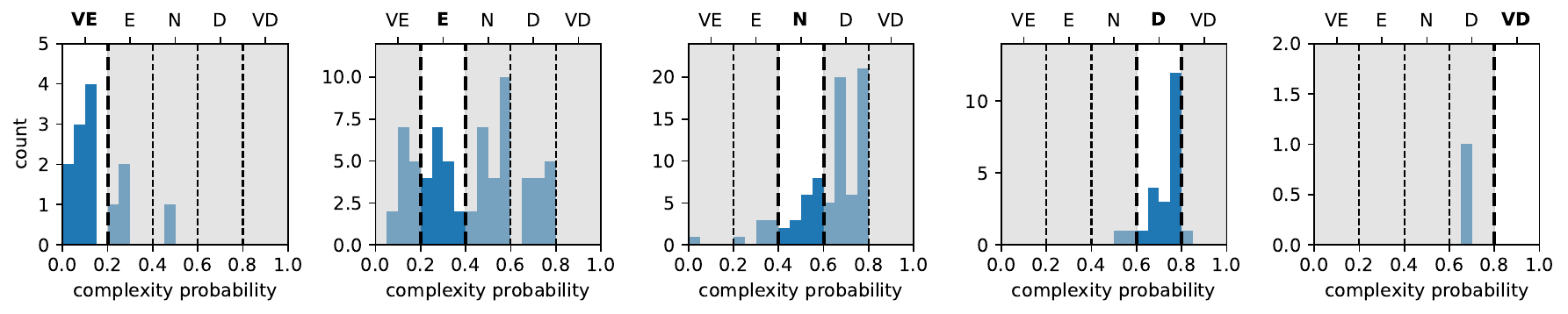}
    \caption{Zero-shot CoT.}
\end{subfigure}
\hfill
\begin{subfigure}{\textwidth}
    \includegraphics[width=\textwidth]{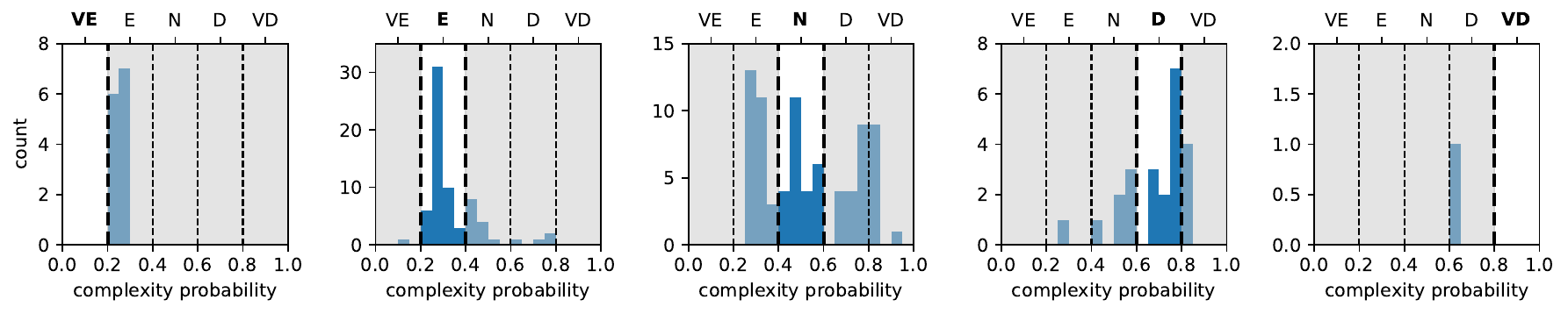}
    \caption{Few-shot.}
\end{subfigure}
\hfill
\begin{subfigure}{\textwidth}
    \includegraphics[width=\textwidth]{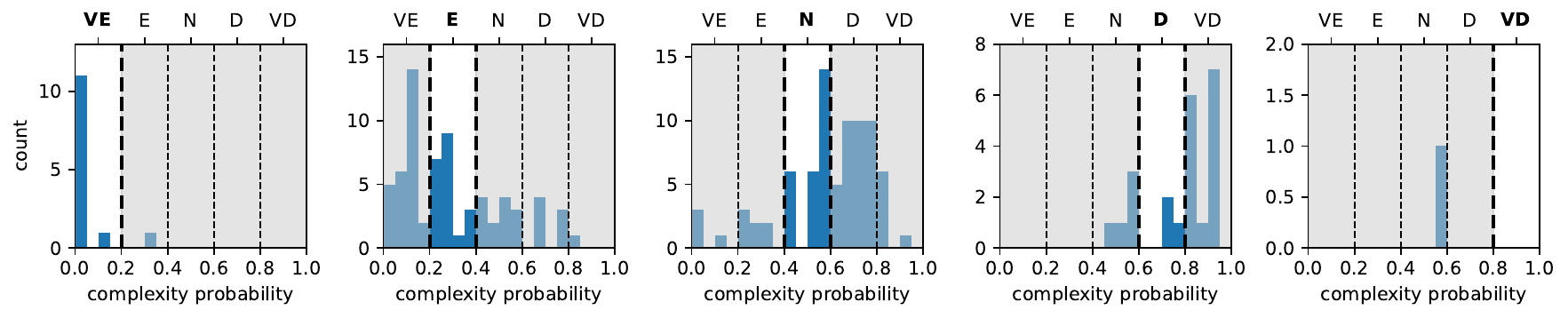}
    \caption{Few-shot CoT.}
\end{subfigure}
\hfill
\begin{subfigure}{\textwidth}
    \includegraphics[width=\textwidth]{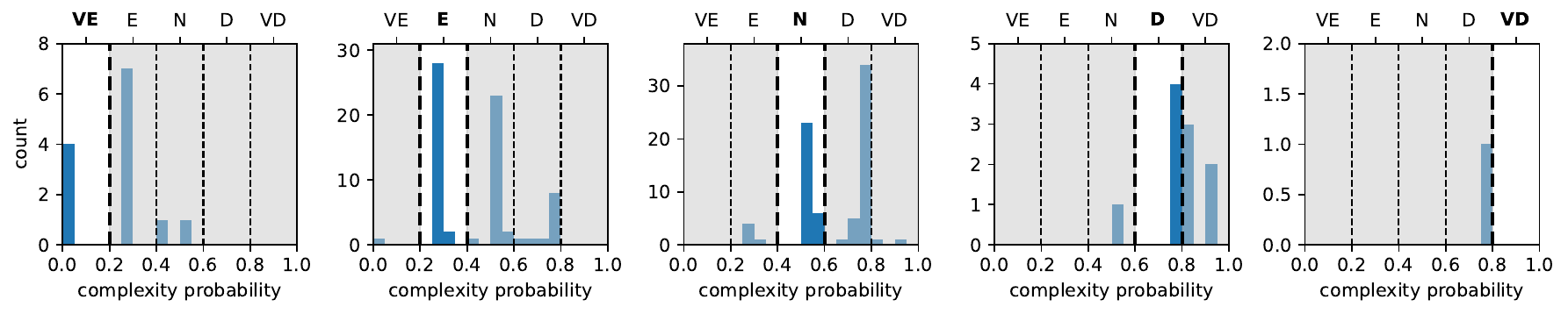}
    \caption{Fine-tune.}
\end{subfigure}

\caption{Predictive probability distribution of ChatGPT-3.5-turbo on LCP 2021 Multi Word dataset. Highlighted in white is the ground truth interval. Neither model predicts in the VD interval. Notation: VE - very easy, E - easy, N - neutral, D - difficult, VD - very difficult.}
\label{fig:lcp_distr_chatgpt35_multi}
\end{figure*}

\begin{figure*}[!ht]
\centering
\graphicspath{{figures/distr/figs/}}
\begin{subfigure}{\textwidth}
    \includegraphics[width=\textwidth]{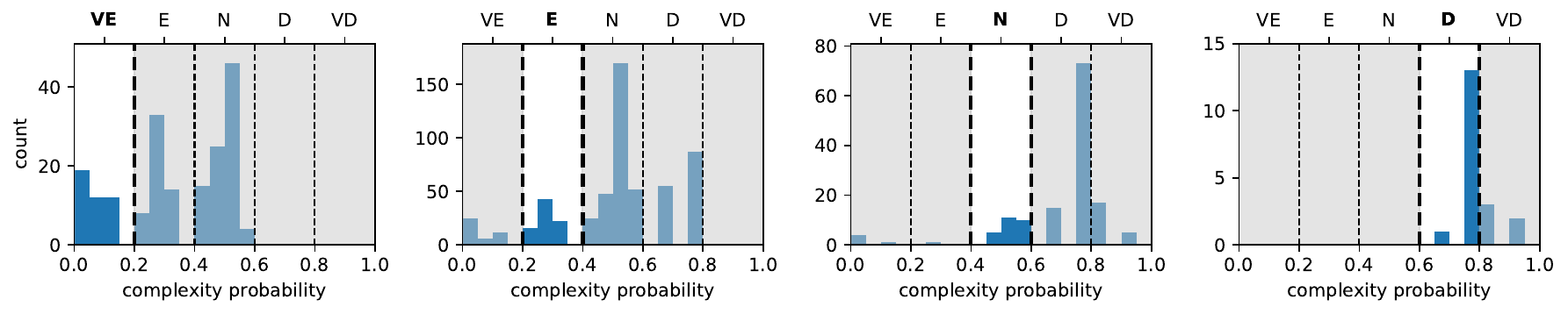}
    \caption{Zero-shot.}
\end{subfigure}
\hfill
\begin{subfigure}{\textwidth}
    \includegraphics[width=\textwidth]{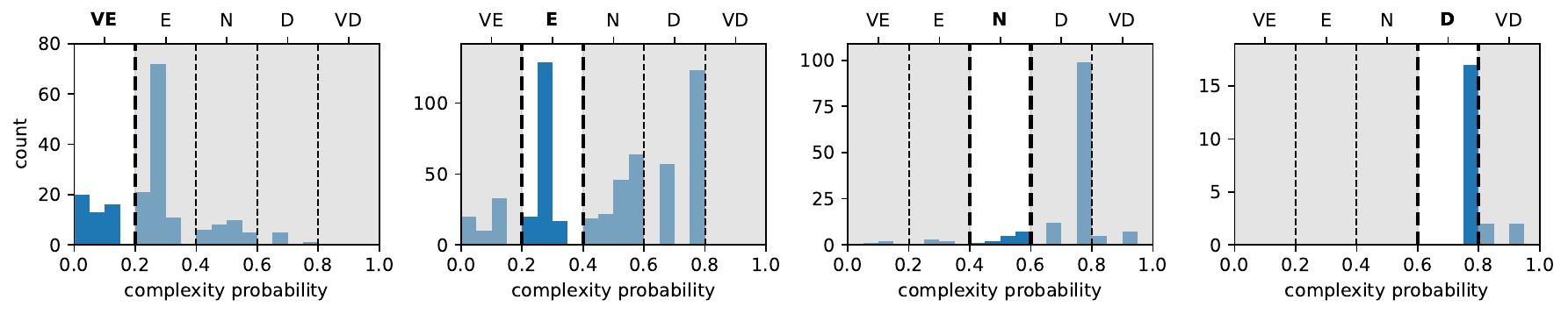}
    \caption{Zero-shot CoT.}
\end{subfigure}
\hfill
\begin{subfigure}{\textwidth}
    \includegraphics[width=\textwidth]{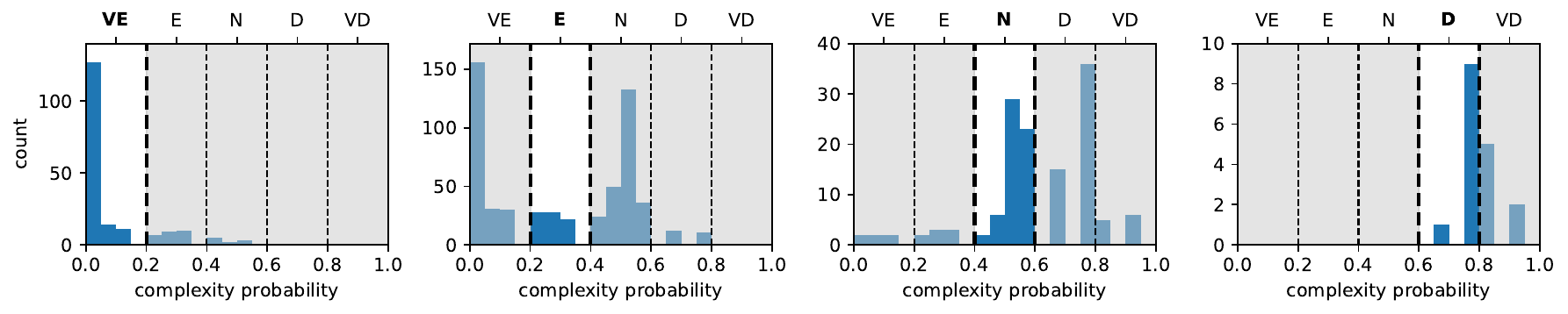}
    \caption{Few-shot.}
\end{subfigure}
\hfill
\begin{subfigure}{\textwidth}
    \includegraphics[width=\textwidth]{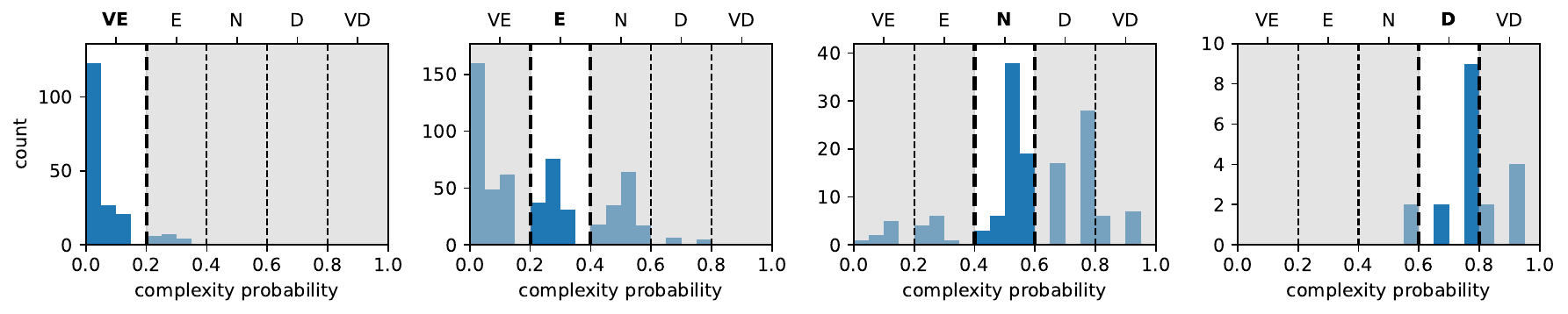}
    \caption{Few-shot CoT.}
\end{subfigure}

\caption{Predictive probability distribution of GPT-4o on LCP 2021 Single Word dataset. Highlighted in white is the ground truth interval. Neither model predicts in the VD interval. Notation: VE - very easy, E - easy, N - neutral, D - difficult, VD - very difficult.}
\label{fig:lcp_distr_gpt4o_single}
\end{figure*}

\begin{figure*}[!ht]
\centering
\graphicspath{{figures/distr/figs/}}
\begin{subfigure}{\textwidth}
    \includegraphics[width=\textwidth]{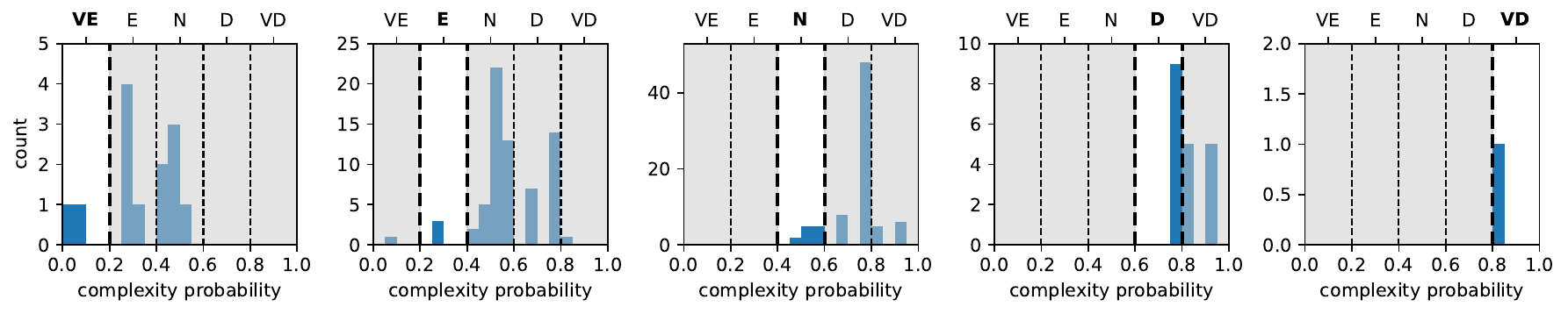}
    \caption{Zero-shot.}
\end{subfigure}
\hfill
\begin{subfigure}{\textwidth}
    \includegraphics[width=\textwidth]{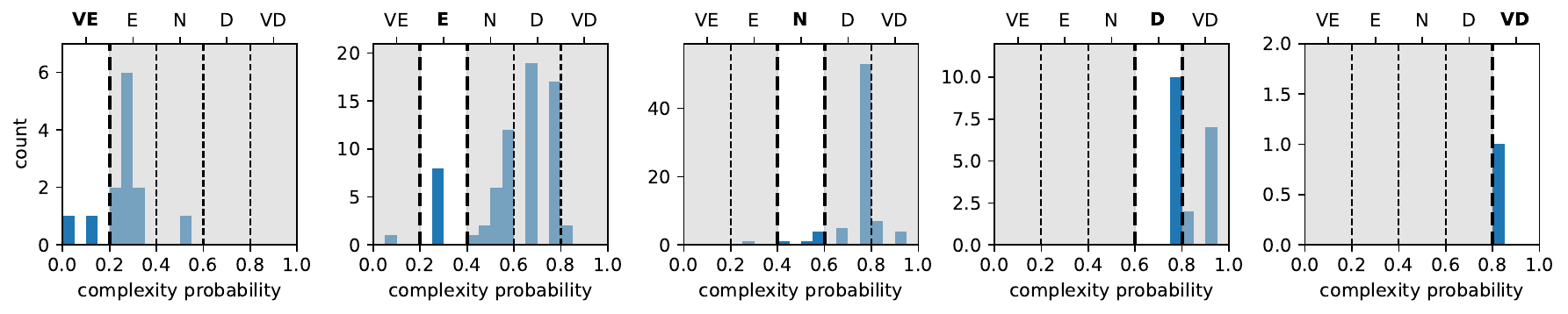}
    \caption{Zero-shot CoT.}
\end{subfigure}
\hfill
\begin{subfigure}{\textwidth}
    \includegraphics[width=\textwidth]{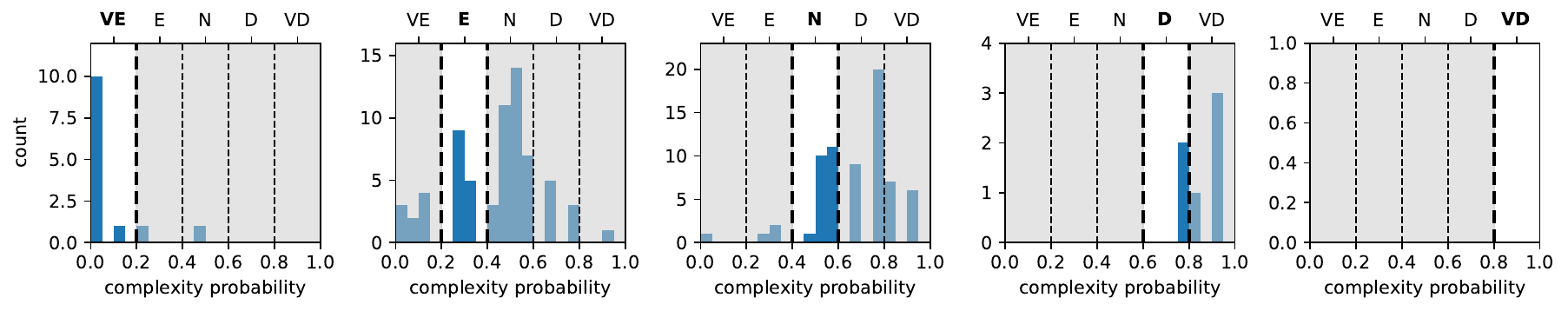}
    \caption{Few-shot.}
\end{subfigure}
\hfill
\begin{subfigure}{\textwidth}
    \includegraphics[width=\textwidth]{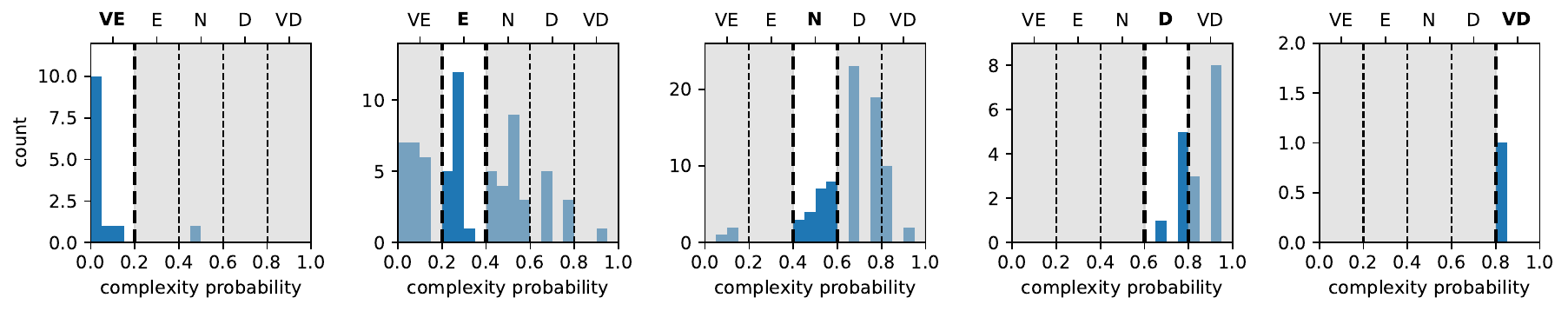}
    \caption{Few-shot CoT.}
\end{subfigure}

\caption{Predictive probability distribution of GPT-4o on LCP 2021 Multi Word dataset. Highlighted in white is the ground truth interval. Neither model predicts in the VD interval. Notation: VE - very easy, E - easy, N - neutral, D - difficult, VD - very difficult.}
\label{fig:lcp_distr_gpt4o_multi}
\end{figure*}

LLMs can grasp word complexity, depending on the model's capabilities. We observed that performances across domains and language, and whether we deal with a word or a phrase, are similar if the model is fine-tuned. In the zero-shot setting, the input prompt and prediction temperature yield a high variance across the results. Also, we noticed that sometimes the models (especially Llama-2-13b-chat, in the zero-shot setting) refused to predict some examples (especially in the Biblical domain) because of racial discrimination, despite that not being the case. Models tend to consider words easier than they are, mainly because if prompted to explain the choice, they could provide another synonym that is not necessarily simpler. Zero-shot prompting is achieved every time poor performances are detected, and the main effect is that models tend to have a high false positive rate in the CWI task. This can be changed during fine-tuning when we notice that imbalanced datasets towards a class lead to the model being biased and producing more often the predominant label from the fine-tuning set. We include supporting figures for confusion matrices on the CWI task in Figures \ref{fig:cm_cwi_llama2-7b}, \ref{fig:cm_cwi_llama2-13b}, \ref{fig:cm_cwi_chatgpt35}, and \ref{fig:cm_cwi_gpt4o} as well as the prediction distributions on the LCP task in Figures \ref{fig:lcp_distr_llama2_7b_single}, \ref{fig:lcp_distr_llama2_7b_multi}, \ref{fig:lcp_distr_llama2_13b_single}, \ref{fig:lcp_distr_llama2_13b_multi}, \ref{fig:lcp_distr_chatgpt35_single}, \ref{fig:lcp_distr_chatgpt35_multi}, \ref{fig:lcp_distr_gpt4o_single}, and \ref{fig:lcp_distr_gpt4o_multi}.

\begin{table*}[!htb]
\centering
\small
\begin{tabular}{p{4.8cm}p{1cm}p{1cm}p{4.8cm}p{1.2cm}}
\toprule
\multicolumn{1}{c}{\textbf{Sentence}} &  \multicolumn{1}{c}{\textbf{Word}} &  \multicolumn{1}{c}{\textbf{Answer}} &  \multicolumn{1}{c}{\textbf{Proof}} &  \multicolumn{1}{p{1.1cm}}{\textbf{Ground Truth}} \\
\midrule
\hline
Toronto traded Stewart to Chicago early in the 1947-48 season in a deal that brought Max Bentley to the Leafs. & brought & false & The verb 'brought' is not complex as it is a commonly used verb in everyday language with a straightforward meaning & true \\
\hline
His use of Russian language formed the basis of the style of novelists Ivan Turgenev, Ivan Goncharov, and Leo Tolstoy, as well as that of subsequent lyric poets such as Mikhail Lermontov. & basis & false & The word 'basis' is not particularly complex, but the concept it represents (i.e., the foundation or starting point of something) may be unfamiliar to some beginner English learners & true \\
\hline
Vertical distance measurements in the "down" direction are commonly referred to as depth. & depth & true & The word 'depth' has multiple meanings, including a vertical extent or height, making it potentially complex for beginning English learners to understand without proper context or explanation & true \\
\hline
The lack of oxygen above 2,400 meters (8,000 ft) can cause serious illnesses such as altitude sickness, high altitude pulmonary edema, and high altitude cerebral edema. & ft & true & The abbreviation 'ft' is commonly used in English to represent feet, which is a unit of measurement. However, in this context, it may be challenging for beginners to understand because they might not be familiar with the abbreviation. & false \\
\bottomrule
\end{tabular}
\caption{Examples of predictions and proofs for the Llama-2-13b-chat model on the CWI English Wikipedia dataset in the zero-shot chain-of-thought setting.}
\label{tab:llm_proofs_llama2_13b_cwi}
\end{table*}

\end{document}